**Sampling-enabled scalable manifold learning unveils the discriminative cluster structure of high-dimensional data**

Dehua Peng[1, 2, 3], Zhipeng Gui[1, 3, *], Wenzhang Wei[1], Fa Li[4], Jie Gui[5], Huayi Wu[2, 3], and Jianya Gong[2, 3]

[1]School of Remote Sensing and Information Engineering, Wuhan University, Wuhan, China.
[2]State Key Laboratory of Information Engineering in Surveying, Mapping and Remote Sensing, Wuhan University, Wuhan, China.
[3]Collaborative Innovation Center of Geospatial Technology, Wuhan University, Wuhan, China.
[4]Department of Earth System Science, Stanford University, Stanford, CA, USA.
[5]School of Cyber Science and Engineering, Southeast University, Nanjing, China.
*Correspondence to: Email: zhipeng.gui@whu.edu.cn (Z.G.)

**Abstract:** As a pivotal branch of machine learning, manifold learning uncovers the intrinsic low-dimensional structure within complex nonlinear manifolds in high-dimensional space for visualization, classification, clustering, and gaining key insights. Although existing techniques have achieved remarkable successes, they suffer from extensive distortions of cluster structure, which hinders the understanding of underlying patterns. Scalability issues also limit their applicability for handling large-scale data. We hence propose a sampling-based Scalable manifold learning technique that enables Uniform and Discriminative Embedding, namely SUDE, for large-scale and high-dimensional data. It starts by seeking a set of landmarks to construct the low-dimensional skeleton of the entire data, and then incorporates the non-landmarks into the learned space based on the constrained locally linear embedding (CLLE). We empirically validated the effectiveness of SUDE on synthetic datasets and real-world benchmarks, and applied it to analyze single-cell data and detect anomalies in electrocardiogram (ECG) signals. SUDE exhibits distinct advantage in scalability with respect to data size and embedding dimension, and has promising performance in cluster separation, integrity, and global structure preservation. The experiments also demonstrate notable robustness in embedding quality as the sampling rate decreases.

# Main

Dimension reduction plays a crucial role in machine learning to serve data exploration and visualization [1, 2]. By transforming data into an embedding space, it reveals the intrinsic structures and patterns of data, in turn boosting the efficiency and generalization of machine learning models, and evading curse of dimensionality [3, 4]. As classical dimension reduction methods, principal component analysis (PCA) [5] identifies an ordered sequence of orthogonal vectors (principal components, PCs) such that, for any given $k$, the first $k$ PCs collectively span the $k$-dimensional subspace of largest variance in the data [6]. While multidimensional scaling (MDS) minimizes the differences between pairwise distances in the original and embedding spaces [7]. However, both methods neglect to consider inter-cluster separation and cause a crowding problem [4]. Linear discriminant analysis (LDA) can separate different clusters with the objective of maximizing the between-class variance while minimizing the within-class variance [8]. It requires true class annotations, but the labeled data is rare in practice. Moreover, when dealing with the data lying on nonlinear manifolds where clusters are non-separable by linear boundaries, LDA fails to achieve satisfactory performance in cluster separation using a linear projection paradigm [9, 10].

Manifold learning surpasses classical dimension reduction methods due to its ability to handle nonlinear high-dimensional data. Its concept was first derived from isometric feature mapping (Isomap) [11] and locally linear embedding (LLE) [12]. Isomap exploits the shortest path distances measured in the K-nearest neighbors (KNN) graph [13], while LLE recovers manifold structure based on the locally linear relationships of data. Subsequently, a series of related methods, e.g., Laplacian eigenmaps [14], Hessian eigenmaps [15], and diffusion maps [16] were proposed successively. These methods focus on the preservation of the manifold structure, but ignore the discrimination of clusters, thereby allowing clusters to overlap in lower-dimensional space. To better preserve between-cluster discrimination while specifically optimizing for low-dimensional visualization, t-distributed stochastic neighbor embedding (t-SNE) [17] innovatively approximates high-dimensional Gaussian distributions that model point neighborhoods in the original space with heavy-tailed t-distributions in the embedding space. By adopting such two different distributions, t-SNE maintains local neighborhoods and creates stronger repulsion between dissimilar points to avoid crowding. Building upon t-SNE's framework, uniform manifold approximation and projection (UMAP) has also achieved success across many fields [18, 19]. It assumes that the local neighborhoods lie on a Riemannian manifold, and uses Laplacian eigenmaps to obtain an initial embedding with a promising global structure. UMAP employs cross entropy as the cost function and performs stochastic gradient descent to optimize it for fast convergence. It can achieve more efficient computation compared to t-SNE, but is insufficient to preserve the global structure only through Laplacian eigenmaps initialization, and still suffers from computational scalability problem for data with large sizes and high dimensions [20].

The high time consumption of t-SNE-like methods primarily arises from the computation of a large matrix over hundreds of epochs during the learning stage [21]. Selecting a subset of key points as landmarks to participate in the embedding learning process can significantly improve time efficiency and scalability [22]. Random sampling from data identifiers is a straightforward approach that enables fast landmark selection [23, 24]. However, outcomes tend to be unstable, especially when processing data with large heterogeneity in densities and cluster sizes. Considering that points with higher density should be given priority in selection, a tree-based landmark sampling algorithm, pruned spanning tree (PST), was introduced to expedite diffusion maps [25]. Nonetheless, the high time complexity of PST in generating a random spanning tree is against the purpose of speed-up. To capture the entire data structure, maximum-minimum sampling (MMS) was designed that recursively adds points into the landmark set by maximizing the minimum path-based distance to the current set [26]. Although it can establish a uniform landmark set, MMS cannot ensure the same sampling rate for each cluster and leads to a higher sampling rate skewed toward the sparse clusters. Therefore, for producing a stable and faithful embedding, a scalable manifold learning method equipped with a robust and uniform sampling strategy that is capable of overcoming data heterogeneity is highly desired.

In this work, we develop a scalable manifold learning method SUDE by uniform landmark sampling and constrained locally linear embedding (CLLE). SUDE samples a subset of points as landmarks for participation in the embedding learning process, and then incorporates the non-landmarks into the learned low-dimensional space. This method has significant advantages in scalability and robustness over the mainstream baselines, as shown in the experimental results on a total of 27 datasets derived from different fields. Moreover, it can achieve a discriminative layout of clusters, while preserving cluster integrity and long-distance relationships. Our contributions can be summarized into three facets. Firstly, we propose a landmark sampling strategy called plum pudding sampling (PPS) that enables stable and uniform landmark sampling through neighborhood exclusion. Secondly, we develop CLLE to incorporate the non-landmarks into the learned space for discriminative embedding. Thirdly, we redesign the framework of embedding learning to further enhance sampling robustness and computational efficiency, including early aggregation, logarithmic low-dimensional probability, and momentum-based gradient descent.

# Results

## Overview of SUDE

The conceptual framework of SUDE, depicted in Fig. 1a, outlines its workflow, along with the missions and applications explored in this study. Specifically, SUDE consists of three major stages, sampling, learning, and incorporating. In the first stage, the PPS algorithm is performed to choose uniform landmarks from the original data. These landmarks are then fed into the learning stage to obtain the landmark embedding by an optimized gradient descent framework for generating the global skeleton of data. Subsequently, CLLE is employed to incorporate the non-landmarks into the learned lower-dimensional space through the local linear relationships with neighboring landmarks. In summary, the SUDE method constructs the global skeleton of data through landmark embedding learning, and then recovers the local neighborhood structure of each non-landmark at the third stage.

The essential of landmark sampling in manifold learning lies in how to preserve the distribution of the original data, e.g., relative positions of clusters and the proportion of points in each cluster. We expect that landmarks are uniformly distributed throughout the data (Supplementary Note 6), and locally high-density points should have high priority to be sampled (Supplementary Note 7) since they contain more important information of data structure. Based on the idea of neighborhood exclusion, we propose the plum pudding sampling (PPS) algorithm. It gets the name from the fact that the landmarks obtained by PPS have an approximately uniform distribution, resembling "plums" embedded in the original data "pudding". PPS arranges all points in descending order based on the number of reverse nearest neighbors (RNN) in advance, where RNN is a local density measurement that finds points whose KNN include the given query point [27, 28]. The points with high RNN values (i.e., hubs) enables preservation of graph structure and centrality properties [29]. Then, PPS recursively selects the first point in the sorted point queue as a landmark and moves its KNN from the queue to the non-landmark set. This operation continues until there are no points left in the queue. The sampling rate is determined by the KNN parameter $k_1$, with a larger $k_1$ leading to a lower sampling rate. The workflow is illustrated in Fig. 1b and has been summarized as Algorithm 1 in Supplementary Note 1.

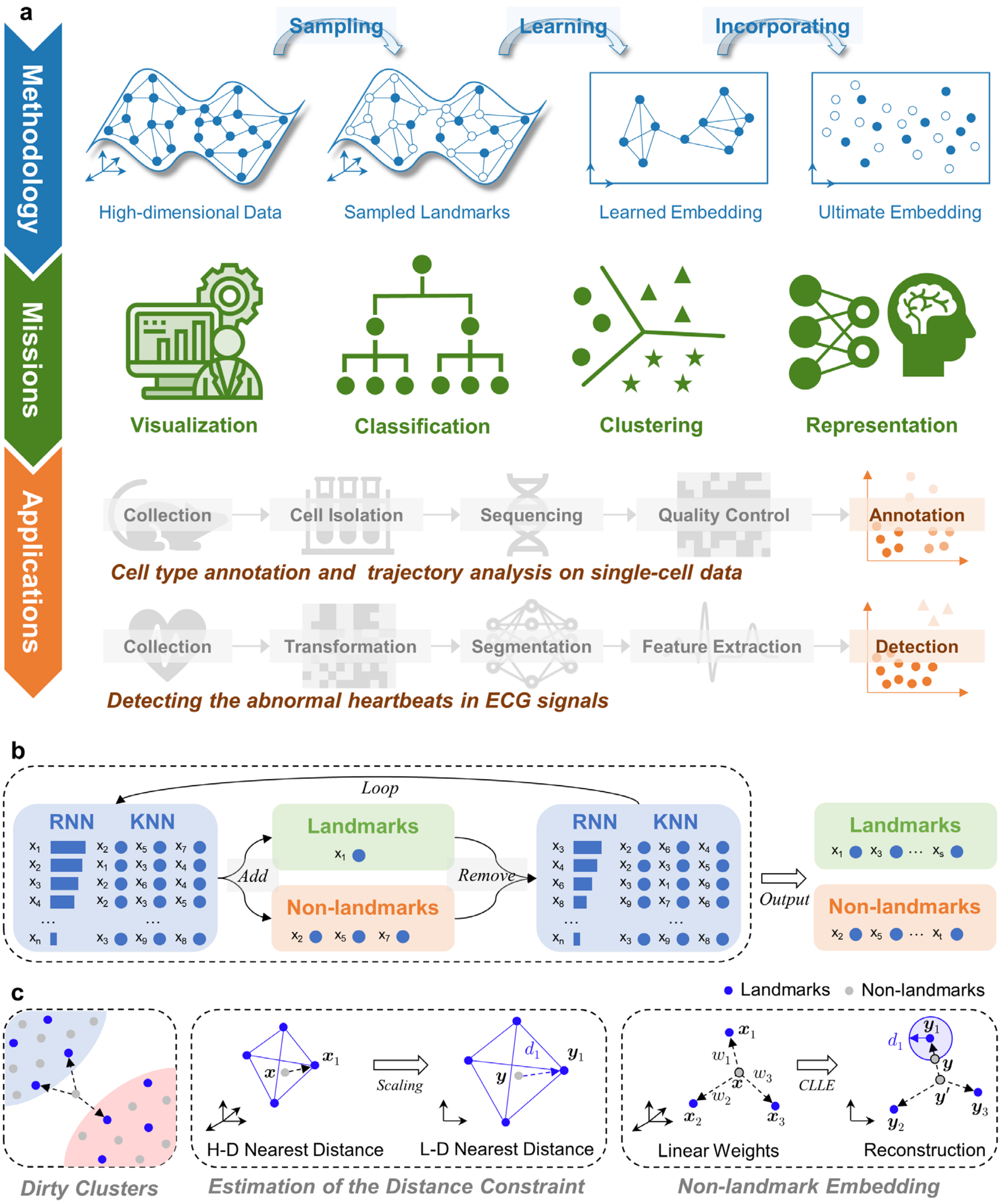

**Fig. 1. Illustration diagram of SUDE.** (**a**), The overall conceptual framework of SUDE includes a three-stage workflow and delineates the missions and application cases in this study. The workflow consists of three main stages: (*i*) sampling landmarks from the original high-dimensional data via PPS; (*ii*) embedding learning to generate the lower-dimensional embedding of landmarks; (*iii*) obtaining the ultimate layout by incorporating non-landmarks to the lower-dimensional space using CLLE. (**b**), The workflow of PPS. All points are arranged in descending order according to the number of RNN. In each loop, the first point in the queue is added to the landmark set, and its KNN are moved from the queue to the non-landmark set at the same time. We repeat this operation until there are no points left in the queue. (**c**), Illustration of the dirty clusters and two key components of CLLE. CLLE imposes a constraint of the nearest distance on the objective of LLE to eliminate dirty clusters. It estimates the low-dimensional (L-D) distance constraint by scaling the high-dimensional (H-D) distance between the query non-landmark

and its nearest landmark. The non-landmark embedding can be determined by integrating local linear representations and distance relationships.

In the learning stage, we approximate a high-dimensional probability distribution $\boldsymbol{P}$ over pairs of points with a low-dimensional probability distribution $\boldsymbol{Q}$, and optimize its core components to adapt to the PPS sampling strategy and improve the efficiency of gradient descent. Undersampling would generate sparse landmarks and make them far apart from each other. The intra-cluster compactness will decrease, causing closely located clusters to lack significant distinguishability (Supplementary Note 9). To tackle the undersampling problem, we early aggregate the landmarks based on their associations in the original data (Supplementary Note 10), where these pairwise associations can be measured by the shared nearest neighbor (SNN) method (equation (2) in Methods). We then adopt a Gaussian kernel to compute the high-dimensional conditional probabilities $\boldsymbol{P}$ between landmarks (equation (3)). After generating an initial layout via Laplacian eigenmaps, we compute the low-dimensional probabilities $\boldsymbol{Q}$ using a logarithmic similarity kernel (equation (6)). The logarithmic distribution has a heavier tail than t-distribution, and can achieve stronger intra-cluster compactness within less training epochs (Supplementary Note 8). Kullback-Leibler divergence (KLD) is leveraged as the cost function and can be minimized by the gradient descent method (equation (8)). To speed up the optimization of gradient descent, we exploit a momentum method with an adaptive learning rate and momentum term (equation (9)). The learning rate is determined through a combination of a constant warm-up strategy and a cosine annealing schedule (equation (19)). This approach serves to expedite training and prevent trapping in local optima by periodically tuning the learning rate from an initial higher warm-up value to a minimum value rapidly [30]. The differences between t-SNE, UMAP, and SUDE can be seen in Supplementary Table 1, and more details about preprocessing, optimization, and hyperparameter settings are in Methods.

The linear relationships with the k-nearest landmarks can be used to determine the proper position of the non-landmarks in the learned space. However, owing to the cluster proximity, these landmarks might belong to different clusters, especially when the non-landmarks are near the fuzzy cluster boundaries. In this case, these non-landmarks tend to be inserted into the between-cluster gaps. It makes them out of clusters and results in the occurrence of outliers or dirty clusters, which represent the clusters without distinct boundaries as illustrated in Fig. 1c. We hence develop CLLE by imposing a constraint of the nearest distance on the naïve objective of LLE, which can effectively confine the non-landmarks within a small neighborhood of their nearest landmarks. Its core idea lies in fusing local linear representations and distance relationships between landmarks and non-landmarks for a well-separated layout. CLLE first reconstructs an initial position using the linear weights obtained by LLE, then estimates the distance between the query non-landmark and its nearest landmark in the learned space by scaling their corresponding distance in high-dimensional space (equation (17)). The embedding of the query landmark can be finally determined by solving the objective in equation (10). More details of CLLE can be found in Methods, Algorithm 6 in Supplementary Note 1, and Supplementary Note 11.

## Validating the sampling robustness on synthetic datasets

We assessed the sampling advantage of the PPS algorithm by comparing with three landmark sampling baselines, i.e., random sampling (RS), pruned spanning tree (PST), and maximum-minimum sampling (MMS), on two 2-D synthetic datasets, DS1 [31] and DS2 [32], with clusters of different shapes and densities. Since DS1 and DS2 contain different numbers of points, we set $k_1$ for PPS from 1 to 20 on DS1 and 30 to 50 on DS2 for a similar range of sampling rate. In RS, sampling was carried out randomly upon the data identifiers. PST randomly generates a spanning tree and prioritizes selecting points with high vertex degrees, while MMS recursively adds points into the landmark set by maximizing the minimum path-based distance to the current set. The detailed description of PST and MMS can be found in Supplementary Fig. 16. For a fair comparison, these four methods were required to output the same number of landmarks. To evaluate the distribution consistency, we used the offset distance of centroids (ODOC) to detect the change in cluster positions, which calculates the total offset distances of all cluster centroids before and after sampling. While we measured the congruence of cluster size (CCS) between the sampled landmarks and the original data by the cosine similarity of the ratio of points contained in each cluster. As shown in Fig. 2a, the landmarks produced by RS and PST are randomly distributed and many large blank regions exist, which introduces inescapable errors in preserving the topological structure of these regions. The inherent randomness also makes the outcomes unstable. In comparison, MMS and PPS generate more uniform landmarks and obtain much smaller ODOC scores (Fig. 2b), thereby beneficial for depicting the global structure of the data in a stable manner. However, MMS yields lower CCS scores in Fig. 2c, which causes the size differences of clusters in the original data to be lost in the embedding. Meanwhile, we compared the computational efficiency of PST, MMS, and PPS using a simulated 2-D dataset in Fig. 2d. PST requires a significant amount of time to build a spanning tree, while MMS takes a lot of time to compute the shortest path distances. Compared with them, PPS has a better scaling performance with increasing data size and runs in just 0.6 seconds on 50k points.

To evaluate the impact of landmark sampling on manifold learning, we designed a 5-D synthetic dataset DS3 containing three same-sized circles that are perpendicular to each other in Fig. 2e. This structure forms a 1-D varifold, which is a generalization of manifolds that permits singularities like intersection points while preserving manifold properties elsewhere. We applied SUDE equipped with four sampling algorithms to DS3. The congruence coefficient (CC) [33] was utilized to quantify global structure preservation as it can capture the long-distance dependency between clusters. CC calculates the cosine similarity of the geodesic distance between high-dimensional data and low-dimensional embedding. As shown in Fig. 2f, the CC scores of SUDE jointly with RS and PST fluctuate markedly and decrease rapidly as $k_1$ increases (the sampling rate decreases). By contrast, the CC scores obtained by PPS exhibit a high and stable level, which remains as high as 0.987 even at $k_1 = 50$ (the sampling rate is only 3.67%). We also compared the 2-D embeddings of t-SNE, UMAP, and SUDE in Fig. 2g. SUDE better restores the circle structure (CC=0.989 at $k_1 = 10$) than t-SNE (CC=0.937) and UMAP (CC=0.942), and preserves its symmetry and coherence.

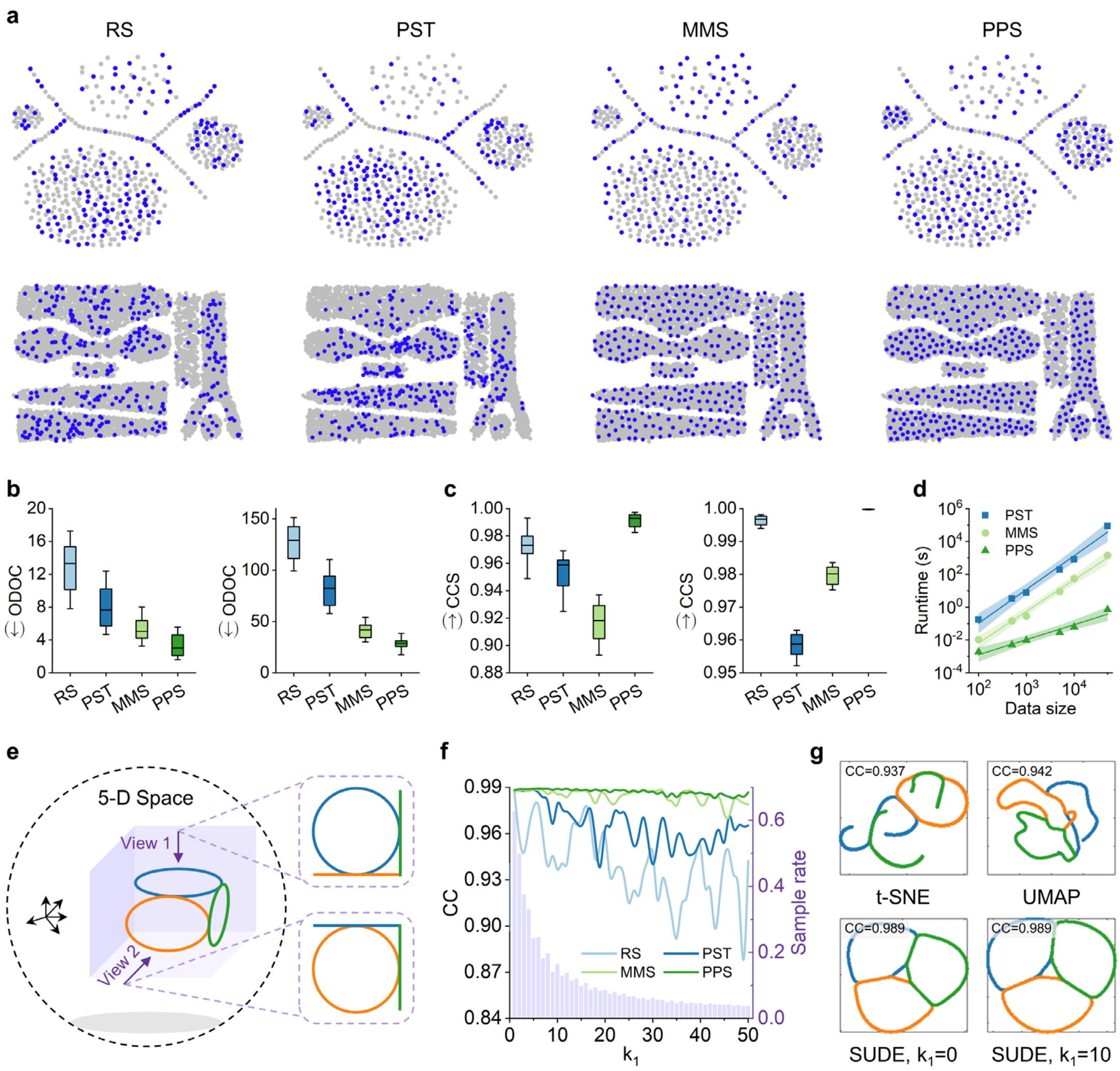

**Fig. 2. Performance of the PPS algorithm on synthetic datasets.** (**a**), Landmarks generated by RS, PST, MMS, and PPS from two 2-D synthetic datasets (DS1 and DS2) with 524 and 7,677 points, respectively. (**b**), ODOC and (**c**), CCS scores obtained by the four sampling algorithms. The boxes show the median and 5%-95% score range. (**d**), Runtimes of PST, MMS and PPS on various sized sub-samples of a simulated 2-D dataset with 50k points. (**e**), A 5-D synthetic dataset DS3 contains three same-sized mutually orthogonal circles, where each circle is a 1-D manifold and composes of 500 points. (**f**), Trends of the CC score obtained by SUDE jointly with four sampling algorithms, and sampling rates of PPS under different $k_1$. (**g**), 2-D embeddings of t-SNE, UMAP, and SUDE with $k_1 = 0$ (*without applying sampling procedure*) and $k_1 = 10$ on DS3.

## Practical efficacy on real-world datasets

We examined the practical efficacy of SUDE by comparing with five typical baselines, Barnes-Hut-SNE (BH-t-SNE), UMAP, TriMap [20], TopoAE [34], and P-UMAP [35], on 13 real-world datasets, including eight UC Irvine benchmarks (Wine, Dermatology, Breast-Cancer, Mfeat, Rice, Spambase, Dry-Bean, and Shuttle) [36], three image datasets (CIFAR10 [37], MNIST [38], and FMNIST [39]), and two text datasets of news articles (AG's News, and Yahoo) [40]. We specified the default parameters suggested by BH-t-SNE, UMAP, and TriMap, and built network architectures consisting of a three-layer encoder, a 2-neuron bottleneck layer, and a symmetric decoder for TopoAE and P-UMAP. More details about the datasets and parameter settings can be found in Supplementary Tables 2 and 3, respectively. For SUDE, we set $k_1 = 20$ for the eight UC Irvine datasets, and $k_1 = 50$ for CIFAR10, MNIST, FMNIST, and AG's News by considering the data size. This setting makes the sampling rate lie in a unified range of 10%~20% for all datasets. We implemented SUDE, BH-t-SNE, and UMAP in MATLAB R2022a on a commodity desktop computer with an 8-core Intel i7 processor and 256 GB RAM, and utilized the cloud version of TriMap at OMIQ platform. TopoAE was executed on a server with Intel Xeon Platinum 8255C CPU running at 2.50 GHz and RTX 2080 Ti, while P-UMAP was performed on a server with AMD EPYC 9654 96-core processor.

The 2-D embeddings from the seven small datasets are shown in Supplementary Fig. 17, while that from Shuttle, CIFAR10, MNIST, FMNIST, and AG's News are illustrated in Fig. 3. In general, SUDE yields better cluster integrity and discrimination, and more accurate global structure. For Shuttle, BH-t-SNE and UMAP split the clusters of "High" (red) and "Bypass" (purple) into several parts, and these parts are blocked by the cluster of "Rad Flow" (blue). While the integrity of "High" (red) and "Bypass" (purple) clusters are better preserved in the TopoAE, P-UMAP, and SUDE embeddings, as the points in these two clusters are distributed closely, although they are split into multiple subclusters. In terms of CIFAR10, the ten clusters are densely mixed together without clear boundaries in the TopoAE embedding. While SUDE pulls apart each cluster, generates less overlaying, and keeps the relative global relationship. For instance, due to the similarity of image features, "truck" (cyan) and "automobile" (orange) are more similar compared to "ship" (olive) in the SUDE embedding (also in UMAP and TriMap). However, this relationship is damaged in the layout of BH-t-SNE, where "truck" is closer to "ship" than to "automobile". BH-t-SNE also fails to preserve the global structure of MNIST, and breaks up the cluster of digit "1". This may also be related to the implementation of BH-t-SNE. Modern implementations like openTSNE [41] might improve the results due to smarter initialization and more appropriate hyperparameter settings. Nonetheless, we still the MATLAB's built-in function for comparison in this paper, as it remains the widely adopted version in current practice. TriMap generates an overlarge layout of digit "1". Such heterogeneity in cluster size hinders the exploration of relationships of small-sized clusters and is not conducive to unsupervised cluster partitioning. TopoAE cannot separate these clusters and suffers from a severe crowding problem. SUDE presents a more uniform and well-separated visualization. It clearly separates digit "8" from the clump composed of digits "3" and "5", while also isolating digit "7" from the clump of digits "4" and "9". The fashion items in FMNIST are quite similar, resulting in most categories being mixed and indistinguishable. Both P-UMAP and SUDE push the categories of "Trouser" (orange) and "Bag" (olive) away from other categories, while SUDE further generates a distinct gap between two clumps, one containing classes of "T-shirt" (blue) and "Dress" (red), and the other having categories of "Pullover" (green), "Coat" (purple) and "Shirt" (pink). In AG's News, BH-t-SNE completely damages the global structure, while SUDE generates a more distinguishable layout, and separates the clusters of "Business" (green) and "Science/Technique" (red) that are adjoined in the layouts of UMAP, TriMap, TopoAE, and P-UMAP.

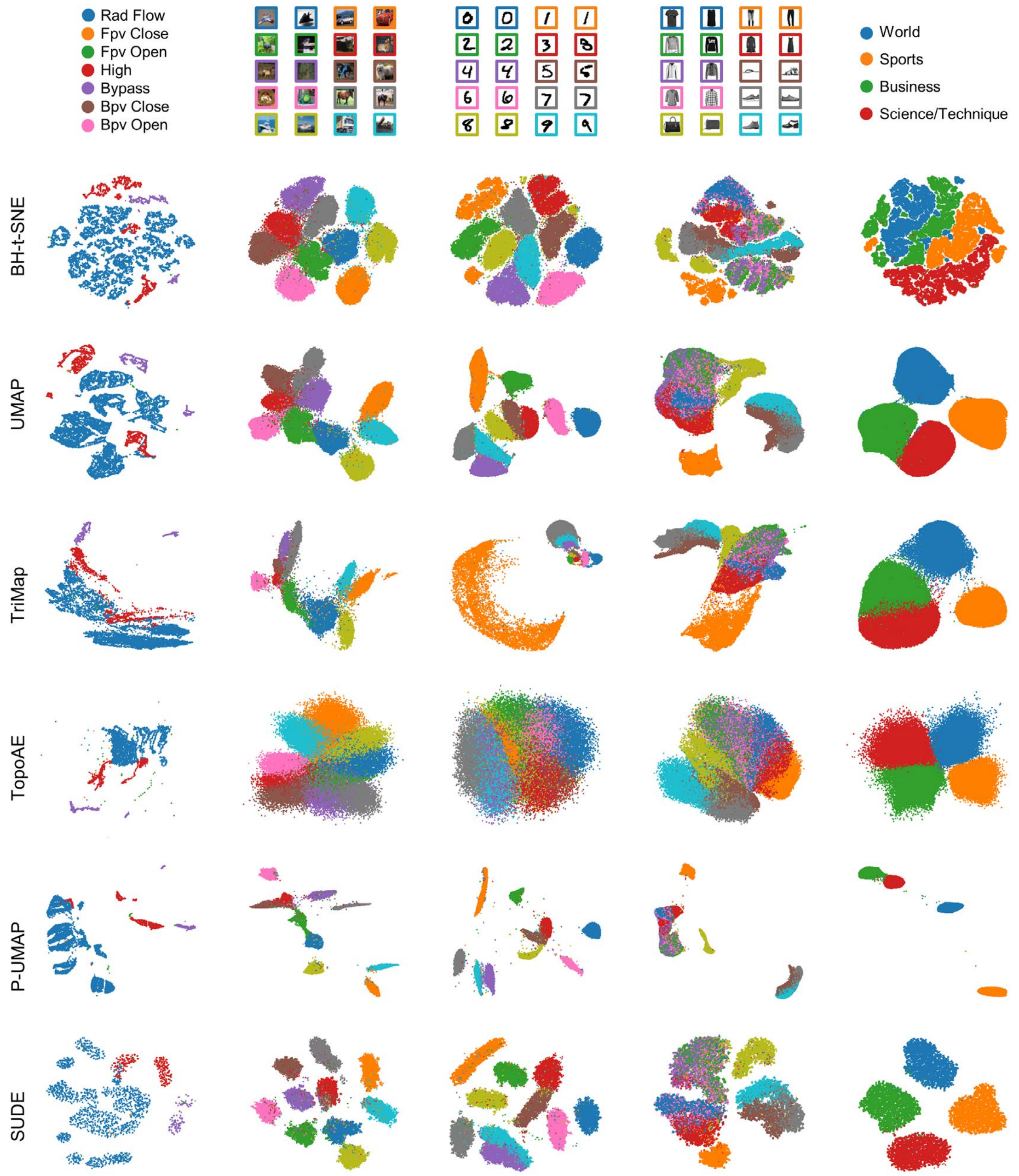

**Fig. 3. 2-D embeddings of BH-t-SNE, UMAP, TriMap, TopoAE, P-UMAP, and SUDE from five real-world datasets including Shuttle, CIFAR10, MNIST, FMNIST, and AG's News.**

For quantitative assessment of embedding quality, we adopted a multi-faceted evaluation framework, comprising classification accuracies of KNN (knnACC) and SVM (svmACC) classifiers [42], clustering accuracy of K-means (clusACC), and preservation of global and local structures reported by CC score and KNN recall (knnRec), respectively. The results are illustrated in Fig. 4a, and Supplementary Fig. 18 and Table 4. In general, SUDE yields the highest scores in 23 out of 60 benchmark tests, outperforming the competitors. Specifically, SUDE obtains the highest classification and clustering accuracy metrics (knnACC, svmACC, and clusACC) on seven, six, and nine datasets, respectively. For global structure preservation, SUDE achieves an average CC score of 0.921 across all datasets, which is higher than t-SNE (0.901) and UMAP (0.911) but lower than TopoAE (0.950) and P-UMAP (0.931). In terms of local structure preservation, BH-t-SNE demonstrates a distinct advantage and yields the highest knnRec scores on ten datasets. SUDE shows relatively low knnRec scores on the first seven small datasets, while it ranks first on Shuttle and takes the second place across the last four large datasets. This is because each point is subject to a growing number of attractive and repulsive forces as the data size increases, which leads to a slower convergence and more severe loss of neighborhood structures. Although the landmarks in SUDE are also affected by the increased forces during the embedding optimization, CLLE allows non-landmarks can partially preserve their neighborhood structures and mitigates the decline in knnRec score as the data size grows. As the relative runtimes shown in Fig. 4b, SUDE exhibits a significant advantage in time efficiency and rank at the forefront on most datasets, where it runs around 8-fold, 4-fold, 2-fold, 18-fold, and 17-fold faster than BH-t-SNE, UMAP, TriMap, TopoAE, and P-UMAP on AG's News, respectively. Fig. 4c provides a comprehensive comparison in six performance dimensions. The score of each dimension reports the advantage of each method against all competitors, with a higher value indicating a greater advantage. SUDE ranks first in terms of knnACC, svmACC, clusACC, and scalability, with its average ranking in scalability approaching one. BH-t-SNE dominates knnRec score. TopoAE has distinct advantage in global structure preservation by fusing the topological loss and reconstruction loss. While the powerful representational capability of neural networks enables better isometry preservation, but suffers from considerable computational overhead.

For further testifying to the scalability, we analyzed the time complexity in Supplementary Note 2 and benchmarked BH-t-SNE, UMAP, TriMap, TopoAE, P-UMAP, and SUDE on the sub-samples of the full Yahoo dataset with 1.4 million samples [21]. The result in Fig. 4d indicates that SUDE has superior scaling performance with data size, and only costs around 32 minutes on the full samples (6.5-fold, 10.9-fold, and 8.3-fold faster than BH-t-SNE, TopoAE, and P-UMAP, respectively). Moreover, we validated the scaling performance with respect to the embedding dimension on MNIST. Since BH-t-SNE and TriMap only support embedding high-dimensional data into 2- or 3-D space, we selected the exact t-SNE algorithm and UMAP for comparison. The result in Fig. 4e reveals that SUDE scales better with the embedding dimension compared to the competitors, which is 5.4-fold and 1.2-fold more efficient than BH-t-SNE and UMAP in dimensions of 500. To testify the validity of the embedding learning framework, we replaced the landmark embedding component of SUDE with t-SNE and UMAP while keeping all other components and parameter settings identical. Fig. 4f shows the distributions of clusACC scores for the three methods on MNIST as $k_1$ varies from 10 to 50. The landmark embedding component of SUDE exhibits higher and more stable performance. We also compared the logarithmic similarity kernel function $q_{sude}$ with that of t-SNE and UMAP, $q_{tsne}$ and $q_{umap}$, by integrating them with other components of SUDE under the same configurations. The clusACC scores shown in Fig. 4g indicate that $q_{sude}$ can enhance the convergence speed of the embedding learning compared to $q_{tsne}$ and $q_{umap}$. More detailed about the formulas and 2-D embeddings are illustrated in Supplementary Figs. 20 and 21. We further conducted ablation experiments using five evaluation metrics on MNIST to validate the effectiveness of key components, including the RNN sequence in PPS, initialization, early aggregation, and the constraint in CLLE (Supplementary Figs. 19 and 22-24).

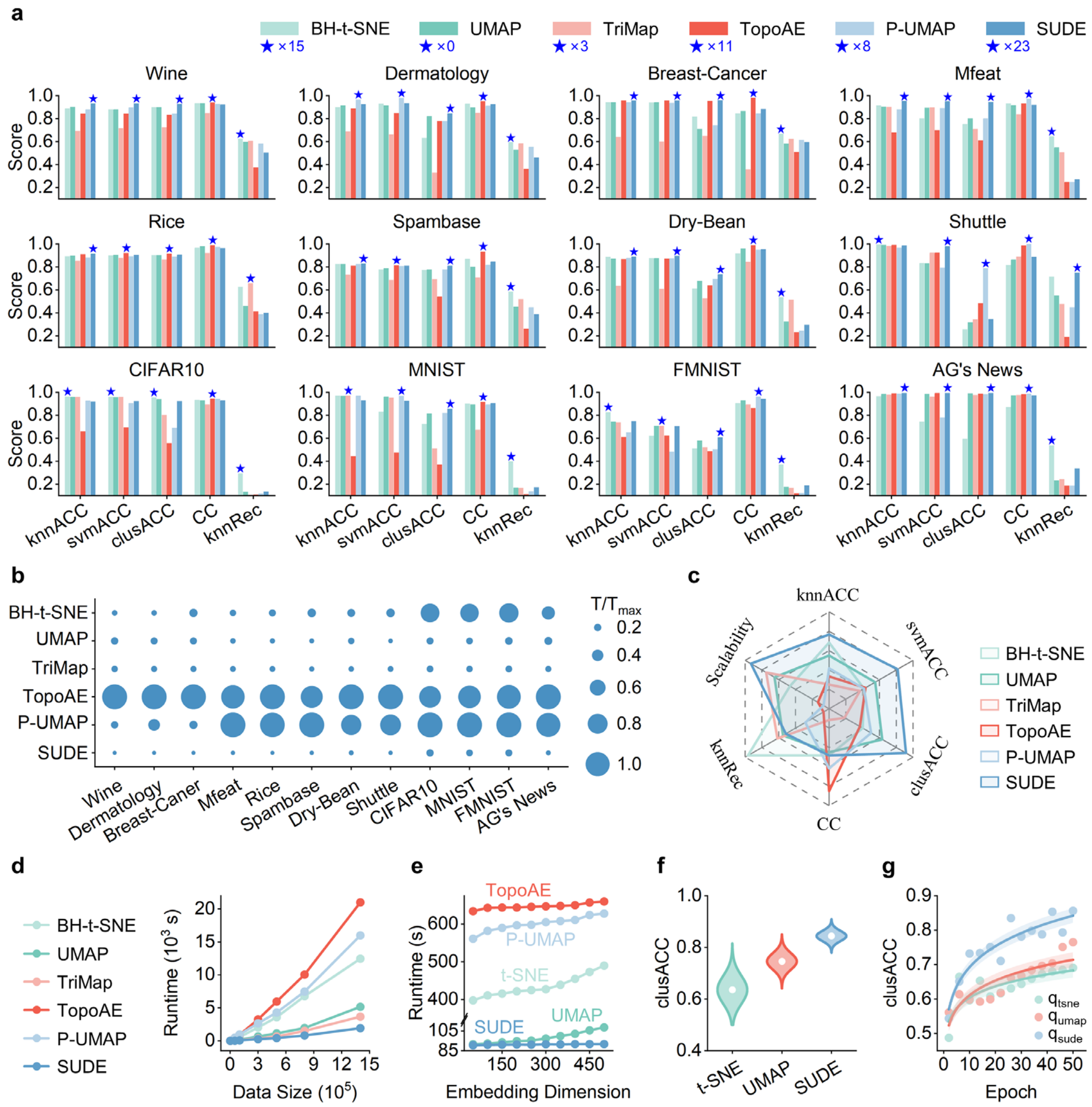

**Fig. 4. Quantitative comparison of the embedding quality and time efficiency on 13 real-world datasets.** (**a**), Six evaluation metrics of BH-t-SNE, UMAP, TriMap, TopoAE, P-UMAP, and SUDE on eight UC Irvine benchmarks, three image datasets, and a text dataset of news articles, where the blue star indicates the highest score. (**b**), The relative runtimes of these six methods, where the relative runtime denotes the absolute runtime divided by the maximum runtime on each dataset. (**c**), Comparison of these six manifold learning methods in six performance dimensions, where the score of each dimension is equal to seven minus the average ranking of the corresponding evaluation metric across all datasets, ranging from 1 to 6. (**d**), Runtimes scaling on various sized sub-samples of the full Yahoo dataset, where $k_1$ of SUDE is specified as $\frac{n}{1000}$ with $n$ denoting the total number of points. (**e**), Runtimes of t-SNE, UMAP, TopoAE, P-UMAP, and SUDE under different embedding dimensions on MNIST. (**f**), A violin plot shows the clusACC scores of SUDE against variants where the landmark embedding component was replaced with t-SNE and UMAP, under $k_1$ from 10 to 50 on MNIST. The white dots represent the mean values. (**g**), clusACC scores obtained by SUDE equipped with three low-dimensional similarity kernels under different epochs on MNIST, with shading denoting a 95% confidence band.

## Application on single-cell data

In life science, dimension reduction enables revealing biological data structure with similar phenotypes, annotating cell populations, and delineating the complex development processes, from the single-cell RNA sequencing (scRNA-seq) and mass cytometry (CyTOF) data [43, 44]. To evaluate the applicability, we performed our method on a published scRNA-seq datasets from wild-type (WT) mouse retinas [45], which cover 7,695 cells with 27,998 genes. Seurat software [46] was leveraged to conduct the standard preprocessing pipeline for scRNA-seq data, including quality control, normalization, selecting the top 2,000 highly variable genes (HVG), scaling, and PCA embedding [47]. Fig. 5a present the 2-D embeddings of UMAP and SUDE generated from the 50-D PCA results. Both methods can generate layouts with separable clusters. However, UMAP tends to over-cluster the populations of cone bipolar cells and amacrine cells, while SUDE better preserves the cluster integrity. Further discussion including differentially gene expression analysis and over-clustering significance test can be seen in Supplementary Note 12. Identification of distinct cell types in transcriptome data is important and often non-trivial. To validate the contribution to unsupervised cell type identification, we performed a cutting-edge clustering algorithm, clustering by direction centrality (CDC) [48], on the 2-D embeddings by considering the heterogeneity in cluster size and density. The parameter $ratio$ of CDC was varied from 0.80 to 0.99, and the corresponding accuracies by comparing with the ground truth (GT) annotations of the retinal cells are presented in Fig. 5b. SUDE leads UMAP by a large margin and the advantage grows as $ratio$ increases. The Sankey plot in Fig. 5c illustrate the best CDC results and indicate that the predicted results based on the SUDE embedding stay more closely aligned with the true cell type annotations than UMAP embedding. We further verified the validity of SUDE on two larger scRNA-seq datasets from human retinal pigment epithelium and choroid (39,326 cells and 36,601 genes) [49] and nasopharyngeal epithelium (55,319 cells and 32,994 genes) [50] in Supplementary Fig. 26.

We also investigated the performance of SUDE on a CyTOF dataset Levine [51], containing 265,627 bone marrow cells with 32 surface markers and 14 manually gated populations. We followed the standard preprocessing steps in Weber et al. 2016 [52] including the application of an arcsin transformation with a standard cofactor of five. BH-t-SNE, UMAP, and SUDE were performed on the entire dataset and the embeddings of the 104,184 manually gated cells are illustrated in Fig. 5d and Supplementary Fig. 27a. SUDE successfully grouped cells of the same population together and cells of different subsets separately. Compared to BH-t-SNE, it more effectively maintains the connectivity of four types of B cells, plasma B, mature B, pre-B, and pro-B, and avoids the misplacement of $CD16^+$ NK cells within the monocyte cluster. SUDE embedding is broadly close to the result of UMAP. By referring to the lineage tree of these 14 human bone marrow cell types in Supplementary Fig. 28, we can observe that UMAP and SUDE can better preserve the phenotypic similarity of HSCs/HSPCs, myeloid cells, and lymphoid cells including NK cells, T cells, and B cells than BH-t-SNE. Their embeddings present the development landscape and transition states from HSPCs to myeloid cells and B cells [53]. We calculated the nearest cell type for each population based on their centroid positions in the 2-D embeddings. As shown in Fig. 5g, SUDE correctly preserves the nearest cell types for 11 populations, achieving the highest accuracy of 78.57% among the three methods. Meanwhile, it can be observed that basophils and pDCs, both of which belong to the myeloid lineage, had pDCs as the nearest type to basophils in the original data. Although UMAP clearly separates basophils and pDCs into two distinct clusters, the neighborhood topology of basophils is disrupted, where their nearest types shifting to $CD34^+CD38^+CD123^-$ HSPCs instead of pDCs or their progenitor $CD34^+CD38^+CD123^+$ HSPCs. Although SUDE does not distinctly separate basophils and pDCs, it preserves the fact that pDCs is the closest cell type to basophils. In addition, microscopic patterns can be discovered in the SUDE embedding. For example, a clear gap can be observed between the two subsets of monocytes, CD11b$^+$ and CD11b$^-$ monocytes. By estimating the cell density using Gaussian mixture models, we can find two density peaks in the SUDE embedding that is consistent with the two subpopulations of monocytes (Fig. 5e). The comparison among BH-t-SNE, UMAP, and SUDE in CC score and scalability score is presented in Fig. 5f. SUDE performs better in global structure preservation and keeps consistently high CC scores even as the sampling rate decreases (illustrated by the error bar). SUDE also exhibits remarkable scalability advantages with being 6.3-fold and 3.0-fold more efficient than t-SNE and UMAP when $k_1 = 3{,}000$, respectively. We further performed the CDC and K-means algorithms on their 2-D embeddings. The results in Supplementary Fig. 29 demonstrate that SUDE embedding is more competent for unsupervised cell type annotation tasks.

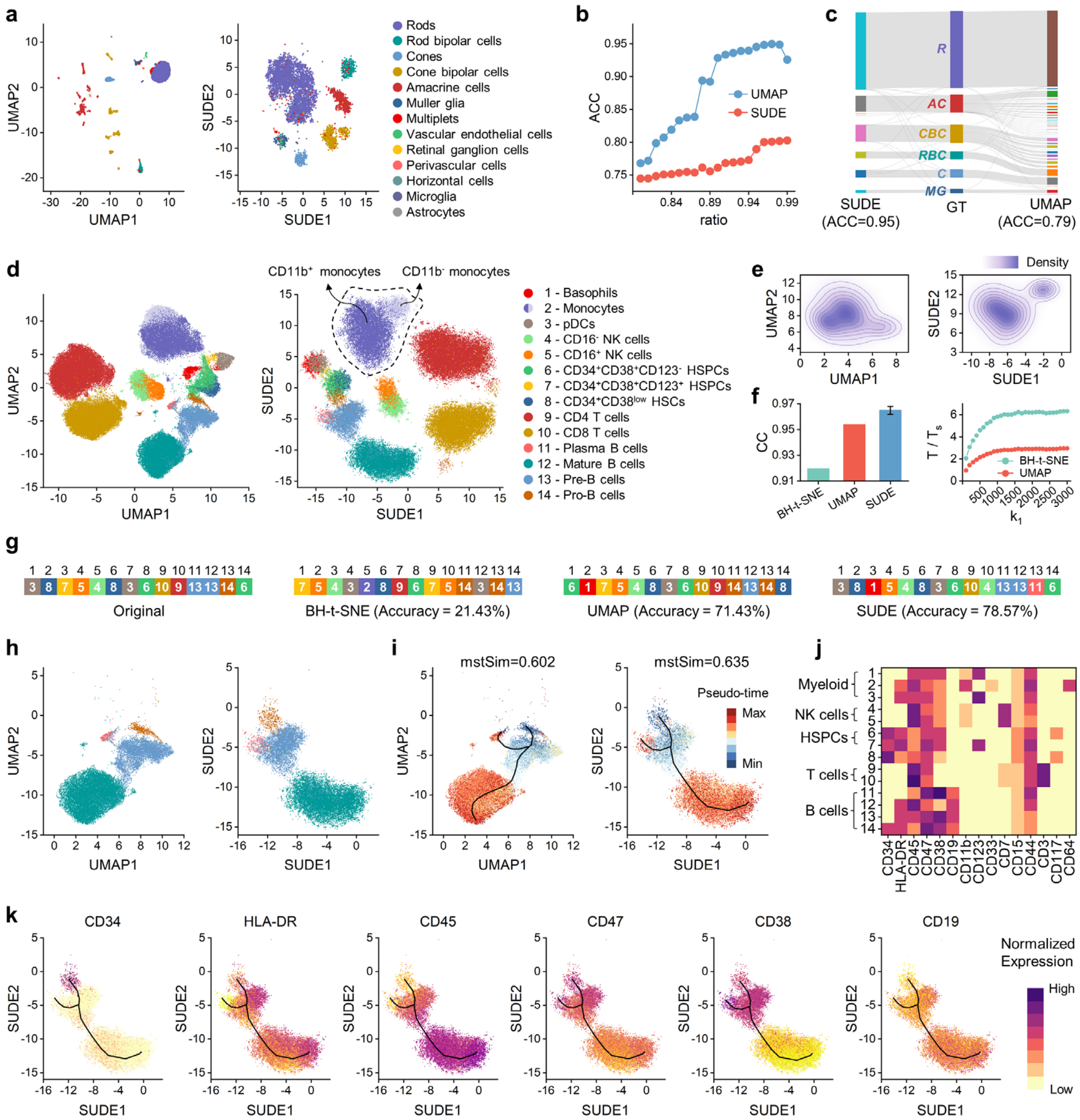

**Fig. 5. Performance comparison on scRNA-seq and CyTOF data.** (**a**), 2-D embeddings of UMAP and SUDE ($k_1 = 20$) on the WT scRNA-seq dataset. (**b**), ACC scores of the CDC clustering algorithm by varying $ratio$ from 0.80 to 0.99 on the UMAP and SUDE embeddings. (**c**), A Sankey diagram shows the match between the best CDC clustering results on the two embeddings and the ground truth (GT) annotations of the retinal cell types (R: rods, RBC: rod bipolar cells, C: cones, CBC: cone bipolar cells,

AC: amacrine cells, and MG: Muller glia). (**d**), 2-D embeddings generated by UMAP and SUDE on the Levine CyTOF dataset. (**e**), Contour plots illustrate the cell density estimated of monocytes by Gaussian mixture models. (**f**), CC and $T/T_s$ scores of BH-t-SNE, UMAP, and SUDE, where $k_1$ is changed from 100 to 3,000 with an interval of 100, and the $T/T_s$ score is defined as the runtime divided by the SUDE's runtime. (**g**), The color bars show the nearest cell type for each cell population by calculating the distances between 14 cluster centroids. The digits in the first row represent the cell type IDs, and the colored blocks indicate their corresponding nearest cell types. (**h**), Four types of B cells (plasma B, mature B, pre-B, and pro-B) on UMAP and SUDE plots. (**i**), Pseudo-time and lineage trajectories of B cell differentiation inferred by Slingshot in the UMAP and SUDE plots. (**j**), A heatmap presents the average expression for each cell type on 15 characteristic markers. (**k**), Six highly expressed surface markers (CD34, HLA-DR, CD45, CD47, CD38, and CD19) in B cells during the differentiation process.

Furthermore, we leveraged SUDE embedding to delineate the differentiation landscape of B cells. Fig. 5h displays the UMAP and SUDE maps of B cells. In the UMAP embedding, some cells on the top are isolated and distant from the clusters, whereas the SUDE map presents a cleaner distribution. We further used Slingshot [54] to infer the pseudo-time and reconstruct the lineage trajectory on the original high-dimensional data. The inferred trajectory consists of two lineages as shown in Fig. 5i, including pro-B – pre-B – mature B and pro-B – pre-B – plasma B. To validate the consistency between the embedding and the inferred trajectory, we built a minimum spanning tree upon the 2-D embeddings to simulate the cell differentiation trajectory like the Monocle method [55], and then employed the mstSim score to measure the similarity between the inferred pseudo-time and path-based time. The detailed calculation process can be seen in Supplementary Fig. 27b. SUDE achieves higher mstSim of 0.635 in Supplementary Fig. 27c. We also explored the patterns of six highly expressed surface markers in B cells, CD34, HLA-DR, CD45, CD47, CD38, and CD19 (Fig. 5j) to reveal the cellular states at different development stages (Fig. 5k). CD34, HLA-DR, and CD45 unravel the potential order of differentiation from pro-B to pre-B and then to mature B cells; while CD47, CD38, and CD19 delineate the second lineage from pro-B to pre-B and then to plasma B cells. It illustrates that SUDE can be applied to analyzing the continuous transition states of B cells. Besides, SUDE is robust to the marker selection. We excluded a single canonical marker of NK cells, HSPCs, T cells, and B cells (CD7, CD34, CD3, and CD19, respectively). However, cell populations can still be identified in the SUDE maps, and their global structures are highly consistent with that of the map using all markers (Supplementary Fig. 30).

## Application for detecting the abnormal heartbeats in ECG signals

The electrocardiogram (ECG) is a non-invasive diagnostic technique used to assess the electrical activities of the heart [56]. It records the electrical signals during each heartbeat, providing valuable information about cardiac function, rhythm, and conduction system [57]. Cardiac anomalies can be reflected in the abnormal waveforms of an ECG signal. Detecting the anomalies contributes to the prompt diagnosis and management of cardiac diseases, ultimately improving patient outcomes [58]. To automatically detect abnormal heartbeats in ECG signals, we design an end-to-end workflow in Fig. 6a. Specifically, the Fourier synchrosqueezed transform (FSST) [59] is used to obtain time-frequency representations of the sample points in the signal. These representations are then fed into a pre-trained deep recurrent neural network to identify the P/QRS/T waves (Fig. 6b). Subsequently, we split the entire signal into beat-wise segments using the beginning points of the QRS complex as breakpoints. Each ECG beat corresponds to a cardiac cycle, and comprises of a P wave, a QRS complex, and a T wave. From each beat segment, we extract eight waveform features, including four amplitude and four interval features (Fig. 6c). We apply three dimension reduction techniques, PCA, UMAP, and SUDE, to embed the features into a lower-dimensional space, and exploit KNN and SVM classifiers to distinguish the normal and abnormal heartbeats.

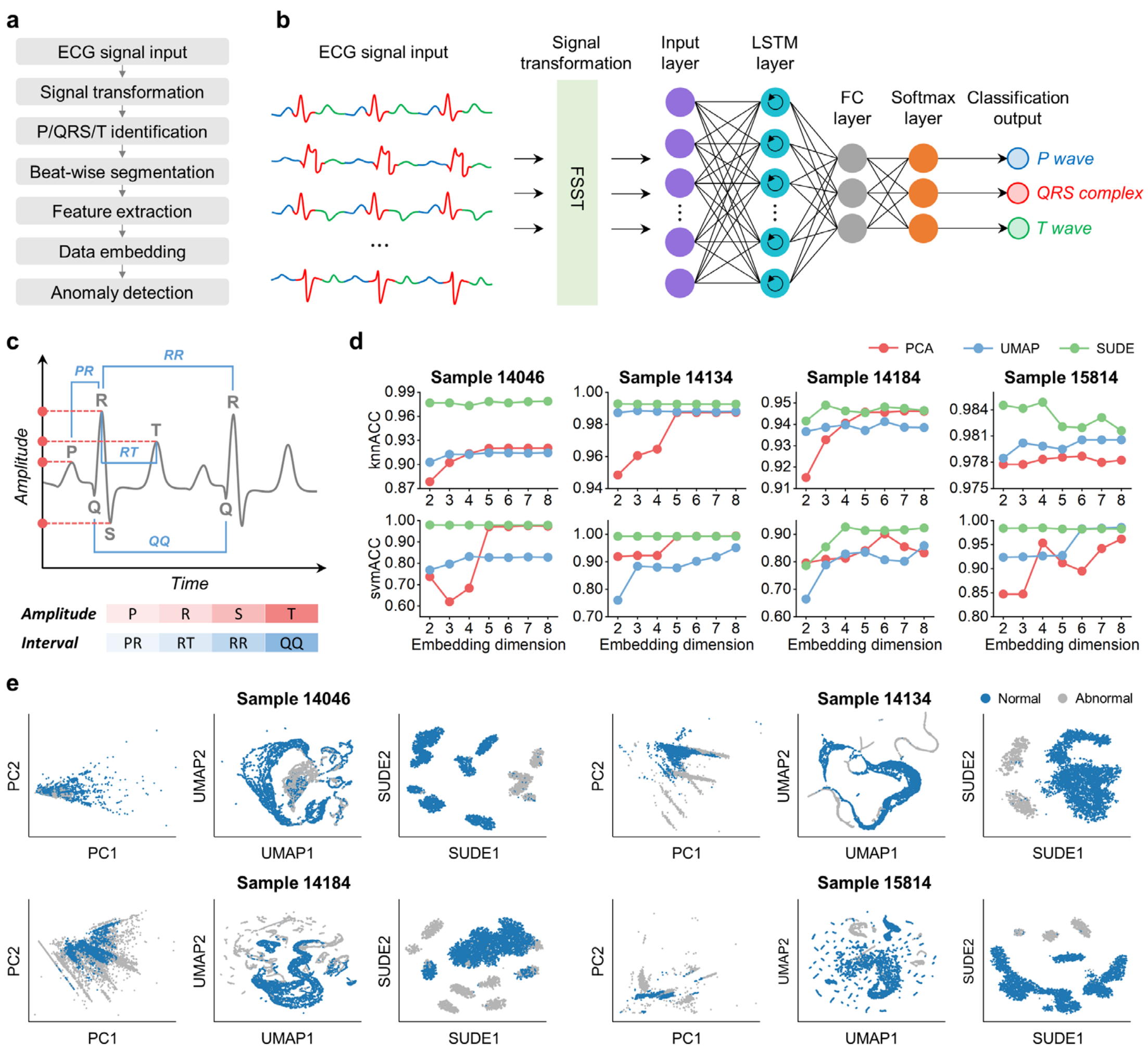

**Fig. 6. Recognition of the abnormal heartbeats in ECG signals.** (**a**), The procedure of abnormal heartbeat detection. (**b**), The deep neural network architecture used for identifying the wave types in the training phase, composed of an input layer, an LSTM recurrent layer with 200 hidden units, a fully connected layer, a softmax layer, and a classification output layer. (**c**), Schematic of eight waveform features, including four peak amplitudes of P, R, S, and T waves, and four interval features, PR, RT, RR and QQ intervals. (**d**), knnACC and svmACC scores of the PCA, UMAP, and SUDE ($k_1 = 20$) under different embedding dimensions. (**e**), The PCA, UMAP, and SUDE embeddings on four ECG datasets.

Identifying the P/QRS/T waves is the fundamental step in the overall workflow. Considering the powerful capability of the long short-term memory (LSTM) network in learning and representing chronological sequences [60], an LSTM-based deep learning architecture was constructed for this purpose. The training phase of this architecture is depicted in Fig. 6b. We pre-trained this model using the published available QT database, which contains 210 fifteen-minute two-lead ECG recordings from 105 patients. These recordings include the annotations of P, QRS, and T waves [61]. Each ECG signal consists of 225,000 sample points with a sampling rate of 250 Hz. We randomly selected 70% of the signals as the training set and used the remaining as the test set. The FSST of the first 10,000 sample points in each signal were fed to the model. We used the Adam optimizer and set the mini-batch size, maximum number of epochs, initial learning rate, and gradient threshold to 2,000, 10, 0.01, and 1 respectively. This model finally yielded a training accuracy of 94.73% and a prediction accuracy of 95.59%.

To testify to the effectiveness of this workflow and the performance of SUDE, we chose four ECG recordings (sample 14046, 14134, 14184, and 15814) with a sampling rate of 128 Hz from the MIT-BIH long-term database, which were collected from four cardiac patients of different ages [62]. Each sample point in the ECG signals has been manually annotated into five types, namely normal beat, premature ventricular contraction, fusion of ventricular and normal beat, junctional premature beat, and supraventricular premature beat. The four beat types, excluding the normal beat, were considered as anomalies. We selected the first two and a half hours of signals from each complete recording as input for this workflow. Each extracted ECG signal contains 1,152,000 sample points with 21.6% anomalies in average and is sufficient for evaluating the effectiveness of our method. After calculating the waveform features of each beat segment, we computed the ACC scores of KNN and SVM classifiers under different embedding dimensions (Fig. 6d). In general, SUDE outperforms PCA and UMAP both in knnACC and svmACC on all four ECG datasets. Notably, SUDE exhibits greater stability as the embedding dimension varies. This indicates that SUDE can avoid crowded clusters even in low-dimensional spaces. The 2-D embeddings generated by PCA, UMAP, and SUDE are displayed in Fig. 6e. In the PCA embeddings, normal and abnormal beats are mixed, making them hard to separate. While UMAP produces too many tiny clusters that disrupt the cluster integrity, especially on samples 14184 and 15184, since it struggles to generate a coherent KNN graph for data with heterogeneous density. The ECG signals of the patients with arrhythmias are usually unstable, thus the extracted waveform features exhibit significant variations during cardiac cycles. As a result, the points in feature space tend to exhibit density heterogeneity. In contrast, SUDE preserves cluster integrity better via early aggregation and neighborhood range expansion through landmark sampling.

# Discussion

The goal of manifold learning is to uncover the intrinsic low-dimensional structure of high-dimensional data by preserving geometric relationships among data points. The manifold hypothesis posits that high-dimensional observational data, such as natural images and sensor measurements, typically lie near or on a smooth low-dimensional manifold embedded within the ambient space [63] (A theoretical proof is provided in Supplementary Note 5). The representation of such manifold data in high-dimensional ambient space is not conducive to classification and clustering tasks due to the data sparsity and risk of overfitting, and imposes burdens on data storage and computation [64]. Although many excellent manifold learning methods, such as t-SNE and UMAP, have achieved success in uncovering the intrinsic structure of high-dimensional data, the scalability presents a challenge. In this work, our method SUDE can generate a uniform and discriminative embedding, and exhibits promising performance in scalability (Fig. 4b-d), sampling robustness (Supplementary Fig. 18), and preservation of global structure (Fig. 2f and 5f, and Supplementary Fig. 31). The improvement of computational efficiency is derived from reducing the number of samples involved in embedding learning through PPS, and faster convergence enabled by the logarithmic similarity kernel and momentum-driven gradient descent. Lower sampling rates boost computational efficiency but necessarily sacrifice the information of data structure. To enhance sampling robustness, SNN-based early aggregation is proposed to cope with the potential undersampling problem and maintain the intra-cluster compactness. CLLE jointly utilizes local linear representations and distance relationships to generate a well-separated layout, while reducing the losses of local and global structures as sampling rate decreases. For the preservation of global structure, PPS enables SUDE to capture long-distance relationships by expanding the neighborhood range for landmark embedding optimization. Besides, Laplacian eigenmaps can yield an initial layout with more accurate positional relationships between clusters [14].

Dimension reduction is a trade-off process, where certain information must be sacrificed to retain other essential features when projecting data into lower-dimensional spaces [65]. In this paper, we evaluated the performances of SUDE in classification accuracy, clustering accuracy, preservation of global/local structures, and scalability. The results demonstrate that SUDE exhibits significant advantages in cluster separation and computational efficiency. It shows relatively weaker performance in local structure preservation across small datasets, but outperforms the competitors except BH-t-SNE on most large datasets as discussed in Fig. 4a. In terms of global structure, neural network-based methods like TopoAE and P-UMAP excelling at preserving pairwise distance relationships. SUDE (0.921) slightly leads UMAP (0.911) with respect to the average CC sore across all real-world datasets in Fig. 4. Besides, SUDE also demonstrates ability in continuous variation preservation as illustrated in Supplementary Fig. 31. In general, users can select appropriate manifold learning methods based on their specific requirements. When prioritizing well-separated clusters and time efficiency, SUDE can be a suitable choice. Nonetheless, there remain room for further improvement. The number of landmarks generated by PPS decreases as $k_1$ increases, and lies between $n/(k_1+1)$ and $n-k_1$. Due to data heterogeneity, the exact number is hard to determine by $k_1$, which makes it tough to accurately estimate the complexity. Hence, the accurate relation between $k_1$ and the sampling rate should be investigated. Furthermore, the hyperparameter settings can be refined. Apart from $k_1$, the hyperparameters have either been fixed or adaptively specified based on empirical attempts. For example, SUDE fixes the aggregation coefficient $\gamma = 1.2$ in early aggregation, but the most appropriate $\gamma$ probably changes with the sampling rate. Therefore, it would be beneficial to investigate better hyperparameter settings and combination strategies.

# Methods

## Preprocessing

After inputting the vector-based data, we first remove the duplicate observations and utilize min-max normalizing to standardize the scale of each feature into [0, 1]. Then, we conduct a KNN search accordingly. SUDE involves two rounds of KNN search. The first round is dedicated to landmark sampling with $k_1$ nearest neighbors, while the second is for constructing the high-dimensional probabilities of landmarks with $k_2$ nearest neighbors. In the implementation of SUDE, K-D tree is used by default to accelerate KNN search. However, when handling excessive dimensions, the construction of K-D tree becomes time-consuming. Thus, for data with more than 5,000 points (denoted as $n > 5{,}000$) and 50 dimensions (denoted as $D_X > 50$), we employ PCA to project the data into a lower-dimensional space beforehand, whose dimensionality is selected to ensure that the cumulative contribution rate is slightly above 0.8. To be noted, we only use PCA for the two rounds of KNN search, while manipulating the data of original dimensionality in other steps. Besides, the implementation of SUDE also provides alternative approximate KNN (ANN) methods such as the HNSW (Hierarchical Navigable Small World) [66] algorithm and the NN-Descent algorithm [67].

## Early aggregation

When clusters are close and the sampling rate is excessively low, the undersampling problem may occur. Undersampling refers to the situation where the landmarks generated by PPS are sparse and far apart from each other under a very low sampling rate, which weakens the intra-cluster compactness and discrimination between clusters. As a result, the landmarks from different clusters are more likely to be mixed together in the low-dimensional embedding space. To tackle this problem, we expect to maintain intra-cluster compactness by measuring the landmark-pair dissimilarity based on their associations in the original data. We employed the shared nearest neighbor (SNN) method to query the common nearest neighbors and measure the degree of association. Landmark pairs with a significant number of SNN in the original data should also have high similarity in the sampled data. Additionally, we consider that SNN with a high local density (e.g., RNN) should carry more weight. Hence, we define a pairwise SNN similarity based on the KNN query results from PPS

$$SNN_{ij} = \sum_{\boldsymbol{x}_u \in KNN(\boldsymbol{x}_i) \cap KNN(\boldsymbol{x}_j)} RNN_u \tag{1}$$

where $KNN(\boldsymbol{x}_i)$ refers to the set of the $k_1$ nearest neighbors of landmark $\boldsymbol{x}_i \in \mathbb{R}^{D_X}$, and $RNN_u$ denotes the RNN of point $\boldsymbol{x}_u$, both in the original data. Then, the dissimilarity from landmark $\boldsymbol{x}_j$ to landmark $\boldsymbol{x}_i$ can be defined as

$$d_{j|i} = \left(1 - \frac{SNN_{ij}}{\max_i SNN_{ij}}\right)^{\gamma} \|\boldsymbol{x}_i - \boldsymbol{x}_j\| \tag{2}$$

where $\gamma$ is the aggregation coefficient. Early aggregation can restore the between-cluster separation as illustrated in Supplementary Note 10.

## High- and low-dimensional probabilities

As depicted in Algorithm 2 of Supplementary Note 1, we construct the high-dimensional probabilities using a neighborhood graph, considering only the $k_2$ nearest neighbors of each landmark while excluding the landmarks that are far away. Based on early aggregation, we employ a Gaussian kernel to calculate the conditional probabilities of landmark $\boldsymbol{x}_j$ to landmark $\boldsymbol{x}_i$

$$p_{j|i} = \begin{cases} \exp\left(-\frac{{d_{j|i}}^2}{2{\sigma_i}^2}\right), \boldsymbol{x}_j \in KNN(\boldsymbol{x}_i) \\ 0, else \end{cases} \tag{3}$$

The higher the SNN similarity between landmarks $\boldsymbol{x}_i$ and $\boldsymbol{x}_j$, the more significantly their high-dimensional probability will be enhanced. Here, $KNN(\boldsymbol{x}_i)$ represents the $k_2$ nearest neighbors of landmark $\boldsymbol{x}_i$ determined by the dissimilarity in equation (2). The variance $\sigma_i$ of Gaussian contributes to producing uniform point density in the embedding. In densely populated regions, it is more appropriate to use a smaller $\sigma_i$. t-SNE and UMAP compute $\sigma_i$ via binary search to align the Shannon entropy of high-dimensional probabilities with a predefined perplexity. To smooth the point density in a more time efficient manner, we define $\sigma_i$ as the average KNN dissimilarity of landmark $\boldsymbol{x}_i$

$$\sigma_i = \frac{1}{k_2} \sum_{\boldsymbol{x}_j \in KNN(\boldsymbol{x}_i)} d_{j|i} \tag{4}$$

Subsequently, the high-dimensional probabilities can be normalized and symmetrized as

$$p_{ij} = \frac{p_{i|j} + p_{j|i}}{2\sum_{k \neq l} p_{k|l}} \tag{5}$$

We adopt a logarithmic function to measure the low-dimensional probabilities

$$q_{ij} = \frac{\left(1 + \log(1 + \|\boldsymbol{y}_i - \boldsymbol{y}_j\|^2)\right)^{-1}}{\sum_{k \neq l}\left(1 + \log(1 + \|\boldsymbol{y}_k - \boldsymbol{y}_l\|^2)\right)^{-1}} \tag{6}$$

Such a logarithmic similarity kernel has a heavier tail than t-distribution, and can achieve more significant cluster separation within fewer training epochs.

## Gradient descent

Kullback-Leibler divergence (KLD) is leveraged as the cost function

$$\mathcal{L} = \sum_{i \neq j} p_{ij} \log \frac{p_{ij}}{q_{ij}} \tag{7}$$

We can derive the gradient as

$$\frac{\partial \mathcal{L}}{\partial \boldsymbol{y}_i} = 4 \sum_j \frac{(p_{ij} - q_{ij})(\boldsymbol{y}_i - \boldsymbol{y}_j)}{(1 + \|\boldsymbol{y}_i - \boldsymbol{y}_j\|^2)\left(1 + \log(1 + \|\boldsymbol{y}_i - \boldsymbol{y}_j\|^2)\right)} \tag{8}$$

The detailed derivation can be seen in Supplementary Note 3. By using gradient descent method with a momentum term, the update is given by

$$\boldsymbol{y}_i{}^{(t)} = \boldsymbol{y}_i{}^{(t-1)} - \eta(t)\left(\frac{\partial \mathcal{L}}{\partial \boldsymbol{y}_i{}^{(t)}} + \alpha(t)\frac{\partial \mathcal{L}}{\partial \boldsymbol{y}_i{}^{(t-1)}}\right) \tag{9}$$

where $\boldsymbol{y}_i{}^{(t)} \in \mathbb{R}^{D_Y}$, $\eta(t)$ and $\alpha(t)$ are the low-dimensional coordinate of landmark $\boldsymbol{x}_i$, learning rate and momentum term in the $t$ th epoch respectively. $\boldsymbol{y}_i{}^{(0)}$ is initialized by Laplacian eigenmaps (Algorithm 3 in Supplementary Note 1), and the ultimate low-dimensional coordinates of landmarks can be obtained using Algorithm 4 in Supplementary Note 1.

## Constrained locally linear embedding

After obtaining the low-dimensional coordinates of landmarks, we proceed to incorporate the non-landmarks into the learned space. Since LLE has high computational efficiency and is robust to rotations, rescalings, and translations, it can be exploited to reconstruct the linear weight $w_i$ between the query non-landmark $\boldsymbol{x} \in \mathbb{R}^{D_X}$ and its $i$-th nearest landmark $\boldsymbol{x}_i$. In $D_Y$-dimensional embedding space, we use the $D_Y + 1$ nearest landmarks to determine the initial position of $\boldsymbol{x}$ as $\boldsymbol{y}' = \sum_{i=1}^{D_Y+1} w_i \boldsymbol{y}_i$.

However, these landmarks may come from different clusters. In this case, $\boldsymbol{x}$ tends to be mistakenly inserted into the inter-cluster gaps or incorrect clusters, especially when $\boldsymbol{x}$ is near the fuzzy cluster boundaries. This can result in dirty clusters and outliers. To evade this issue, we leverage the distance $d_1$ between $\boldsymbol{y}$ and $\boldsymbol{y}_1$ (embedding of the nearest landmark $\boldsymbol{x}_1$) and impose such a distance constraint on the naïve objective function of LLE as follows

$$\operatorname{argmin} \|\boldsymbol{y}-\boldsymbol{y}'\|^2,\ s.t.\, \|\boldsymbol{y}-\boldsymbol{y}_1\|^2 = {d_1}^2 \tag{10}$$

Supplementary Note 4 provides a detailed solving process using Lagrange multiplier technique of this objective. Overall, we have the solution of $\boldsymbol{y}$

$$\boldsymbol{y} = \boldsymbol{y}_1 - d_1 \frac{\boldsymbol{y}_1 - \boldsymbol{y}'}{\|\boldsymbol{y}_1 - \boldsymbol{y}'\|} \tag{11}$$

The key to solve $\boldsymbol{y}$ is to obtain $w_i$ and $d_1$. Suppose that $\boldsymbol{W} = [w_1, w_2, \dots, w_{D_Y+1}]^T \in \mathbb{R}^{D_Y+1}$ denotes the reconstructed linear weights of the current non-landmark $\boldsymbol{x}$, and let the matrix $\boldsymbol{X} = [\boldsymbol{x}_1, \boldsymbol{x}_2, \dots, \boldsymbol{x}_{D_Y+1}] \in \mathbb{R}^{D_X \times (D_Y+1)}$ includes its $D_Y + 1$ nearest landmarks, then we can restate the object of LLE as

$$\operatorname{argmin} \phi = \boldsymbol{W}^T \boldsymbol{G} \boldsymbol{W}, s.t.\, \mathbf{1}^T \boldsymbol{W} = 1 \tag{12}$$

where $\boldsymbol{G} = (\boldsymbol{x}\mathbf{1}^T - \boldsymbol{X})^T(\boldsymbol{x}\mathbf{1}^T - \boldsymbol{X}) \in \mathbb{R}^{(D_Y+1)\times(D_Y+1)}$ and $\mathbf{1} = [1,1,..,1]^T \in \mathbb{R}^{D_Y+1}$. It is worth noting that sometimes the Gram matrix $\boldsymbol{G}$ is singular or nearly singular, in which case it should be replaced by $\boldsymbol{G} + \left(\frac{\Delta^2}{D_Y+1}\right) \cdot \operatorname{tr}(\boldsymbol{G}) \cdot \boldsymbol{I}$ [68, 69], where $\Delta = 0.1$ and $\boldsymbol{I} \in \mathbb{R}^{(D_Y+1)\times(D_Y+1)}$ is the identity matrix, thus we can obtain the solution

$$\boldsymbol{W} = \frac{\boldsymbol{G}^{-1}\mathbf{1}}{\mathbf{1}^T \boldsymbol{G}^{-1} \mathbf{1}} \tag{13}$$

To compute $d_1$, we first calculate the high-dimensional distance between $\boldsymbol{x}$ and $\boldsymbol{x}_1$, then transform it by the optimal distance scale. The optimal scale of $\boldsymbol{y}_1$ can be solved by seeking the mapping relationship between the high- and low-dimensional distances between the KNN of $\boldsymbol{y}_1$

$$\operatorname{argmin} C = \sum_{i=1}^{\frac{k_2(k_2-1)}{2}} (scale \cdot d_{1i} - d'_{1i})^2 \tag{14}$$

where $d_{1i}$ and $d'_{1i}$ denote the $i$-th pairwise distances between the KNN of $\boldsymbol{y}_1$ in high- and low-dimensional spaces respectively. Set the derivative of $C$ to 0

$$\frac{\partial C}{\partial scale} = 2 \sum_{i=1}^{\frac{k_2(k_2-1)}{2}} d_{1i}(scale \cdot d_{1i} - d'_{1i}) = 0 \tag{15}$$

Then, we can solve the optimal scale

$$scale = \frac{\sum_{i=1}^{\frac{k_2(k_2-1)}{2}} d_{1i} d'_{1i}}{\sum_{i=1}^{\frac{k_2(k_2-1)}{2}} {d_{1i}}^2} \tag{16}$$

Hence, we can obtain $d_1$

$$d_1 = \|\boldsymbol{y} - \boldsymbol{y}_1\| = scale \cdot \|\boldsymbol{x} - \boldsymbol{x}_1\| \tag{17}$$

The calculation of optimal scale and CLLE are summarized as Algorithms 5 and 6 in Supplementary Note 1, respectively, and the efficacy of CLLE can be found in Supplementary Note 11.

## Hyperparameter settings

In the first-round KNN search, the parameter $k_1$ controls the sampling rate and affects the time efficiency of SUDE. Users have the flexibility to specify it according to their requirements by trading off the efficiency and quality. For $k_2$ in the second-round KNN search, existing studies indicate that it is typically associated with the number of landmarks $N$ [70], and a linear or logarithmic relation between them has been demonstrated in the KNN-based cluster analysis [71, 72]. In our previous work, we revealed the insensitivity of the KNN parameter and leveraged a continuous piecewise function to establish this relation that ensures the growth speed to flatten out when $N > 1{,}000$. Its effectiveness has been validated through an empirical study on 26 datasets [48]. Based on this idea, we conducted a series of empirical tests on real-world datasets (Supplementary Fig. 25), and summarized a continuous piecewise function to determine $k_2$

$$k_2 = \begin{cases} \lceil \log_2 N \rceil + 18, N \geq 1{,}000 \\ \left\lceil \dfrac{N}{50} \right\rceil + 8, 50 \leq N \leq 1{,}000 \\ 9, 9 \leq N \leq 50 \\ N, N < 9 \end{cases} \tag{18}$$

where $\lceil \cdot \rceil$ denotes the nearest integer upwards.

In the learning stage, we fix the aggregation coefficient to $\gamma = 1.2$ for SNN-based early aggregation. Then, we adaptively adjust the learning rate using a constant warm-up strategy and cosine annealing schedule [30]

$$\eta(t) = \begin{cases} \eta_{max}, t \leq T_{warm} \\ \eta_{min} + (\dfrac{\eta_{max} - \eta_{min}}{2})(1 + \cos((\dfrac{t - T_{warm}}{T_{epoch} - T_{warm}})\pi)), else \end{cases} \tag{19}$$

where the maximum and minimum learning rates are set to $\eta_{max} = 2.5N$ and $\eta_{min} = 2N$, the number of warm-up epochs and training epochs are specified to $T_{warm} = 10$ and $T_{epoch} = 50$ respectively. Besides, the momentum term can be adaptively set to $\alpha(t) = \frac{t-1}{t+2}$ in the $t$th epoch.

## Evaluation metrics

**Offset distance of centroids:** The offset distance of centroids (ODOC) is used to quantify the consistency of cluster position between the data before and after sampling, which is formulated as

$$ODOC = \sum_{i=1}^{c} \|\boldsymbol{\mu}_i - \boldsymbol{\mu}'_i\|^2 \tag{20}$$

where $c$ refers to the number of clusters, $\boldsymbol{\mu}_i$ and $\boldsymbol{\mu}'_i$ denote the centroid of the $i$th cluster before and after sampling.

**Congruence of cluster size:** The congruence of cluster size (CCS) measures the cosine similarity of the ratio of points contained in each cluster between the data before and after sampling, which is defined as

$$CCS = \frac{\sum_{i=1}^{c} n_i \cdot n_i'}{\sqrt{\sum_{i=1}^{c} {n_i}^2} \cdot \sqrt{\sum_{i=1}^{c} {n_i'}^2}} \tag{21}$$

where $c$ refers to the number of clusters, $n_i$ and $n_i'$ denote the number of points in the $i$th cluster before and after sampling.

**Congruence coefficient:** The congruence coefficient (CC) is able to evaluate the preservation of the global or macroscopic structure in embedding [33]. It is defined as the cosine similarity between pairwise distances in the high-dimensional and embedding spaces

$$CC = \frac{\sum_{i=1}^{\frac{n(n-1)}{2}} d_i \cdot d_i'}{\sqrt{\sum_{i=1}^{\frac{n(n-1)}{2}} {d_i}^2} \cdot \sqrt{\sum_{i=1}^{\frac{n(n-1)}{2}} {d_i'}^2}} \tag{22}$$

where $d_i$ and $d_i'$ denote the $i$th corresponding pairwise distances in high- and low-dimensional spaces respectively.

**Accuracy of KNN, SVM classifiers, and K-means clustering:** To measure the separation of clusters in the lower-dimensional space, we leverage the accuracy (ACC) of two supervised classifiers, i.e., KNN ($k = 5$) and support vector machine (SVM), along with the K-means clustering algorithm. KNN and SVM classifiers involve the following steps: (*i*) apply a nonlinear dimension reduction method to obtain the $D_Y$-dimensional embedding; (*ii*) randomly split the data into training (25%) and test (75%) sets, repeating this process five times to ensure robustness; (*iii*) report the average ACC score across these five trails. For K-means clustering, we perform 200 iterations on the $D_Y$-dimensional embeddings with the true number of clusters.

ACC refers to the accuracy rate of the classification or clustering results compared with the true labels, where the true labels are derived from the ground-truth class assignments of the original dataset. We set the true label vector and the predicted label vector as $\boldsymbol{l} = (l_1, l_2, \dots, l_n) \in \mathbb{R}^n$ and $\boldsymbol{r} = (r_1, r_2, \dots, r_n) \in \mathbb{R}^n$, respectively, and the ACC can be defined as:

$$ACC = \frac{\sum_{i=1}^{n} \delta\big(l_i, map(r_i)\big)}{n} \tag{23}$$

where $\delta(\cdot)$ denotes an indicator function:

$$\delta(a, b) = \begin{cases} 1 & if \quad a = b \\ 0 & otherwise \end{cases} \tag{24}$$

$map(\cdot)$ is a mapping function that maps each predicted label to one of the true cluster labels. Commonly, the best mapping can be found by using the Kuhn-Munkres or Hungarian algorithms. These algorithms first construct a confusion matrix to quantify the association strength between predicted and ground truth labels, and then find an optimal one-to-one mapping that maximizes the global matching score [73].

**KNN recall:** The KNN recall (knnRec) measures how well the local neighborhood structure is preserved. It calculates the fraction of true KNN in the original high-dimensional space that remain as neighbors in the low-dimensional embedding, which is defined as

$$knnRec = \frac{1}{n}\sum_{i=1}^{n}\frac{|KNN_H(\boldsymbol{x}_i) \cap KNN_L(\boldsymbol{x}_i)|}{k} \tag{25}$$

where $KNN_H(\boldsymbol{x}_i)$ and $KNN_L(\boldsymbol{x}_i)$ represent the KNN sets of point $\boldsymbol{x}_i$ in the high-dimensional and embedding spaces, respectively. $k$ is the number of nearest neighbors and is fixed as 10 in this paper.

## Data availability

The real-world datasets used in this study are publicly available: Wine (http://archive.ics.uci.edu/dataset/109/wine), Dermatology (http://archive.ics.uci.edu/dataset/33/dermatology), Breast-Cancer (https://www.csie.ntu.edu.tw/~cjlin/libsvmtools/datasets/binary/breast-cancer), Mfeat (https://archive.ics.uci.edu/dataset/72/multiple+features), Rice (https://archive.ics.uci.edu/dataset/545/rice+cammeo+and+osmancik), Spambase (https://archive.ics.uci.edu/dataset/94/spambase), Dry-Bean (https://archive.ics.uci.edu/dataset/602/dry+bean+dataset), Shuttle (https://archive.ics.uci.edu/dataset/148/statlog+shuttle), CIFAR10 (https://www.cs.toronto.edu/~kriz/cifar.html), MNIST (https://yann.lecun.com/exdb/mnist/), FMNIST (https://github.com/zalandoresearch/fashion-mnist), AG's News (http://www.di.unipi.it/~gulli/AG_corpus_of_news_articles.html), Yahoo (https://github.com/zjulearning/AtSNE), WT (GEO database: accession no. GSE125708), HRPEAC (https://singlecell.broadinstitute.org/single_cell/study/SCP3025/scrna-seq-and-scatac-seq-atlas-of-human-retinal-pigment-epithelium-and-choroid?#study-download), SARS-CoV-2 (https://singlecell.broadinstitute.org/single_cell/study/SCP2593/impact-of-variants-and-vaccination-on-nasal-immunity-across-three-waves-of-sars-cov-2#study-download), Levine (FlowRepository: FR-FCM-ZZPH), QT database (https://www.mathworks.com/supportfiles/SPT/data/QTDatabaseECGData.zip), MIT-BIH long term database (https://archive.physionet.org/physiobank/database/ltdb/).

## Code availability

The source code of SUDE in MATLAB and Python versions has been deposited via Code Ocean at https://doi.org/10.24433/CO.9866909.v3. The TriMap algorithm is available at http://www.omiq.ai. The Seurat software can be found at https://satijalab.org/seurat. The TopoAE model is available at https://github.com/BorgwardtLab/topological-autoencoders. The P-UMAP model is available at https://umap-learn.readthedocs.io/en/latest/parametric_umap.html. The Slingshot toolkit is available at https://github.com/kstreet13/slingshot.

## References

[1] Busch, E. L. et al. Multi-view manifold learning of human brain-state trajectories. *Nat. Comput. Sci.* **3**, 240-253 (2023).

[2] Floryan, D. & Graham, M. D. Data-driven discovery of intrinsic dynamics. *Nat. Mach. Intell.* **4**, 1113-1120 (2022).

[3] Islam, M. T. et al. Revealing hidden patterns in deep neural network feature space continuum via manifold learning. *Nat. Commun.* **14**, 8506 (2023).

[4] Vogelstein, J. T. et al. Supervised dimensionality reduction for big data. *Nat. Commun.* **12**, 2872 (2021).

[5] Hotelling, H. Analysis of a complex of statistical variables into principal components. *J. Educ. Psychol.* **25**, 417-441 (1933).

[6] Peng, D., Gui, Z., Gui, J. & Wu, H. A robust and efficient boundary point detection method by measuring local direction dispersion. IEEE Trans. Circuits Syst. Video Technol., (2025). https://doi.org/10.1109/TCSVT.2025.3546401.

[7] Kruskal, J. B. Multidimensional scaling by optimizing goodness of fit to a nonmetric hypothesis. *Psychometrika* **29**, 1-27 (1964).

[8] Fisher, R. A. The use of multiple measurements in taxonomic problems. *Ann. Eugenics* **7**, 179-188 (1936).

[9] Yan, S. et al. Nonlinear Discriminant Analysis on Embedded Manifold. *IEEE Trans. Circuits Syst. Video Technol.* **17**, 468-477 (2007).

[10] Zhang, L., Zhou, W. & Chang, P.-C. Generalized nonlinear discriminant analysis and its small sample size problems. *Neurocomputing* **74**, 568-574 (2011).

[11] Tenenbaum, J. B., Silva, V. D. & Langford, J. C. A global geometric framework for nonlinear dimensionality reduction. *Science* **290**, 2319-2323 (2000).

[12] Roweis, S. T. & Saul, L. K. Nonlinear dimensionality reduction by locally linear embedding. *Science* **290**, 2323-2326 (2000).

[13] Cover, T. & Hart, P. Nearest neighbor pattern classification. *IEEE Trans. Inf. Theory* **13**, 21-27 (1967).

[14] Belkin, M. & Niyogi, P. Laplacian eigenmaps and spectral techniques for embedding and clustering. In *Advances in neural information processing systems*, pp. 585-591 (NeurIPS, 2002).

[15] Donoho, D. L. & Grimes C. Hessian eigenmaps: Locally linear embedding techniques for high-dimensional data. *Proc. Natl. Acad. Sci. U. S. A.* **100**, 5591-5596 (2003).

[16] Coifman, R. R. et al. Geometric diffusions as a tool for harmonic analysis and structure definition of data: diffusion maps. *Proc. Natl. Acad. Sci. U. S. A.* **102**, 7426-7431 (2005).

[17] van der Maaten, L. & Hinton, G. Visualizing data using t-SNE. *J. Mach. Learn. Res.* **9**, 2579-2605 (2008).

[18] McInnes, L., Healy, J., Saul, N. & Großberger, L. UMAP: uniform manifold approximation and projection. *J. Open Source Softw.* **3**, 861 (2018).

[19] Becht, E. et al. Dimensionality reduction for visualizing single-cell data using UMAP. *Nat. Biotechnol.* **37**, 38-44 (2019).

[20] Amid, E. & Warmuth, TriMap: large-scale dimensionality reduction using triplets. Preprint at https://arxiv.org/abs/1910.00204 (2022).

[21] Fu, C., Zhang, Y., Cai, D. & Ren, X. AtSNE: efficient and robust visualization on GPU through hierarchical optimization. In *Proceedings of the 25th ACM SIGKDD International Conference on Knowledge Discovery & Data Mining*, pp. 176-186 (2019).

[22] Moon, K. R. et al. Visualizing structure and transitions in high-dimensional biological data. *Nat. Biotechnol.* **37**, 1482-1492 (2019).

[23] Kobak, D. & Berens, P. The art of using t-SNE for single-cell transcriptomics. *Nat. Commun.* **10**, 5416 (2019).

[24] Shen, C. & Wu, H.-T. Scalability and robustness of spectral embedding: landmark diffusion is all you need. *Inf. Inference* **11**, 1527-1595 (2022).

[25] Long, A. W. & Ferguson, A. L. Landmark diffusion maps (L-dMaps): Accelerated manifold learning out-of-sample extension. *Appl. Comput. Harmon. Anal.* **47**, 190-211 (2019).

[26] Persad, S. et al. SEACells infers transcriptional and epigenomic cellular states from single-cell genomics data. *Nat. Biotechnol.* **41**, 1746-1757 (2023).

[27] Stanoi, I., Riedewald, M., Agrawal, D. & El Abbadi, A. Discovery of influence sets in frequently updated databases. In *Proceedings of the 27th International Conference on Very Large Databases*, pp. 99-108 (2001).

[28] Tao, Y., Papadias, D. & Lian, X. Reverse kNN search in arbitrary dimensionality. In *Proceedings of the 30th International Conference on Very Large Databases*, pp. 744-755 (2004).

[29] Radovanovic, M., Nanopoulos, A. & Ivanovic, M. Hubs in space: Popular nearest neighbors in high-dimensional data. *J. Mach. Learn. Res.* **11**, 2487-2531 (2010).

[30] Loshchilov, I. & Hutter, F. SGDR: stochastic gradient descent with warm restarts. Preprint at https://arxiv.org/abs/1608.03983 (2016).

[31] Peng, D., Gui, Z. & Wu, H. MeanCut: a greedy-optimized graph clustering via path-based similarity and degree descent criterion. Preprint at https://arxiv.org/abs/2312.04067 (2023).

[32] Karypis, G., Han, E.-H. & Kumar, V. Chameleon: hierarchical clustering using dynamic modeling. *Computer*, **32**, 68-75 (1999).

[33] France, S. & Carroll, D. Development of an agreement metric based upon the Rand Index for the evaluation of dimensionality reduction techniques, with applications to mapping customer data. In *Machine Learning and Data Mining in Pattern Recognition* (eds Perner, P.) vol. 4571, pp. 499-517 (2007).

[34] Moor, M., Horn, M., Rieck, B. & Borgwardt, K. Topological autoencoders. In *Proceedings of the 37th International Conference on Machine Learning*, pp. 7045-7054, (2020).

[35] Sainburg, T., McInnes, L. & Gentner, T. Q. Parametric UMAP embeddings for representation and semisupervised learning. *Neural Comput.* **33**, 2881-2907 (2021).

[36] Dua, D. & Graff, C. UCI machine learning repository. University of California, School of Information and Computer Science, Irvine, CA, (2019).

[37] Krizhevsky, A. & Hinton, G. Learning multiple layers of features from tiny images. Technical Report (Univ. Toronto, 2009).

[38] Lecun, Y., Bottou, L., Bengio, Y. & Haffner, P. Gradient-based learning applied to document recognition. *Proceedings of the IEEE* **86**, 2278-2324 (1998).

[39] Xiao, H., Rasul, K. & Vollgraf, R. Fashion-MNIST: a novel image dataset for benchmarking machine learning algorithms. Preprint at https://arxiv.org/abs/1708.07747 (2017).

[40] Corso, G. M. D., Gulli, A. & Romani, F. Ranking a stream of news. In *Proceedings of the 14th International World Wide Web Conference*, pp. 97-106 (2005).

[41] Poličar, P. G., Stražar, M. & Zupan, B. openTSNE: a modular Python library for t-SNE dimensionality reduction and embedding. *J. Stat. Softw.* **109**, 1-30 (2024).

[42] Huang, H., Wang, Y., Rudin, C. & Browne, E. P. Towards a comprehensive evaluation of dimension reduction methods for transcriptomic data visualization. *Commun. Biol.* **5**, 719 (2022).

[43] Amir, Ea. et al. viSNE enables visualization of high dimensional single-cell data and reveals phenotypic heterogeneity of leukemia. *Nat. Biotechnol.* **31**, 545-552 (2013).

[44] Khan, S. A. et al. Reusability report: Learning the transcriptional grammar in single-cell RNA-sequencing data using transformers. *Nat. Mach. Intell.* **5**, 1437-1446 (2023).

[45] Heng, J. S. et al. Hypoxia tolerance in the Norrin-deficient retina and the chronically hypoxic brain studied at single-cell resolution. *Proc. Natl. Acad. Sci. U. S. A.* **116**, 9103-9114 (2019).

[46] Satija, R. et al. Spatial reconstruction of single-cell gene expression data. *Nat. Biotechnol.* **33**, 495-502 (2015).

[47] Kiselev, V. Y., Andrews, T. S. & Hemberg, M. Challenges in unsupervised clustering of single-cell RNA-seq data. *Nat. Rev. Genet.* **20**, 273-282 (2019).

[48] Peng, D. et al. Clustering by measuring local direction centrality for data with heterogeneous density and weak connectivity. *Nat. Commun.* **13**, 5455 (2022).

[49] Wang, S. K. et al. Single-cell multiome and enhancer connectome of human retinal pigment epithelium and choroid nominate pathogenic variants in age-related macular degeneration. Preprint at https://doi.org/10.1101/2025.03.21.644670 (2025).

[50] Ziegler, C. G. et al. Impaired local intrinsic immunity to SARS-CoV-2 infection in severe COVID-19. *Cell* **184**, 4713-4733.

[51] Levine, J. H. et al. Data-driven phenotypic dissection of AML reveals progenitor-like cells that correlate with prognosis. *Cell* **162**, 184-197 (2015).

[52] Weber, L. M. & Robinson, M. D. Comparison of clustering methods for high-dimensional single-cell flow and mass cytometry data. *Cytom. Part A* **89**, 1084-1096 (2016).

[53] Ferrari, G., Thrasher, A. J. & Aiuti, A. Gene therapy using haematopoietic stem and progenitor cells. *Nat. Rev. Genet.* **22**, 216-234 (2021).

[54] Street, K. et al. Slingshot: cell lineage and pseudotime inference for single-cell transcriptomics. *BMC Genomics* **19**, 477 (2018).

[55] Trapnell, C. et al. The dynamics and regulators of cell fate decisions are revealed by pseudotemporal ordering of single cells. *Nat. Biotechnol.* **32**, 381-386 (2014).

[56] Young, W. J. et al. Genetic analyses of the electrocardiographic QT interval and its components identify additional loci and pathways. *Nat. Commun.* **13**, 5144 (2022).

[57] Niroshana, S. M. I., Kuroda, S., Tanaka, K. & Chen, W. Beat-wise segmentation of electrocardiogram using adaptive windowing and deep neural network. *Sci. Rep.* **13**, 11039 (2023).

[58] Huo, R. et al. ECG segmentation algorithm based on bidirectional hidden semi-Markov model. *Comput. Biol. Med.* **150**, 106081 (2022).

[59] Oberlin, T., Meignen, S. & Perrier, V. The fourier-based synchrosqueezing transform. In *IEEE International Conference on Acoustics, Speech and Signal Processing*, pp. 315-319 (ICASSP, 2014).

[60] Hochreiter, S. & Schmidhuber, J. Long short-term memory. *Neural Comput.* **9**, 1735-1780 (1997).

[61] Laguna, P., Mark, R. G., Goldberg, A. & Moody, G. B. A database for evaluation of algorithms for measurement of QT and other waveform intervals in the ECG. *Comput. Cardiol.* **24**, 673-676 (1997).

[62] Goldberger, A. L. et al. PhysioBank, PhysioToolkit, and PhysioNet: components of a new research resource for complex physiologic signals. *Circulation* **101**, e215-e220 (2000).

[63] Bengio, Y., Courville, A. & Vincent, P. Representation learning: A review and new perspectives. *IEEE Trans. Pattern Anal. Mach. Intell.* **35**, 1798-1828 (2013).

[64] Peng, D., Gui, Z. & Wu, H. Interpreting the curse of dimensionality from distance concentration and manifold effect. Preprint at https://arxiv.org/abs/2401.00422 (2023).

[65] Healy, J., McInnes, L. Uniform manifold approximation and projection. Nat. Rev. Methods Primers **4**, 82 (2024).

[66] Malkov, Y. A. & Yashunin, D. A. Efficient and robust approximate nearest neighbor search using hierarchical navigable small world graphs. *IEEE Trans. Pattern Anal. Mach. Intell.* **42**, 824-836 (2018).

[67] Dong, W., Moses, C. & Li, K. Efficient k-nearest neighbor graph construction for generic similarity measures. In *Proceedings of the 20th International Conference on World Wide Web*, pp. 577-586, (2011).

[68] Saul, L. K. & Roweis, S. T. Think globally, fit locally: unsupervised learning of low dimensional manifolds. *J. Mach. Learn. Res.* **4**, 119-155 (2003).

[69] Wang, J. Real local-linearity preserving embedding. *Neurocomputing* **136**, 7-13 (2014).

[70] Wang, Z. et al. Clustering by local gravitation. *IEEE Trans. Cybern.* **48**, 1383-1396 (2018).

[71] von Luxburg, U. A tutorial on spectral clustering. *Stat. Comp.* **17**, 395-416 (2007).

[72] Ding, J., Shah, S. & Condon, A. densityCut: an efficient and versatile topological approach for automatic clustering of biological data. *Bioinformatics* **32**, 2567-2576 (2016).

[73] Kuhn, H. W. The Hungarian method for the assignment problem. *Nav. Res. Logist. Q.* **2**, 83-97 (1955).

# Supplementary Information

## Contents

### Notes

### Figures

## Tables

## References

## Supplementary Note 1: Pseudocode of the algorithms

**Algorithm 1** Plum pudding sampling (PPS)

*Input*: Point set $\boldsymbol{X}$ and the KNN parameter $k_1$

Search the KNN and obtain the RNN of $\boldsymbol{X}$ with $k_1$
Arrange $\boldsymbol{X}$ as $\{\boldsymbol{x}_1, \boldsymbol{x}_2, \dots, \boldsymbol{x}_n\}$ in RNN descending order
Define an empty landmark set $\boldsymbol{X_l}$ and non-landmark set $\boldsymbol{X_{nl}}$
**for** the first point $\boldsymbol{x}_i$ in $\boldsymbol{X}$
  Add $\boldsymbol{x}_i$ to $\boldsymbol{X_l}$
  Move $\boldsymbol{x}_a, \boldsymbol{x}_b, \dots, \boldsymbol{x}_s \in KNN(\boldsymbol{x}_i)$ from $\boldsymbol{X}$ to $\boldsymbol{X_{nl}}$
**end for**
**return** $\boldsymbol{X_l}$ and $\boldsymbol{X_{nl}}$

**Algorithm 2** Constructing high-dimensional probabilities

*Input*: Landmark set $\boldsymbol{X_l}$ and the KNN parameter $k_2$

Search the KNN of $\boldsymbol{X_l}$ with $k_2$
**for** each point $\boldsymbol{x}_i$ in $\boldsymbol{X_l}$
  **for** each point $\boldsymbol{x}_j$ in $\boldsymbol{X_l}$
    **if** $\boldsymbol{x}_j \in KNN(\boldsymbol{x}_i)$
      Compute pair-wise SNN using equation (1)
      Compute pair-wise dissimilarity using equation (2)
      $p_{j|i} = \exp\left(-{d_{j|i}}^2 / 2{\sigma_i}^2\right)$ where $\sigma_i$ can be obtained using equation (4)
    **else**
      $p_{j|i} = 0$
    **end if**
  **end for**
**end for**
Normalize and symmetrize the probability matrix as $\boldsymbol{P}$ using equation (5)
**return** $\boldsymbol{P}$

**Algorithm 3** Initialization using Laplacian eigenmaps

*Input*: High-dimensional probability matrix $\boldsymbol{P}$ and embedding dimension $D_Y$

Compute the degree matrix $\boldsymbol{D}$
Compute the Laplacian matrix $\boldsymbol{L} = \boldsymbol{D}^{1/2}(\boldsymbol{D} - \boldsymbol{P})\boldsymbol{D}^{1/2}$
Compute the eigenvectors $[\boldsymbol{v}_1, \boldsymbol{v}_2, \dots, \boldsymbol{v}_N]$ of $\boldsymbol{L}$ in ascending order of eigenvalues
$\boldsymbol{Y_l} = [\boldsymbol{v}_2, \boldsymbol{v}_3, \dots, \boldsymbol{v}_{D_Y+1}]$
**return** $\boldsymbol{Y_l}$

**Algorithm 4** Embedding learning

*Input*: High-dimensional probability matrix $\boldsymbol{P}$, initial landmark embedding $\boldsymbol{Y_l}$, the maximum and minimum learning rates $\eta_{max}$ and $\eta_{min}$, the number of warm-up epochs $T_{warm}$ and training epochs $T_{epoch}$

Compute the initial low-dimensional probability matrix $\boldsymbol{Q}$ using equation (6)<br>
**while** $t < T_{epoch}$<br>
    Compute the learning rate using equation (19)<br>
    Compute the gradients using equation (8)<br>
    Update $\boldsymbol{Y_l}$ using equation (9)<br>
    Update $\boldsymbol{Q}$ using equation (6)<br>
    $t = t + 1$<br>
**end while**<br>
**return** $\boldsymbol{Y_l}$

**Algorithm 5** Calculating the optimal scale

*Input*: KNN and embedding $\boldsymbol{Y_l}$ of the landmark set $\boldsymbol{X_l}$

**for** each point $\boldsymbol{x}_i$ in $\boldsymbol{X_l}$<br>
    Compute the pair-wise distances of $KNN(\boldsymbol{x}_i)$ using $\boldsymbol{X_l}$<br>
    Compute the pair-wise distances of $KNN(\boldsymbol{x}_i)$ using $\boldsymbol{Y_l}$<br>
    Compute the optimal scale $scale_i$ of point $\boldsymbol{x}_i$ using equation (16)<br>
**end for**<br>
**return** $scale = [scale_1, scale_2, \dots, scale_N]$

**Algorithm 6** Constrained locally linear embedding

*Input*: Non-landmark set $\boldsymbol{X_{nl}}$, embedding $\boldsymbol{Y_l}$ of the landmark set $\boldsymbol{X_l}$, optimal scale $scale$, and embedding dimension $D_Y$

**for** each point $\boldsymbol{x}_i$ in $\boldsymbol{X_{nl}}$<br>
    Search $D_Y + 1$ nearest landmarks of $\boldsymbol{x}_i$ in $\boldsymbol{X_l}$<br>
    Compute the locally linear weights using equation (13)<br>
    Compute the nearest distance with $scale_i$ using equation (17)<br>
    Compute the embedding $\boldsymbol{y}_i$ of $\boldsymbol{x}_i$ with $\boldsymbol{Y_l}$ using equation (11)<br>
**end for**<br>
Obtain the embedding $\boldsymbol{Y_{nl}} = [\boldsymbol{y}_1, \boldsymbol{y}_2, \dots, \boldsymbol{y}_{n-N}]$ of $\boldsymbol{X_{nl}}$<br>
**return** $\boldsymbol{Y_{nl}}$

**Supplementary Note 2: Time complexity analysis**

We analyze the time complexity of SUDE to assess its computational efficiency. It is mainly composed of seven parts, including preprocessing (complexity is denoted as $O(T_0)$) and six algorithms ($O(T_1) - O(T_6)$) presented in Supplementary Note 1. The notations used in the complexity analysis are defined as follows:

| Notation | Description |
|---|---|
| $D_X$ | The number of dimensions of the original data |
| $D_{\widetilde{X}}$ | The number of embedding dimensions for PCA preprocessing |
| $n$ | The number of samples in the original data |
| $N$ | The number of landmarks selected by PPS |
| $k_1$ | The number of nearest neighbors in PPS |
| $k_2$ | The number of nearest neighbors in embedding learning |
| $T_{epoch}$ | The number of training epochs |
| $p$ | The sampling probability of positive samples in UMAP |
| $m$ | The number of negative samples to select per positive sample in UMAP |

- **Preprocessing**

PCA is an optional step for the two rounds of KNN search. If the original data satisfies $n > 5{,}000$ and $D_X > 50$, we recommend applying PCA to project the data into a lower-dimensional space without losing too much of the important structure. The embedding dimension of PCA can be determined to ensure that the cumulative contribution rate is slightly above 80%. PCA involves two steps, i.e., covariance matrix computation and eigen-value decomposition. Computing covariance matrix takes $O({D_X}^2 n)$. Suppose that the reduced dimension is $D_{\widetilde{X}}$, so finding the $D_{\widetilde{X}}$ largest eigenvalues of this dense matrix using traditional eigenvalue algorithms like the QR algorithm or the Arnoldi method has a complexity of $O({D_X}^2 D_{\widetilde{X}})$. So, the complexity of PCA preprocessing is

$$O(T_0) = O({D_X}^2 n + D_{\widetilde{X}} {D_X}^2)$$

- **Algorithm 1: Plum pudding sampling**

The PPS algorithm consists of KNN search and landmark sampling. Suppose that the dimensionality of $\boldsymbol{X}$ is reduced to $D_{\widetilde{X}}$, the brute-force method for KNN search has a complexity of $O(k_1 D_{\widetilde{X}} n^2)$. There are many KNN acceleration methods available, such as K-D tree, Ball tree, and DCI (Dynamic Continuous Indexing) [1]. Additionally, various approximate KNN (ANN) methods have been developed, including the HNSW (Hierarchical Navigable Small World) algorithm used in TriMap, the NN-Descent algorithm used in LargeVis and UMAP, which achieve a theoretical complexity of $O(n^{1.14})$. Here, we take K-D tree as an example to analyze the complexity, but offer users a variety of KNN methods to choose from in practical implementations. K-D tree construction has a complexity of $O(D_{\widetilde{X}} n \log n)$, and querying the nearest neighbors is $O(n \log n + k_1 n)$. Thus, K-D tree can improve the complexity to $O(D_{\widetilde{X}} n \log n)$. Landmark sampling first arranges the RNN queue in $O(n \log n)$ and then selects landmarks from the queue point by point in $O(n)$. Overall, Algorithm 1 has a complexity of

$$O(T_1) = O(D_{\widetilde{X}} n \log n + n \log n + n) = O(D_{\widetilde{X}} n \log n)$$

- **Algorithm 2: Constructing high-dimensional probabilities**

Constructing the high-dimensional probabilities involves KNN search, similarity computation, and matrix normalization. Suppose that Algorithm 1 generates $N$ landmarks, KNN search takes $O(D_{\widetilde{X}} N \log N)$. The complexity of similarity computation and matrix normalization is $O(k_2 N)$. Hence, the overall complexity is

$$O(T_2) = O(D_{\widetilde{X}} N \log N + 2k_2 N) = O(D_{\widetilde{X}} N \log N)$$

- **Algorithm 3: Initialization using Laplacian eigenmaps**

In Algorithm 3, computing the Laplacian matrix is $O(k_2N)$, while solving the $D_Y$ smallest eigenvalues of the sparse Laplacian matrix is $O(k_2D_YN)$ using the Lanczos iteration. So, the total complexity of Algorithm 3 is

$$O(T_3) = O(k_2N + k_2D_YN) = O(k_2D_YN)$$

- **Algorithm 4: Embedding learning**

Optimizing the embedding of landmarks is an iterative process. In each epoch, computing the gradient involves the calculation of $\boldsymbol{Q} \in \mathbb{R}^{N\times N}$ using equation (6) and the Hadamard product of multiple matrices, both of which have a time complexity of $O(N^2)$. Besides, the update of low-dimensional coordinates is $O(N)$. To sum up, embedding learning has a complexity of

$$O(T_4) = O(T_{epoch}N^2 + T_{epoch}N) = O(T_{epoch}N^2)$$

- **Algorithm 5: Calculating the optimal scale**

Algorithm 5 computes the optimal scales of non-landmarks. The main time-consuming step is to find the $D_Y + 1$ nearest landmarks for each non-landmark. Using the K-D tree constructed in Algorithm 2, it only takes $O(\log N + D_Y + 1)$ for a non-landmark to query its nearest landmarks. For $n - N$ non-landmarks, the complexity of Algorithm 5 is

$$O(T_5) = O((n-N)(\log N + D_Y + 1)) = O((n-N)(\log N + D_Y))$$

- **Algorithm 6: Constrained locally linear embedding**

CLLE needs to be conducted $n - N$ times to obtain the low-dimensional coordinates of non-landmarks. It mainly includes the computation of $\boldsymbol{G} \in \mathbb{R}^{(D_Y+1)\times(D_Y+1)}$ and solving the reconstructed weights $\boldsymbol{W}$ using equation (13). $\boldsymbol{G}$ is determined by matrix multiplication with a complexity of $O((D_Y+1)^2D_X(n-N))$, while solving $\boldsymbol{W}$ needs to calculate the inverse of $\boldsymbol{G}$, which takes $O((D_Y+1)^2\log(D_Y+1)(n-N))$. Algorithm 6 gets

$$O(T_6) = O\big((D_Y+1)^2(D_X + \log(D_Y+1))(n-N)\big) = O({D_Y}^2(D_X + \log D_Y)(n-N))$$

- **Overall complexity**

For convenience, we fix the hyperparameters and regard $k_2$, $D_Y$, and $T_{epoch}$ as constants. Then, we can obtain the overall complexity of SUDE

$$O(T_{sude}) = \sum_{i=0}^{6} O(T_i) = O({D_X}^2 n + D_{\widetilde{X}}{D_X}^2 + D_{\widetilde{X}}n\log n + N^2)$$

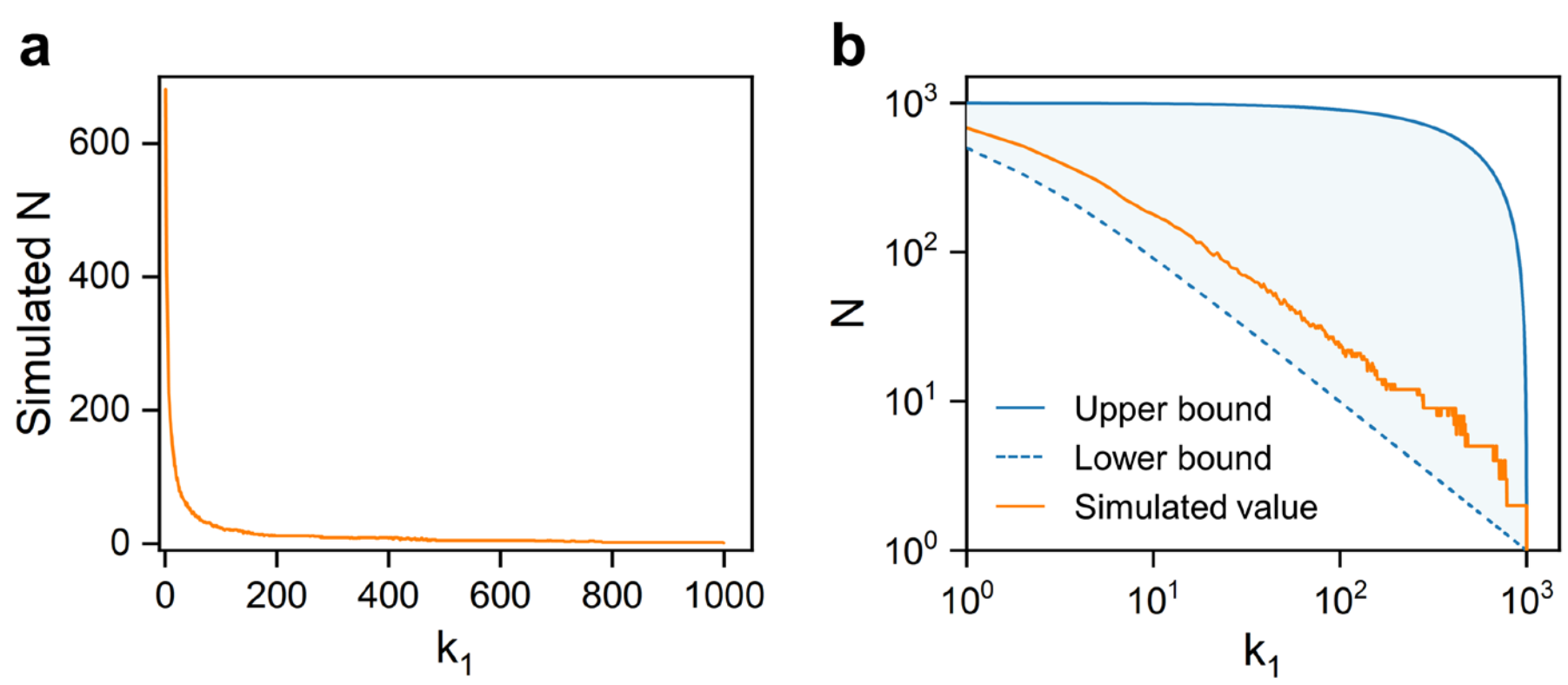

**Supplementary Fig. 1. Estimation of $\boldsymbol{N}$ on a synthetic dataset.** (**a**), The simulated $\boldsymbol{N}$ under different $\boldsymbol{k_1}$. (**b**), The bounds and the simulated results of $\boldsymbol{N}$.

We further analyze the scalability with respect to data size $n$ and assume $D_{\widetilde{X}} < D_X \ll n$. In this case, the complexity can be rewritten as $O(n\log n + N^2)$. The number of landmarks $N$ can be regulated by $k_1$, which lies between $n/(k_1+1)$ (lower bound) and $n - k_1$ (upper bound) in theory. To estimate an empirical value of $N$, we generated 1,000 random points and count $N$ by varying $k_1$ from 1 to 999. $N$ presents a power law distribution with $k_1$ as shown in Supplementary Fig. 1a and approximates to $\epsilon \cdot \frac{n}{k_1+1}$ in Supplementary Fig. 1b, where $\epsilon$ is a constant. In real-world application, we can specify $k_1$ as $\varepsilon \cdot n$. For example, we set $\varepsilon = 10^{-3}$ in Fig. 4d for a million-scale dataset. Therefore, we can obtain the empirical value of $N$

$$N = \frac{\epsilon \cdot n}{k_1 + 1} = \frac{\epsilon \cdot n}{\varepsilon \cdot n + 1}$$

In summary, the overall complexity of SUDE can be simplified as

$$O(T_{sude}) = O(n\text{log}n + N^2) = O\left(n\text{log}n + \left(\frac{\epsilon \cdot n}{\varepsilon \cdot n + 1}\right)^2\right) = O(n\text{log}n)$$

- **SUDE vs. t-SNE and UMAP**

Existing complexity analyses of t-SNE and UMAP are relatively simplistic, with the commonly accepted theoretical complexity of $O(n^2)$ for t-SNE and empirical bound of $O(n^{1.14})$ for UMAP. In this note, we refine their complexity analyses through theoretical derivations.

t-SNE also provides a PCA preprocessing option and includes KNN graph construction and embedding learning (corresponding to Algorithms 2 and 4 in SUDE). So, the complexity of t-SNE is

$$O(T_{tsne}) = O({D_X}^2 n + D_{\widetilde{X}} {D_X}^2 + D_{\widetilde{X}} n\text{log}n + T_{epoch} n^2)$$

By regarding the parameter $T_{epoch}$ as a constant, we have $O(T_{tsne}) = O(n^2)$ in the case of $D_{\widetilde{X}} < D_X \ll n$.

While, UMAP mainly involves KNN graph construction, Laplacian initialization, and embedding learning. Let UMAP use the same K-D-tree-based KNN method, so the total complexity of KNN graph construction and Laplacian initialization is $O(D_X n\text{log}n + k_2 D_Y n)$. To speed up the process of embedding learning, UMAP utilizes stochastic gradient descent and negative sampling. Suppose that $p$ is the positive sampling probability, then the total number of positive sample pairs is $p \cdot k_2 \cdot n$. Let each positive sample select $m$ negative samples, so the total number of negative sample pairs is $p \cdot k_2 \cdot n \cdot m$. In this case, UMAP takes $O(T_{epoch} \cdot p \cdot k_2 \cdot n \cdot (m+1))$ for embedding learning. The time complexity of UMAP can be written as

$$O(T_{umap}) = O(D_X n\text{log}n + (k_2 D_Y + T_{epoch} \cdot p \cdot k_2 \cdot (m+1))n)$$

We specify these parameters of UMAP as constants and assume $D_{\widetilde{X}} < D_X \ll n$, and then we can obtain $O(T_{umap}) = O(n\text{log}n)$.

Through the comparison of computational complexity among the three methods, it can be observed that SUDE reduces the complexity of embedding learning phase from $O(n^2)$ to $O(N^2)$, where the number of landmarks $N$ is commonly far smaller than the total sample size $n$. SUDE achieves the same overall complexity $O(n\text{log}n)$ as UMAP when both methods adopt the KD-tree-based KNN search. To further improve the complexity, we incorporate a "KNN method" parameter in SUDE's implementation, allowing users to select the exact KD-tree-based KNN method or ANN methods (HNSW and NN-Descent). If the NN-Descent algorithm is employed, the complexity of SUDE can be further reduced to $O(n^{1.14})$ theoretically. Moreover, SUDE utilizes a logarithmic similarity kernel that converges faster than the kernel functions of t-SNE and UMAP. This allows SUDE to achieve more compact cluster structures with fewer training epochs $T_{epoch}$. In practical implementation, t-SNE and UMAP require 1000 and 200 epochs, respectively, while SUDE requires only 50 epochs by default.

**Supplementary Note 3: Gradient of the KDL cost function**

We leverage the Kullback-Leibler divergence (KLD) as the cost function

$$\mathcal{L} = KL(\boldsymbol{P}||\boldsymbol{Q}) = \sum_{i\neq j} p_{ij}\log\frac{p_{ij}}{q_{ij}} = \sum_{i\neq j} p_{ij}\log p_{ij} - p_{ij}\log q_{ij}$$

Let $d_{ij} = \|\boldsymbol{y}_i - \boldsymbol{y}_j\|^2$, then we have

$$\frac{\partial\mathcal{L}}{\partial\boldsymbol{y}_i} = -\frac{\partial(\sum_{k\neq l} p_{kl}\log q_{kl})}{\partial\boldsymbol{y}_i} = -2(\boldsymbol{y}_i - \boldsymbol{y}_j)\sum_j \frac{\partial(\sum_{k\neq l} p_{kl}\log q_{kl})}{\partial d_{ij}}$$

Let $\Psi(\boldsymbol{y}_i, \boldsymbol{y}_j) = (1 + \log(1 + d_{ij}))^{-1}$, we can obtain

$$\begin{aligned}
\frac{\partial(\sum_{k\neq l} p_{kl}\log q_{kl})}{\partial d_{ij}} &= \sum_{k\neq l} p_{kl}\frac{\partial\log(\Psi(\boldsymbol{y}_k, \boldsymbol{y}_l)) - \partial\log(\sum_{k\neq l}\Psi(\boldsymbol{y}_k, \boldsymbol{y}_l))}{\partial d_{ij}} \\
&= \sum_{k\neq l} p_{kl}\frac{\partial\log(\Psi(\boldsymbol{y}_k, \boldsymbol{y}_l))}{\partial d_{ij}} - \left(\sum_{k\neq l} p_{kl}\right)\left(\frac{\partial\log\left(\sum_{k\neq l}\Psi(\boldsymbol{y}_k, \boldsymbol{y}_l)\right)}{\partial d_{ij}}\right) \\
&= 2p_{ij}\frac{1}{\Psi(\boldsymbol{y}_i, \boldsymbol{y}_j)}\frac{\partial\Psi(\boldsymbol{y}_i, \boldsymbol{y}_j)}{\partial d_{ij}} - 2\frac{1}{\sum_{k\neq l}\Psi(\boldsymbol{y}_k, \boldsymbol{y}_l)}\left(\frac{\partial\Psi(\boldsymbol{y}_i, \boldsymbol{y}_j)}{\partial d_{ij}}\right) \\
&= -2(p_{ij} - q_{ij})\Psi(\boldsymbol{y}_i, \boldsymbol{y}_j)(1 + d_{ij})^{-1}
\end{aligned}$$

Thus, we can derive the gradient as

$$\frac{\partial\mathcal{L}}{\partial\boldsymbol{y}_i} = 4\sum_j \frac{(p_{ij} - q_{ij})(\boldsymbol{y}_i - \boldsymbol{y}_j)}{(1 + \|\boldsymbol{y}_i - \boldsymbol{y}_j\|^2)\left(1 + \log(1 + \|\boldsymbol{y}_i - \boldsymbol{y}_j\|^2)\right)}$$

**Supplementary Note 4: Solving the constrained locally linear embedding**

We modify the native objective function of LLE by imposing a constraint of the nearest distance

$$\text{argmin } \mathcal{J} = \|\boldsymbol{y} - \boldsymbol{y}'\|^2, \ s.t. \|\boldsymbol{y} - \boldsymbol{y}_1\|^2 = {d_1}^2$$

where $\boldsymbol{y}' = \sum_{i=1}^{D_Y+1} w_i \boldsymbol{y}_i$. Using Lagrange multiplier technique, we introduce a new variable $\lambda$ and define a new function as

$$\mathcal{J} = \|\boldsymbol{y} - \boldsymbol{y}'\|^2 - \lambda(\|\boldsymbol{y} - \boldsymbol{y}_1\|^2 - {d_1}^2)$$

We set the gradient of $e$ to a zero vector

$$\frac{\partial \mathcal{J}}{\partial \boldsymbol{y}} = 2(\boldsymbol{y} - \boldsymbol{y}') - 2\lambda(\boldsymbol{y} - \boldsymbol{y}_1) = \mathbf{0}$$

*i*) If $\lambda = 1$, then we have $\boldsymbol{y}_1 = \boldsymbol{y}'$, which means that $\boldsymbol{y}_1$ coincides with the LLE embedding of $\boldsymbol{y}$. In this case, we do not consider the nearest distance constraint and let $\boldsymbol{y} = \boldsymbol{y}_1$.

*ii*) If $\lambda \neq 1$, we can obtain

$$\boldsymbol{y} - \boldsymbol{y}_1 = \frac{\boldsymbol{y}_1 - \boldsymbol{y}'}{\lambda - 1}$$

Considering the constraint, we have

$$\|\boldsymbol{y} - \boldsymbol{y}_1\|^2 = \frac{\|\boldsymbol{y}_1 - \boldsymbol{y}'\|^2}{(\lambda - 1)^2} = {d_1}^2$$

So, the solution of $\boldsymbol{y}$ can be yielded

$$\boldsymbol{y} = \boldsymbol{y}_1 + \frac{\boldsymbol{y}_1 - \boldsymbol{y}'}{\lambda - 1} = \begin{cases} \boldsymbol{y}_1 + d_1 \dfrac{\boldsymbol{y}_1 - \boldsymbol{y}'}{\|\boldsymbol{y}_1 - \boldsymbol{y}'\|}, \lambda > 1 \\ \boldsymbol{y}_1 - d_1 \dfrac{\boldsymbol{y}_1 - \boldsymbol{y}'}{\|\boldsymbol{y}_1 - \boldsymbol{y}'\|}, \lambda < 1 \end{cases}$$

This shows that the objective function has two extrema, with one being a local optimum. To identify the global optimum, we compute and compare their objective function values

$$\mathcal{J} = \begin{cases} (1 + \dfrac{d_1}{\|\boldsymbol{y}_1 - \boldsymbol{y}'\|_2})^2 \|\boldsymbol{y}_1 - \boldsymbol{y}'\|^2, \lambda > 1 \\ (1 - \dfrac{d_1}{\|\boldsymbol{y}_1 - \boldsymbol{y}'\|_2})^2 \|\boldsymbol{y}_1 - \boldsymbol{y}'\|^2, \lambda < 1 \end{cases}$$

Obviously, the objective reaches the minimum when $\lambda < 1$, in which case

$$\boldsymbol{y} = \boldsymbol{y}_1 - d_1 \frac{\boldsymbol{y}_1 - \boldsymbol{y}'}{\|\boldsymbol{y}_1 - \boldsymbol{y}'\|}$$

**Supplementary Note 5: Why we need manifold learning?**

Curse of dimensionality refers to the phenomenon where common patterns and relationships that hold in low-dimensional space may become complex and counterintuitive with the increase of data dimensionality. As three typical effects in high-dimensional spaces, data sparsity, distance concentration [2, 3], and hubness [4, 5] have been widely regarded as the primary culprits leading to inadequate capacities of representation and generalization, and limited accuracy in machine learning models [6, 7]. To investigate the impact of the curse of dimensionality, we have theoretically analyzed the distance concentration occurred in different distance metrics, including Minkowski, Euclidean, Chebyshev, and cosine distances in a separate study [8]. Here, we address a critical manifestation of the curse of dimensionality: the manifold hypothesis. It posits that high-dimensional data inherently lies along low-dimensional manifolds embedded within the ambient space, where the intrinsic dimension is substantially lower than the original feature dimension. The non-Euclidean geometry characteristics of these manifolds fundamentally alter the notion of similarity measurement in machine learning systems. This geometric complexity renders traditional Euclidean distance suboptimal for capturing meaningful relationships between data points, thereby complicating classification and clustering tasks [9, 10]. A canonical illustration is the Swiss Roll structure, where path-based similarity measures outperform Euclidean distance in capturing its inherent nonlinear relationships [11]. Meanwhile, the intrinsic dimension being lower than the feature dimension means that the data contains an excess of redundant features that impose burdens on data storage and computation.

To better understand the manifold hypothesis, we discover the existence of manifolds of in high-dimensional data from the perspective of PCA asymptotic behavior. Manifold effect actually is more remarkable when the number of samples is much smaller than the feature dimension. It is also called the high dimension low sample size (HDLSS) problem ($D \gg n$), and the asymptotic behavior ($D \to \infty$) has been studied for exploring the PCA consistency [12-15]. The eigenvalues also called variances in PCA can reflect the magnitude of information in each dimension of the data. Thus, we propose to reveal the existence of manifolds through the lens of cumulative explained variance (CEV) in PCA.

Before formal derivation, let us present a key lemma below, whose proof can be found in [8].

**Lemma 1:** Given a sequence of random variables with finite variance $\boldsymbol{A} = \{a_1, a_2, \dots a_n\}$, if $\lim_{n\to\infty} \mathbf{E}(\boldsymbol{A}) = c$ and $\lim_{n\to\infty} var(\boldsymbol{A}) = 0$, then we will have $\lim_{n\to\infty} a_n = c$.

**Definition:** Let $\boldsymbol{X} = (\boldsymbol{x}_1, \boldsymbol{x}_2, \dots \boldsymbol{x}_n) \in \mathbb{R}^{D\times n}$ be the data matrix that holds $n$ random and independent samples $\boldsymbol{x}_i = (x_{1i}, x_{2i}, \dots x_{Di})^T$. If $\boldsymbol{\mu} = (\mu_1, \mu_2, \dots \mu_D)^T$ denotes the mean of each row of $\boldsymbol{X}$, where $\mu_i = \sum_{j=1}^{n} x_{ij} / n$, the population covariance matrix can be defined as

$$\boldsymbol{C} = \begin{pmatrix} c_{11} & c_{12} & \cdots & c_{1D} \\ c_{21} & c_{22} & \cdots & c_{2D} \\ \vdots & \vdots & & \vdots \\ c_{D1} & c_{D2} & \cdots & c_{DD} \end{pmatrix}, c_{ij} = \frac{1}{n}\sum_{l=1}^{n}(x_{il} - \mu_i)(x_{jl} - \mu_j)$$

**Proposition 1:** Suppose that the entries in $\boldsymbol{X}$ are independent and identically distributed in a finite range $[s, t]$ with the fixed expectation $\mathbf{E}(x)$ and variance $var(x)$, and the eigenvalues of $\boldsymbol{C}$ are arranged as $0 \le \lambda_1 \le \lambda_2 \le \cdots \le \lambda_D$ in ascending order, if $D > n$, then we have that the CEV of the first $D - n$ smallest eigenvalues converges to 0, which can be restated as

$$\lim_{D\to\infty} CEV = \lim_{D\to\infty} \frac{\sum_{i=1}^{D-n} \lambda_i}{\sum_{i=1}^{D} \lambda_i} = 0$$

**Proof:** We compute the limits of $\sum_{i=1}^{D} \lambda_i$ (Part I) and $\sum_{i=1}^{D-n} \lambda_i$ (Part II) separately as follows

*Part I:* Since $\sum_{i=1}^{D} \lambda_i = \sum_{i=1}^{D} c_{ii}$, we have

$$\sum_{i=1}^{D} \lambda_i = \frac{1}{n}\sum_{i=1}^{D}\sum_{j=1}^{n}(x_{ij} - \mu_i)^2 = \frac{1}{n^2}\sum_{i=1}^{D}\left(\sum_{j=1}^{n} {x_{ij}}^2 - n{\mu_i}^2\right) = \frac{1}{n^2}\sum_{i=1}^{D}\sum_{j\ne l}^{n}({x_{ij}}^2 + {x_{il}}^2 - 2x_{ij}x_{il})$$

We can first obtain the limit of the expectation

$$\lim_{D\to\infty}\mathbf{E}\left(\frac{1}{D}\sum_{i=1}^{D}\lambda_i\right)=\lim_{D\to\infty}\frac{1}{Dn^2}\mathbf{E}\left(\sum_{i=1}^{D}\sum_{j\neq l}^{n}({x_{ij}}^2+{x_{il}}^2-2x_{ij}x_{il})\right)$$
$$=\lim_{D\to\infty}\frac{n-1}{n}\Big(\mathbf{E}(x^2)-\mathbf{E}^2(x)\Big)\ =\frac{n-1}{n}var(x)$$

Then, we can compute the limit of the variance

$$\lim_{D\to\infty}var\left(\frac{1}{D}\sum_{i=1}^{D}\lambda_i\right)=\lim_{D\to\infty}\frac{n-1}{Dn}\Big(var(x^2)-var^2(x)-2\mathbf{E}^2(x)var(x)\Big)$$
$$\leq\lim_{D\to\infty}\frac{n-1}{Dn}\Big(\mathbf{E}(x^4)\Big)\leq\lim_{D\to\infty}\frac{n-1}{Dn}\Big(max(s^4,t^4)\Big)=0$$

Based on Lemma 1, we can conclude that

$$\lim_{D\to\infty}\frac{1}{D}\sum_{i=1}^{D}\lambda_i=\frac{n-1}{n}var(x)$$

*Part II:* Given any eigenvalue $\lambda$ and its eigenvector $\boldsymbol{v}=(v_1,v_2,\dots v_D)^T$, $\boldsymbol{v}^T\boldsymbol{v}=1$, we have

$$\lambda=\boldsymbol{v}^T\boldsymbol{C}\boldsymbol{v}=\frac{1}{n}\sum_{i=1}^{D}\sum_{j=1}^{D}v_iv_j\sum_{l=1}^{n}(x_{il}-\mu_i)(x_{jl}-\mu_j)=\frac{1}{n}\sum_{l=1}^{n}\sum_{i=1}^{D}v_i(x_{il}-\mu_i)v_j\sum_{j=1}^{D}(x_{jl}-\mu_j)$$
$$=\frac{1}{n}\sum_{l=1}^{n}\sum_{i=1}^{D}v_i(x_{il}-\mu_i)\boldsymbol{v}^T(\boldsymbol{x}_l-\boldsymbol{\mu})=\frac{1}{n}\sum_{l=1}^{n}\Big(\boldsymbol{v}^T(\boldsymbol{x}_l-\boldsymbol{\mu})\Big)^2$$

We assume $\widetilde{\boldsymbol{X}}=(\boldsymbol{x}_1-\boldsymbol{\mu},\boldsymbol{x}_2-\boldsymbol{\mu},\dots\boldsymbol{x}_n-\boldsymbol{\mu})\in\mathbb{R}^{D\times n}$ is the mean-centered data matrix, and consider the homogeneous system of linear equations $\boldsymbol{v}^T\widetilde{\boldsymbol{X}}=\boldsymbol{0}$ as follow

$$\begin{cases}\boldsymbol{v}^T(\boldsymbol{x}_1-\boldsymbol{\mu})=0\\ \boldsymbol{v}^T(\boldsymbol{x}_2-\boldsymbol{\mu})=0\\ \qquad\vdots\\ \boldsymbol{v}^T(\boldsymbol{x}_n-\boldsymbol{\mu})=0\end{cases}$$

For $rank\Big(\widetilde{\boldsymbol{X}}\Big)\leq n<D$, that means this system has $D-n$ basic solutions at least. By Gram-Schmidt orthogonalization, these $D-n$ orthonormal vectors $\boldsymbol{v}_1,\boldsymbol{v}_2,\dots\boldsymbol{v}_{D-n}$ can be solved from the linearly independent solutions. These vectors are also the eigenvectors of $\boldsymbol{C}$, and corresponds the $D-n$ smallest eigenvalues, thus we can obtain

$$\lambda_1=\lambda_2=\cdots=\lambda_{D-n}=0$$

In summary, we have

$$\lim_{D\to\infty}CEV=\lim_{D\to\infty}\frac{\sum_{i=1}^{D-n}\lambda_i}{\sum_{i=1}^{D}\lambda_i}=\frac{\lim\limits_{D\to\infty}\frac{1}{D}\sum_{i=1}^{D-n}\lambda_i}{\lim\limits_{D\to\infty}\frac{1}{D}\sum_{i=1}^{D}\lambda_i}=0$$

Proposition 1 indicates that the $D$-dimensional data can be 100% explained using only $n$ principal components (PC) when $D>n$. In fact, manifolds exist not only when $D>n$. $rank(\widetilde{\boldsymbol{X}})$ is always smaller than $D$ in real-world data even when $D\leq n$. This means that the system $\boldsymbol{v}^T\widetilde{\boldsymbol{X}}=\boldsymbol{0}$ has more than one solution, and there exists more than one zero eigenvalue of $\boldsymbol{C}$ accordingly, indicating that the data exhibits a manifold structure. A weaker conclusion is that the contribution of the first smallest eigenvalue tends to decrease monotonically as $D$ increases, which can be depicted as

$$\frac{\lambda_1(\boldsymbol{C}^{(D)})}{\sum_{i=1}^{D}\lambda_i(\boldsymbol{C}^{(D)})} \leq \frac{\lambda_1(\boldsymbol{C}^{(D-1)})}{\sum_{i=1}^{D-1}\lambda_i(\boldsymbol{C}^{(D-1)})} \leq \cdots \leq \frac{\lambda_1(\boldsymbol{C}^{(1)})}{\sum_{i=1}^{1}\lambda_i(\boldsymbol{C}^{(1)})}$$

where $\boldsymbol{C}^{(D)}$ denotes the covariance matrix of the $D$-dimensional data. Using Cauchy Interlace Theorem, this conclusion can be easily proved.

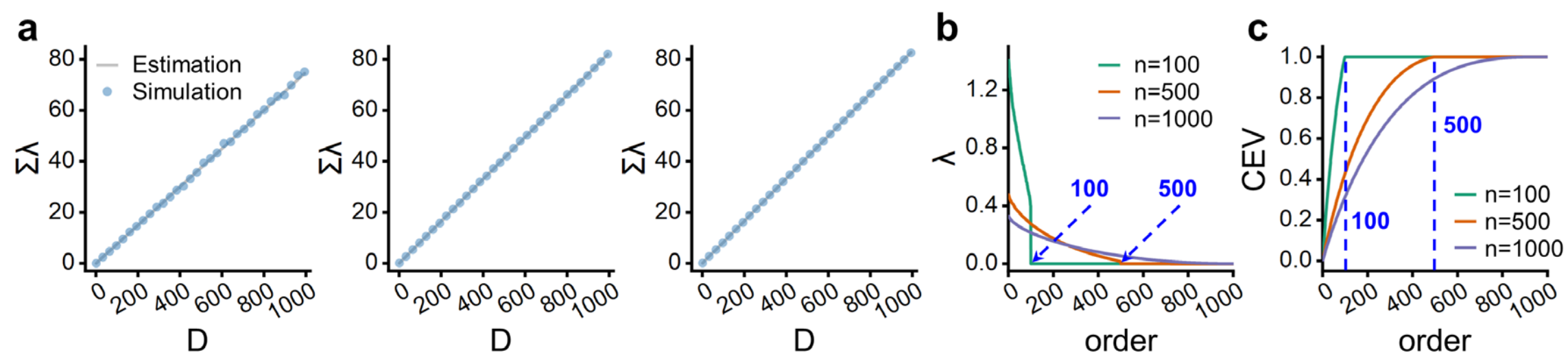

**Supplementary Fig. 2. The asymptotic behavior of PCA.** (**a**), The estimated and simulated values of $\boldsymbol{\sum_{i=1}^{D}\lambda_i}$ as $\boldsymbol{D}$ increases on 100, 500, and 1,000 data points, respectively. The eigenvalues in descending order (**b**), and corresponding CEVs (**c**), on a 1,000-D synthetic dataset with different $\boldsymbol{n}$.

To further demonstrate the manifold hypothesis empirically, we conducted a simulation experiment on synthetic datasets with 100, 500, and 1,000 points and dimensions from 1 to 1,000. The value of each feature is independent and uniformly distributed in [0, 1] with the variance of 1/12. As shown in Supplementary Fig. 2a, the simulated values of $\sum_{i=1}^{D}\lambda_i \propto D$ with the slope closely approximating $\frac{n-1}{n}var(x)$. It is consistent with $\lim_{D\to\infty}\sum_{i=1}^{D}\lambda_i = \frac{n-1}{n}var(x)D$ in the proof of Proposition 1. This agreement intensifies as the sample size $n$ increases. Furthermore, the descending-ordered eigenvalues and the corresponding CEVs are shown in Supplementary Fig. 2b,c respectively. The $D-n$ smallest eigenvalues are 0 and the CEVs of the $n$ largest eigenvalues are equal to 1, which coincides with the above derivations.

**Supplementary Note 6: Why uniform landmark sampling matters?**

To enhance the computational efficiency of manifold learning, we propose a landmark sampling algorithm termed PPS to select a subset of data as landmark points for embedding learning. It is characterized by its uniform sampling, stable outcome, and high time efficiency. It is important to note that the "uniform sampling" mentioned here does not refer to statistical uniform sampling, where each sample has an equal probability of being selected. Instead, it describes that the generated landmarks can often approximately uniformly distributed across each individual component distribution. This has been demonstrated in Fig. 2c of the main text, where the PPS-generated landmarks exhibit a higher CCS score. In terms of sampling probability, PPS is closer to stratified sampling. We explain the motivation of PPS from the following two aspects.

- **Uniform sampling can better preserve the original data distribution**

PPS aims to reduce computational load in manifold embedding procedures while preserving the structure of the original data. To validate this point, a sampling example is presented in Supplementary Fig. 3. Given $n$ high-dimensional data points, we partition the feature space into multiple grid cells of the same size. If there exists $m$ non-empty cells that contain $n_1, n_2, \dots, n_m$ points, respectively, then $\sum_{i=1}^{m} n_i = n$ holds. Now, we plan to select $N$ points from the original data with sampling probabilities $p_1, p_2, \dots, p_m$ $(0 \le p_i \le 1)$ for the corresponding $m$ grid cells. So, we have $\sum_{i=1}^{m} p_i n_i = N$. Let $\boldsymbol{U} = (u_1, u_2, \dots, u_m)$ and $\boldsymbol{V} = (v_1, v_2, \dots, v_m)$ be the distributions of the original data and the sampled data, respectively, where $u_i = \frac{n_i}{n}$ and $v_i = \frac{p_i n_i}{N}$. We can find that if the sampling probabilities satisfy $p_1 = p_2 = \dots = p_m$, then $p_i = \frac{N}{n}$, at which point $u_i = v_i$. It means that the sampled data has the same distribution with the original data when the sampling is uniform.

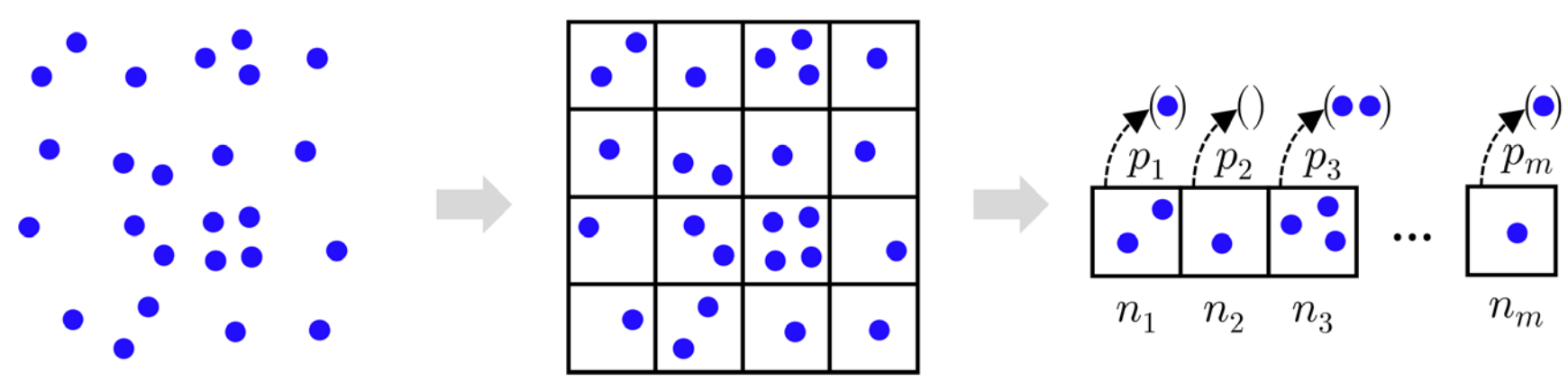

**Supplementary Fig. 3. Illustration of the landmark sampling process.**

We state this insight as Proposition 2 and provide a more rigorous proof using KL divergence (KLD).

**Proposition 2:** If $\boldsymbol{U} = (u_1, u_2, \dots, u_m) \in \mathbb{R}^m$ and $\boldsymbol{V} = (v_1, v_2, \dots, v_m) \in \mathbb{R}^m$ represent the distributions of the original data and the sampled data, respectively, then

$$\mathcal{L} = KL(\boldsymbol{P}||\boldsymbol{Q}) = \sum_{i=1}^{m} u_i \log \frac{u_i}{v_i} \ge 0$$

$\mathcal{L}$ reaches the minimum value of zero if and only if the probabilities satisfy $p_1 = p_2 = \dots = p_m$.

**Proof:** We can rewrite the objective as

$$\underset{p_i}{\operatorname{argmin}} \mathcal{L} = -\sum_{i=1}^{m} \frac{n_i}{n} \log \frac{p_i n_i}{N}, s.t. \sum_{i=1}^{m} p_i n_i = N$$

Using Lagrange multiplier technique, we introduce variable $\lambda$ and define a new function as

$$\mathcal{L} = -\sum_{i=1}^{m} \frac{n_i}{n} \log \frac{p_i n_i}{N} - \lambda \left( \sum_{i=1}^{m} p_i n_i - N \right)$$

We compute the partial derivatives and set them to zero

$$\frac{\partial \mathcal{L}}{\partial p_i} = \frac{\partial (\frac{n_i}{n} \log \frac{p_i n_i}{N} - \lambda p_i n_i)}{\partial p_i} = \frac{n_i}{p_i n} - \lambda n_i = 0$$

Thus, we have

$$p_i = \frac{1}{\lambda n}$$

It means that if and only if $p_1 = p_2 = \cdots = p_m$, then $\mathcal{L}$ reaches its minimum value of zero.

- **Uniform sampling can reduce the uncertainty of non-landmark embedding**

When landmark points are uniformly distributed across the coverage area of the original data, the distances from each non-landmark point to its nearest landmarks are generally small. Supplementary Fig. 4a illustrates two different landmark sampling results. It can be observed that in the first subfigure with approximately uniform sampling, the distances from the non-landmark $\boldsymbol{y}$ to the landmarks $\boldsymbol{y}_1$ and $\boldsymbol{y}_2$ are shorter. In contrast, there are some non-landmarks far from their nearest landmarks when the sampling exhibits significant spatial heterogeneity.

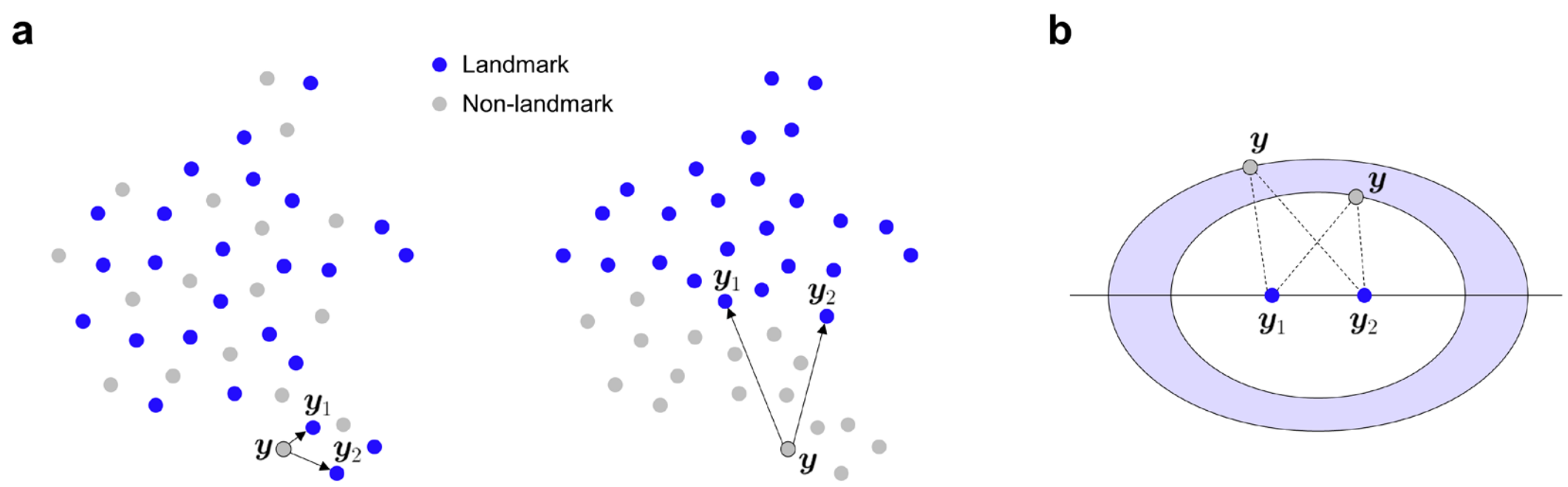

**Supplementary Fig. 4. The impact of landmark sampling on non-landmark embedding.** (**a**), Uniform sampling results in a shorter average distance between landmarks and their nearest non-landmarks. (**b**), Illustration of the uncertainty of non-landmark embedding.

The error magnitude of non-landmark embedding is positively correlated with the distances between the non-landmark and its nearest landmarks. We use a simple example to further illustrate this point. In our proposed method SUDE, the coordinates of non-landmarks are determined by CLLE with $D_Y + 1$ nearest landmarks. But for the sake of simplicity and intuitiveness in explanation, we adopt isometric embedding with two nearest landmarks $\boldsymbol{y}_1$ and $\boldsymbol{y}_2$ to determine the 2-D coordinates of $\boldsymbol{y}$ here.

Let the distances between $\boldsymbol{y}$ to $\boldsymbol{y}_1$ and $\boldsymbol{y}_2$ be $d_1$ and $d_2$ in the high-dimensional space, respectively, then we define the embedding error as $(||\boldsymbol{y}-\boldsymbol{y}_1||+||\boldsymbol{y}-\boldsymbol{y}_2||-d_1-d_2)^2$. Assume that the embedding error tolerance is $\varepsilon \geq 0$ for point $\boldsymbol{y}$, then we have

$$(||\boldsymbol{y}-\boldsymbol{y}_1||+||\boldsymbol{y}-\boldsymbol{y}_2||-d_1-d_2)^2 \leq \varepsilon$$

$$d_1+d_2-\sqrt{\varepsilon} \leq ||\boldsymbol{y}-\boldsymbol{y}_1||+||\boldsymbol{y}-\boldsymbol{y}_2|| \leq d_1+d_2+\sqrt{\varepsilon}$$

In geometry, if the sum of the two distances between a moving point and two focal points is a constant, this moving point lies on an ellipse. So, we have two case as follows:

*i*) If $d_1+d_2 \leq \sqrt{\varepsilon}$, $\boldsymbol{y}$ is inside the ellipse $||\boldsymbol{y}-\boldsymbol{y}_1||+||\boldsymbol{y}-\boldsymbol{y}_2|| \leq d_1+d_2+\sqrt{\varepsilon}$. Its linear eccentricity is equal to $c=\frac{||\boldsymbol{y}_1-\boldsymbol{y}_2||}{2}$, and the length of the semi-major axis is $a=\frac{d_1+d_2+\sqrt{\varepsilon}}{2}$. We can compute the area of this ellipse

$$S=\pi ab=\pi a\sqrt{a^2-c^2}=\frac{\pi(d_1+d_2+\sqrt{\varepsilon})\sqrt{(d_1+d_2+\sqrt{\varepsilon})^2-||\boldsymbol{y}_1-\boldsymbol{y}_2||^2}}{4}$$

*ii*) If $d_1+d_2 > \sqrt{\varepsilon}$, the non-landmark $\boldsymbol{y}$ is located between two ellipses with $\boldsymbol{y}_1$ and $\boldsymbol{y}_2$ as their foci. The lengths of their semi-major axes are $a_1=\frac{d_1+d_2+\sqrt{\varepsilon}}{2}$ and $a_2=\frac{d_1+d_2-\sqrt{\varepsilon}}{2}$, respectively. As shown in Supplementary Fig. 4b, the tolerable range of $\boldsymbol{y}$ is the blue region, whose area equals the difference between the areas of the two ellipses

$$S = \pi(a_1 b_1 - a_2 b_2) = \pi a_1 \sqrt{{a_1}^2 - c^2} - \pi a_2 \sqrt{{a_2}^2 - c^2}$$

Since $a_1 \geq a_2$, we have

$$\pi(a_1 - a_2)\sqrt{{a_1}^2 - c^2} = \pi\sqrt{\varepsilon}\sqrt{{a_1}^2 - c^2} \leq S < \pi a_1 \sqrt{{a_1}^2 - c^2}$$

From the above derivation, it can be concluded that when $d_1 + d_2 \leq \sqrt{\varepsilon}$, the distribution range of non-landmark $\boldsymbol{y}$ is positively correlated with $d_1 + d_2$. While, when $d_1 + d_2 > \sqrt{\varepsilon}$, since both the upper and lower bounds of the area $S$ are positively correlated with $d_1 + d_2$, $S$ also conforms to this law. This indicates that under the same error tolerance, the farther the non-landmark $\boldsymbol{y}$ is from its nearest landmarks, the greater the uncertainty of its position.

**Supplementary Note 7: Why descending RNN sequence is important for PPS?**

The sampling sequence is crucial as it directly impacts the overall structure of landmarks. In the PPS algorithm, we arrange all points in descending order based on the RNN values and select the landmarks in this order. It is primarily driven by three considerations below.

- **Hubs contain more important information of data structure**

The points with high RNN values (*k*-occurrences), known as hubs, exhibit locally high-density regions. Real-world datasets are often better described as a mixture of distributions. It has been observed that hubs tend to be close to the means of individual component distributions of the mixture in a multimodal data [16]. This suggests that hubs tend to reside near cluster centroids, and prioritizing sampling of hubs effectively preserves the integrity of clusters. Moreover, hubs have high degree centrality and betweenness centrality in the KNN graph. Therefore, they are more important to preserve the graph structure. The evidence can be found in [17], where the authors explicitly state in the original text: "*However, (scale-free) network analysis literature often attributes other properties to hubs (Albert and Barabasi, 2002), viewing them as nodes that are important for preserving network structure due to their central positions within the graph, indicated, for example, by their betweenness centrality (Scott, 2000).*"

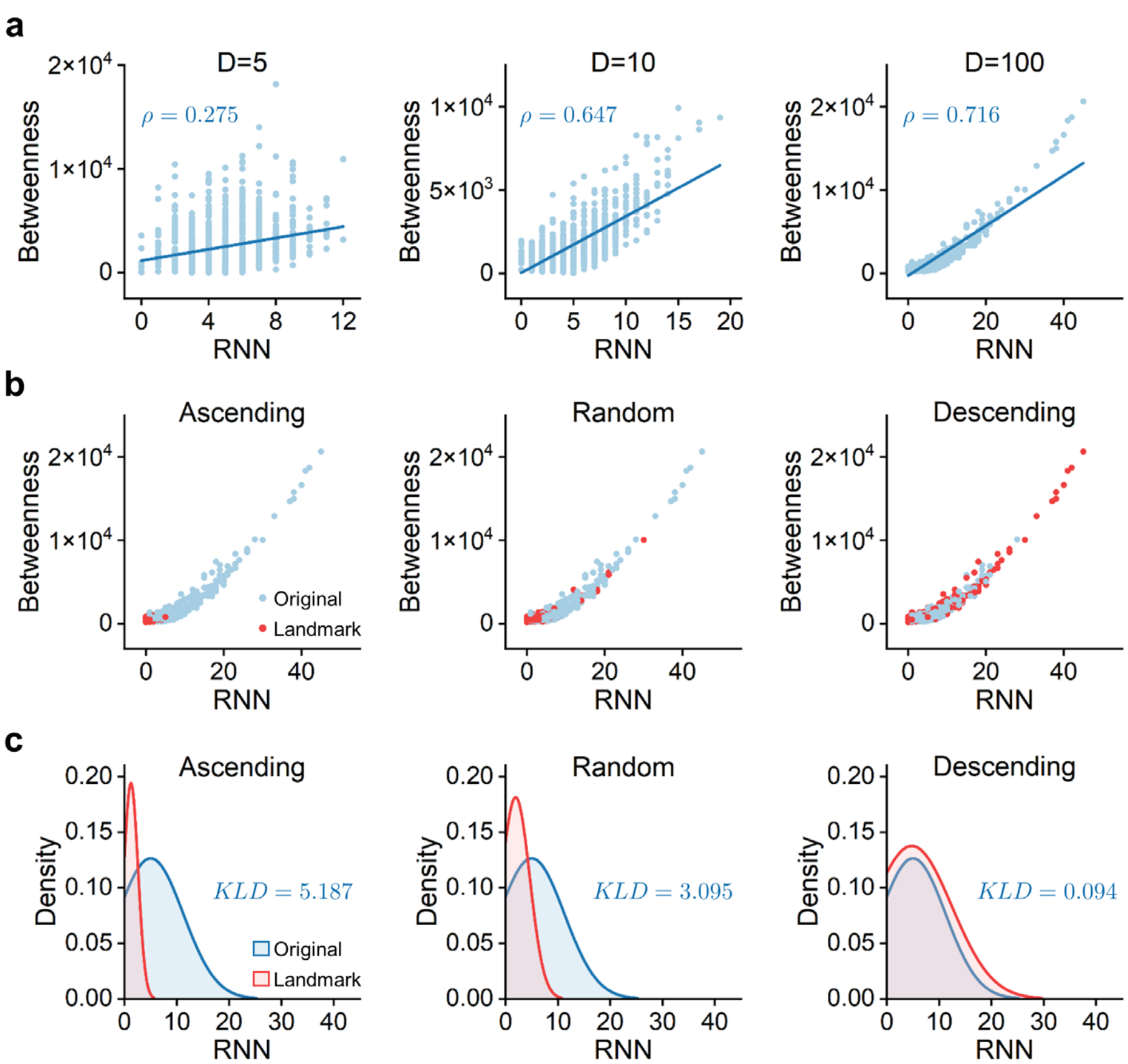

**Supplementary Fig. 5. The importance of hubs for landmark sampling.** (**a**), Spearman correlation analysis between the values of RNN and betweenness centrality. (**b**), RNN-betweenness values and (**c**), RNN density distributions (fitted by normal distribution) of landmarks sampled by PPS with three different RNN sequences in 100-D space, where KLD (Kullback-Leibler divergence) is used to report the similarity between two distributions.

To prove this view, we generated three synthetic datasets containing 1,000 points, with dimensions of 5, 10, and 100, respectively. The value of each feature is independent and uniformly distributed in [0, 1]. The values of RNN and betweenness centrality of the points in the 5-NN graph are calculated. The results of Spearman correlation analysis in Supplementary Fig. 5a reveal that hubs tend to have high betweenness centrality in the KNN graph, and this tendency becomes more significant with increasing dimensionality. We selected landmarks using PPS with ascending, random, and descending RNN sequences on the 100-D dataset. As shown in Supplementary Fig. 5b, the landmarks selected in ascending RNN order have significantly low RNN-betweenness values and fail to cover any hub with high centrality. Random ordering only captures limited hubs. While the descending-order PPS yields landmarks with greater RNN variance and retain those hubs having the highest RNN values. Furthermore, we used normal distribution to fit the RNN distributions of the original data and landmarks in Supplementary Fig. 5c. The hubness phenomenon posits that the RNN values will concentrate toward smaller magnitudes as dimensionality increases. However, the means of the distributions of ascending and random ordering are even smaller than the original distribution, which aggravates the hubness phenomenon. In comparison, descending-order PPS does not aggravate the hubness and generates a closer density distribution with the original data with a much lower $KLD = 0.094$.

Moreover, hubs are viewed as high-quality samples. In the field of image synthesis, a truncated trick has been widely used to randomly sample GAN latents towards their mean based on the observation that the images synthesized from close-to-mean latents are usually of higher quality [18]. Besides, a GAN latent sampling method based on the hubness priors was also proposed [19], which regards that images with high hub values are also of high quality (in Supplementary Fig. 6). The authors described this in the paper: "*Our key insight is that the high dimensionality of the GAN latent space will inevitably lead to the emergence of hub latents that usually have much larger sampling densities than other latents in the latent space. As a result, these hub latents are better trained and thus contribute more to the synthesis of high-quality images.*" Similar observation can be found in protein information mining [20]. An RNN cluster centrality measure was developed to examine the placement of essential proteins based on this insight: "*we have observed that essential proteins tend to be hub proteins in RNN topology. Additionally, essential genes are enriched in clusters containing many hub proteins in RNN topology (RNN protein clusters).*"

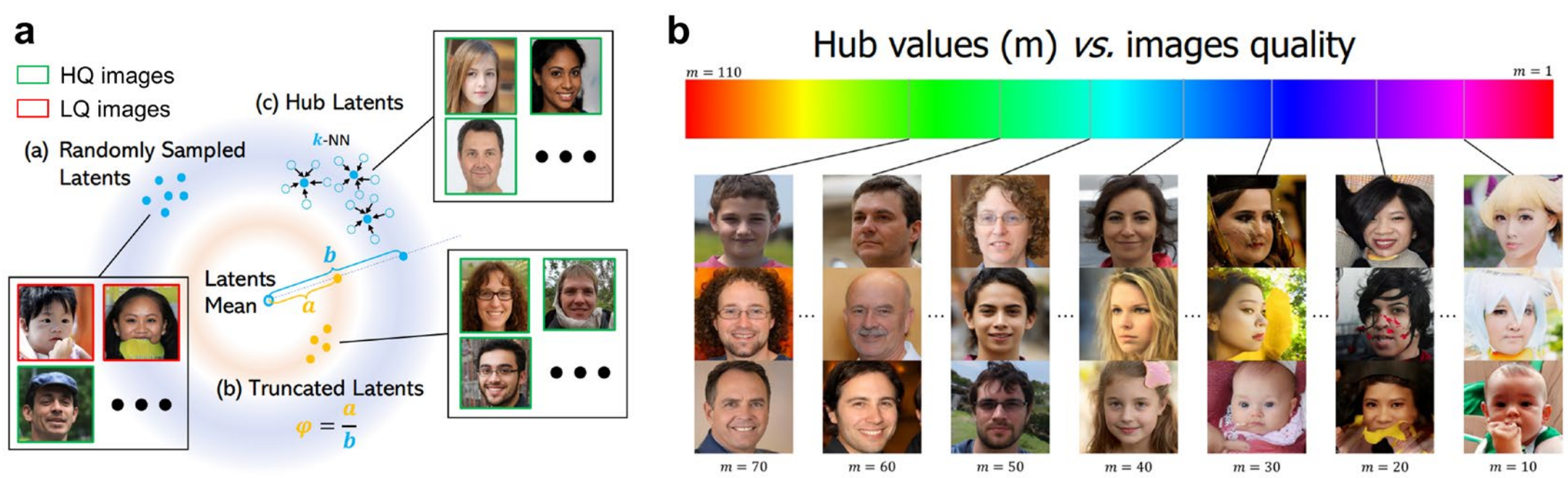

**Supplementary Fig. 6. Exploiting hubness priors for high-quality GAN latent sampling.** (**a**), Illustration of the differences among random sampling, truncation trick, and hub sampling for GAN latents. (**b**), The relationship between hub values and image quality. It indicates that the images who have higher hub values are also of higher quality (Source: https://icml.cc/media/icml-2022/Slides/16224.pdf).

- **Descending RNN sequence makes non-landmarks closer to landmarks**

As mentioned in Supplementary Note 6, the proximity of a non-landmark to its nearest landmark is inversely correlated with the magnitude of embedding error and uncertainty. Hubs display characteristics of spatial centrality. They tend to be close to the center in a unimodal data and be close to the means of individual component distributions of the mixture in a multimodal data. It means that hubs lie in the local density peaks and have lower distances to its neighbors. Supplementary Fig. 7a provides a 5-NN graph that is composed of nine nodes. Nodes 4 and 5 have the highest RNN value of 8; nodes 7, 8, and 9 have an RNN value of 6; while the remaining have an RNN value of 5. Supplementary Fig. 7b show the landmarks sampled by random RNN sequence, where node 1 was selected as a landmark. Since node 1 is an antihub that only attracts nodes 2, 4, and 6, making node 8 have a long distance (i.e., length of the edge) to its nearest landmark node 7. By contrast, descending ordering makes non-landmarks closer to landmarks. In Supplementary Fig. 7c, we give two cases of the descending RNN sequence since nodes 4 and 5 have the same highest RNN value of 8. Suppose we start sampling from node 4 (see the left graph), then nodes 4 and 7 would be chosen as landmarks. As a hub, node 4 attracts five nodes and makes these non-landmarks have a shorter average distance from their nearest landmarks. We further computed the average distances between the non-landmarks to their nearest, second-nearest, and third-nearest landmarks on three synthetic datasets (same dataset as Supplementary Fig. 5), which are denoted as 1NN-Dis, 2NN-Dis, and 3NN-Dis, respectively. Supplementary Fig. 7d indicates that the landmarks selected by descending ordering are closer to non-landmarks.

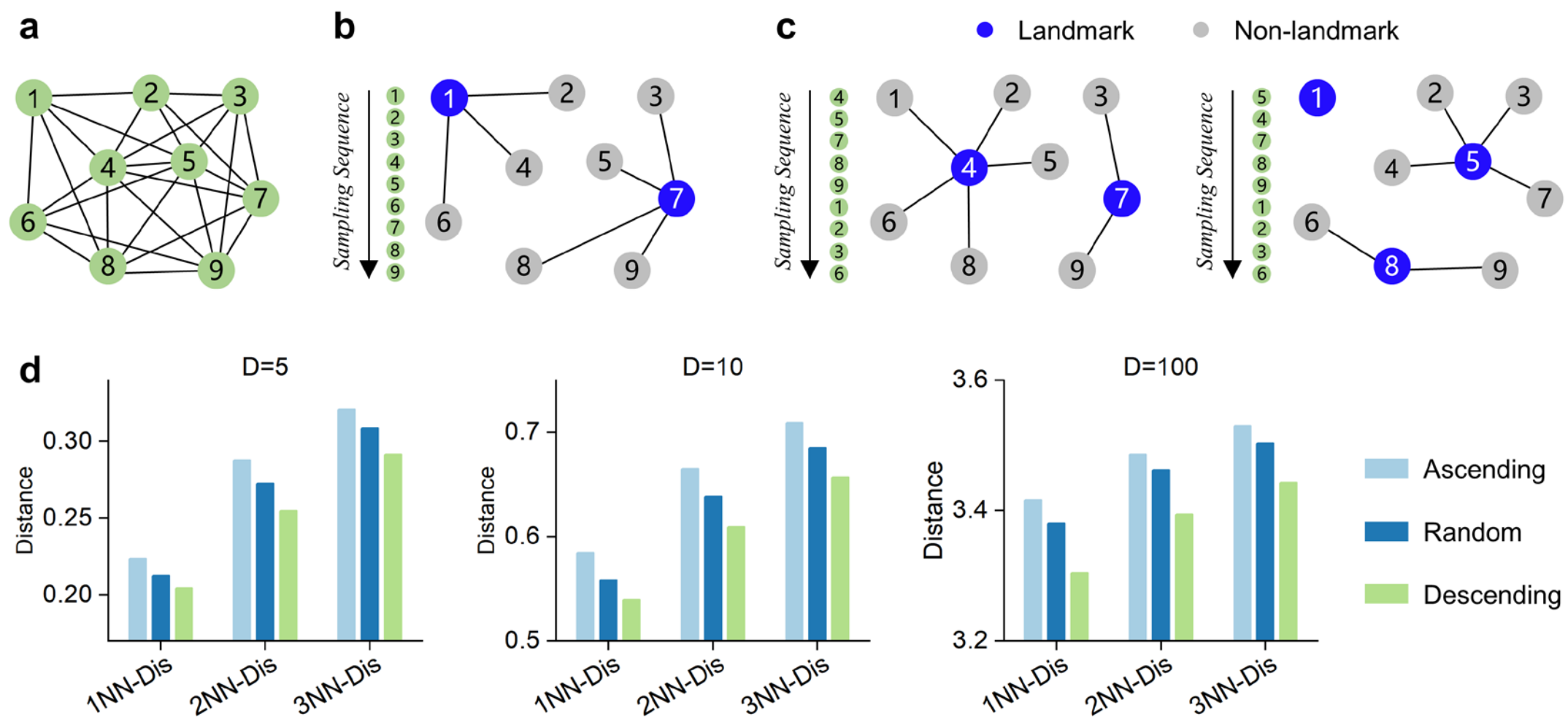

**Supplementary Fig. 7. The impact of RNN sequence on the distance between non-landmarks and their nearest landmarks.** (**a**), The KNN graph of nine nodes with $k = 5$. Landmarks generated by PPS using (**b**), random and (**c**), descending RNN sequences (with two cases). (**d**), A comparison of 1NN-Dis, 2NN-Dis, and 3NN-Dis among the sampling results of three RNN sequences on three synthetic datasets with different dimensions.

- **Hubs have a lower probability of being outliers**

It has been observed that antihubs are more likely to be noise and boundary points of clusters, while hubs tend to reside near the cluster centroids [21]. Based on this observation, an RNN-based outlier detector was proposed [22], and the authors claim an insight in the paper: "*In summary, the emergence of antihubs is closely connected with outliers both in high-dimensional and low-dimensional data. The examples above illustrate this connection, and suggest that antihubs can be used as an alternative to standard outlier-detection methods.*" In this case, we do not want the outliers to have a higher priority to be selected as the landmarks. Assuming that sampling is performed in ascending order of RNN values, most selected landmarks would be outliers or located near the periphery of clusters, leading to the loss of substantial topological information within the clusters.

**Supplementary Note 8: Why the logarithmic kernel function works?**

In this paper, we utilize a logarithmic similarity function to establish the low-dimensional probabilities. Logarithmic $q_{ij}$ promotes a faster convergence for embedding optimization than the Student t-distribution probability and achieves stronger intra-cluster compactness. To illustrate this point, we intend to prove the proposition as follow:

**Definition:** Given a set of pair-wise distances $0 \leq d_1 \leq d_2 \leq \cdots \leq d_n$, the low-dimensional probabilities of t-SNE and SUDE can be written as

$$q_{tsne}(d_i)=\frac{(1+{d_i}^2)^{-1}}{\sum_{j=1}^{n}(1+{d_j}^2)^{-1}} \qquad q_{sude}(d_i)=\frac{\left(1+\log(1+{d_i}^2)\right)^{-1}}{\sum_{j=1}^{n}\left(1+\log(1+{d_j}^2)\right)^{-1}}$$

**Proposition:** We define a function $f(x)$ and a variable $t$ as

$$f(x)=\frac{(1+x^2)^{-1}}{\left(1+\log(1+x^2)\right)^{-1}}\,(x\geq 0) \qquad t=\frac{\sum_{j=1}^{n}(1+{d_j}^2)^{-1}}{\sum_{j=1}^{n}\left(1+\log(1+{d_j}^2)\right)^{-1}}$$

Then, there exists a positive variable $s$ such that $f(s)=t$, and $q_{tsne}(d_i) > q_{sude}(d_i)$ for $d_i < s$ and $q_{tsne}(d_i) < q_{sude}(d_i)$ for $d_i > s$.

**Proof:** We differentiate this function

$$f'(x)=-\frac{\log(1+x^2)}{(1+x^2)^2}\leq 0$$

Thus, $f(x)$ is monotonically decreasing and $0 < f(x) \leq f(0) = 1$. We can further obtain

$$f(x)=\frac{(1+x^2)^{-1}}{\left(1+\log(1+x^2)\right)^{-1}}\leq 1 \Rightarrow (1+x^2)^{-1}\leq\left(1+\log(1+x^2)\right)^{-1}$$

$$\Rightarrow \sum_{j=1}^{n}(1+{d_j}^2)^{-1}\leq\sum_{j=1}^{n}\left(1+\log(1+{d_j}^2)\right)^{-1}\Rightarrow 0<t\leq 1$$

In this case, there must exist $s > 0$ that makes $f(s) = t$. For $d_i < s$, we have $f(d_i) > t$, which is equivalent to

$$\frac{(1+{d_i}^2)^{-1}}{\left(1+\log(1+{d_i}^2)\right)^{-1}} > t = \frac{\sum_{j=1}^{n}(1+{d_j}^2)^{-1}}{\sum_{j=1}^{n}\left(1+\log(1+{d_j}^2)\right)^{-1}}$$

$$\Rightarrow q_{tsne}(d_i)=\frac{(1+{d_i}^2)^{-1}}{\sum_{j=1}^{n}(1+{d_j}^2)^{-1}} > q_{sude}(d_i)=\frac{\left(1+\log(1+{d_i}^2)\right)^{-1}}{\sum_{j=1}^{n}\left(1+\log(1+{d_j}^2)\right)^{-1}}$$

Similarly, for $d_i > s$, we have $q_{tsne}(d_i) < q_{sude}(d_i)$. Proposition hence holds.

By uniformly generating 100 pair-wise distances from [0, 10], the corresponding probabilities using $q_{tsne}$ and $q_{sude}$ are displayed in Supplementary Fig. 8a. As depicted in the above theorem, $q_{tsne}(d_i) > q_{sude}(d_i)$ in the range of smaller distances, while $q_{tsne}(d_i) < q_{sude}(d_i)$ when the distance is large. It means that $q_{sude}$ has a heavier tail than $q_{tsne}$ [23, 24]. In this case, given the same probability, $q_{sude}$ would make the close point pairs more compact in low-dimensional space and push the distant point pairs away, which makes clusters of points are more distinctly separated as shown in Supplementary Fig. 8b.

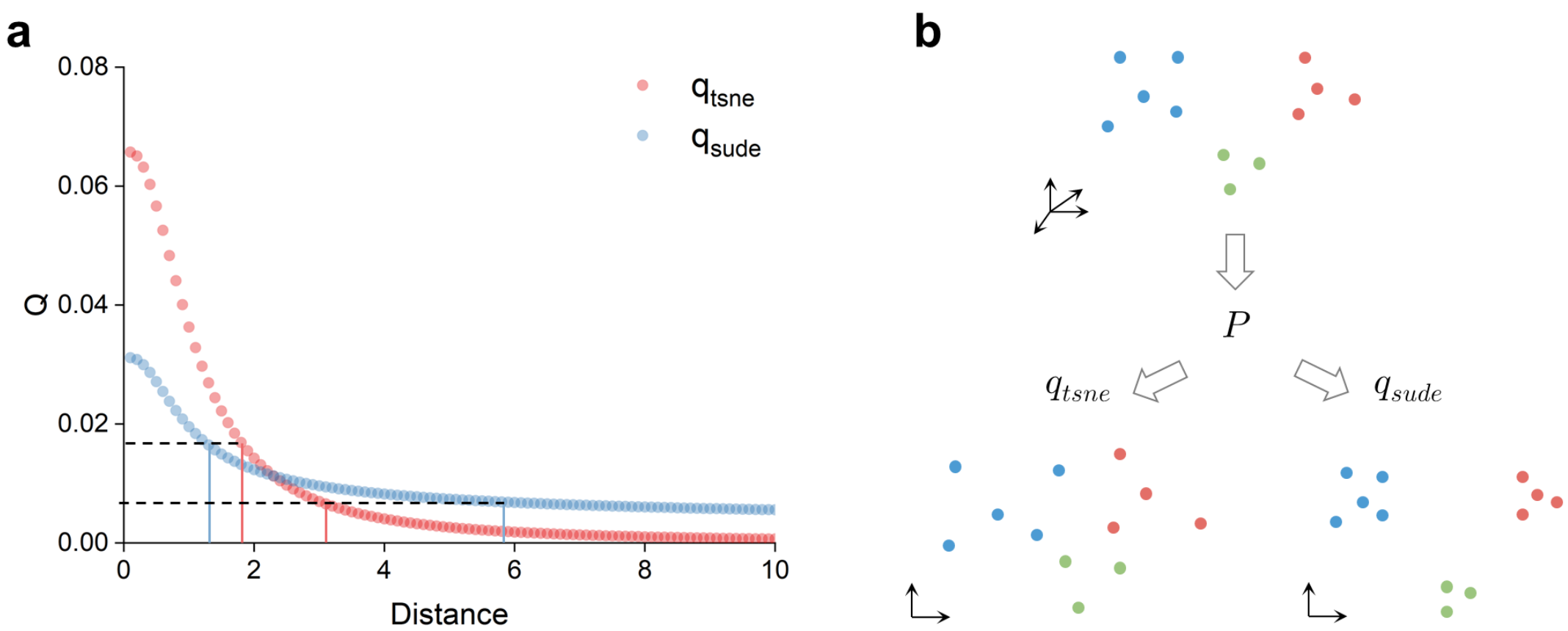

**Supplementary Fig. 8. Comparison between Student-t-based probability and logarithmic probability.** (**a**), Low-dimensional probabilities of 100 pairwise distances. (**b**), Illustration of that $\boldsymbol{q_{sude}}$ can generate more compact embedding than $\boldsymbol{q_{tsne}}$.

In terms of the convergence speed, we can compare the gradients of t-SNE and SUDE

$$Grad_{tsne}(\boldsymbol{y}_i) = 4\sum_j \frac{(p_{ij} - q_{ij})(\boldsymbol{y}_i - \boldsymbol{y}_j)}{(1 + \|\boldsymbol{y}_i - \boldsymbol{y}_j\|^2)}$$

$$Grad_{sude}(\boldsymbol{y}_i) = 4\sum_j \frac{(p_{ij} - q_{ij})(\boldsymbol{y}_i - \boldsymbol{y}_j)}{(1 + \|\boldsymbol{y}_i - \boldsymbol{y}_j\|^2)\left(1 + \log(1 + \|\boldsymbol{y}_i - \boldsymbol{y}_j\|^2)\right)}$$

$Grad_{sude}(\boldsymbol{y}_i)$ is equivalent to attaching a weight $\left(1 + \log(1 + \|\boldsymbol{y}_i - \boldsymbol{y}_j\|^2)\right)^{-1}$ for each point upon $Grad_{tsne}(\boldsymbol{y}_i)$. When updating the coordinates of $\boldsymbol{y}_i$, SUDE pays more attention to the influence of nearby points while weakening the impact of points that are far away. With the assistance of LE initialization, points within the same cluster tend to be close to each other. In this case, the impacts acted on $\boldsymbol{y}_i$ are more likely from the same cluster, and the updates of its coordinates only occurs in the local range. It leads to a faster cluster aggregation in fewer epochs.

## Supplementary Note 9: What is the impact of undersampling problem?

Landmark sampling-based manifold learning inevitably suffers from undersampling problem. It refers to the phenomenon that when the sampling rate is too low, the distribution of landmarks becomes overly sparse. This results in the inter-cluster gaps are no longer distinct, thereby making the manifold learning algorithms to produce crowded low-dimensional embeddings. To interpret this phenomenon intuitively, an example on a toy dataset containing two clusters is provided in Supplementary Fig. 9a. We posit that point $\boldsymbol{x}_0$ in the blue cluster is selected as a landmark by PPS in every sampling iteration, and $\boldsymbol{\mu}_1$ and $\boldsymbol{\mu}_2$ are the centroids of the blue and red clusters, respectively. If we do not apply early aggregation to SUDE, the similarities of $\boldsymbol{\mu}_1$ and $\boldsymbol{\mu}_2$ to $\boldsymbol{x}_0$ can be written as $p_1 = \exp(-\|\boldsymbol{x}_0 - \boldsymbol{\mu}_1\|^2/2\sigma^2)$ and $p_2 = \exp(-\|\boldsymbol{x}_0 - \boldsymbol{\mu}_2\|^2/2\sigma^2)$, respectively, where $\sigma$ is the average KNN distance of $\boldsymbol{x}_0$. To quantify the impact of undersampling problem on cluster distinguishability, we define a clustering quality metric by referring to Davies-Bouldin index (DBI) as

$$\frac{p_1}{p_2} = \frac{\exp(-\|\boldsymbol{x}_0 - \boldsymbol{\mu}_1\|^2/2\sigma^2)}{\exp(-\|\boldsymbol{x}_0 - \boldsymbol{\mu}_2\|^2/2\sigma^2)} = \exp\left(-\frac{\|\boldsymbol{x}_0 - \boldsymbol{\mu}_1\|^2 - \|\boldsymbol{x}_0 - \boldsymbol{\mu}_2\|^2}{2\sigma^2}\right)$$

$p_1/p_2$ measures the ratio of the intra-cluster distance to the inter-cluster distance. A smaller value indicates a better cluster distinguishability. Considering that the landmarks generated by PPS can preserve the structure of the original data well, the cluster centroids $\boldsymbol{\mu}_1$ and $\boldsymbol{\mu}_2$ can be considered to remain unchanged across different sampling iterations. This point has been demonstrated in Fig. 2b of the main text, where PPS achieved the lowest ODOC scores. However, as the sampling rate decreases, the landmarks become sparse, and the average KNN distance of $\boldsymbol{x}_0$ becomes larger, that is, the bandwidth $\sigma$ of its Gaussian distribution increases as illustrated in Supplementary Fig. 9b. Thus, we have $\lim_{\sigma\to\infty} p_1/p_2 = 1$. This implies that with the decrease of the sampling rate, the intra-cluster similarity gradually approaches the inter-cluster similarity, and the two clusters no longer exhibit significant distinguishability between them.

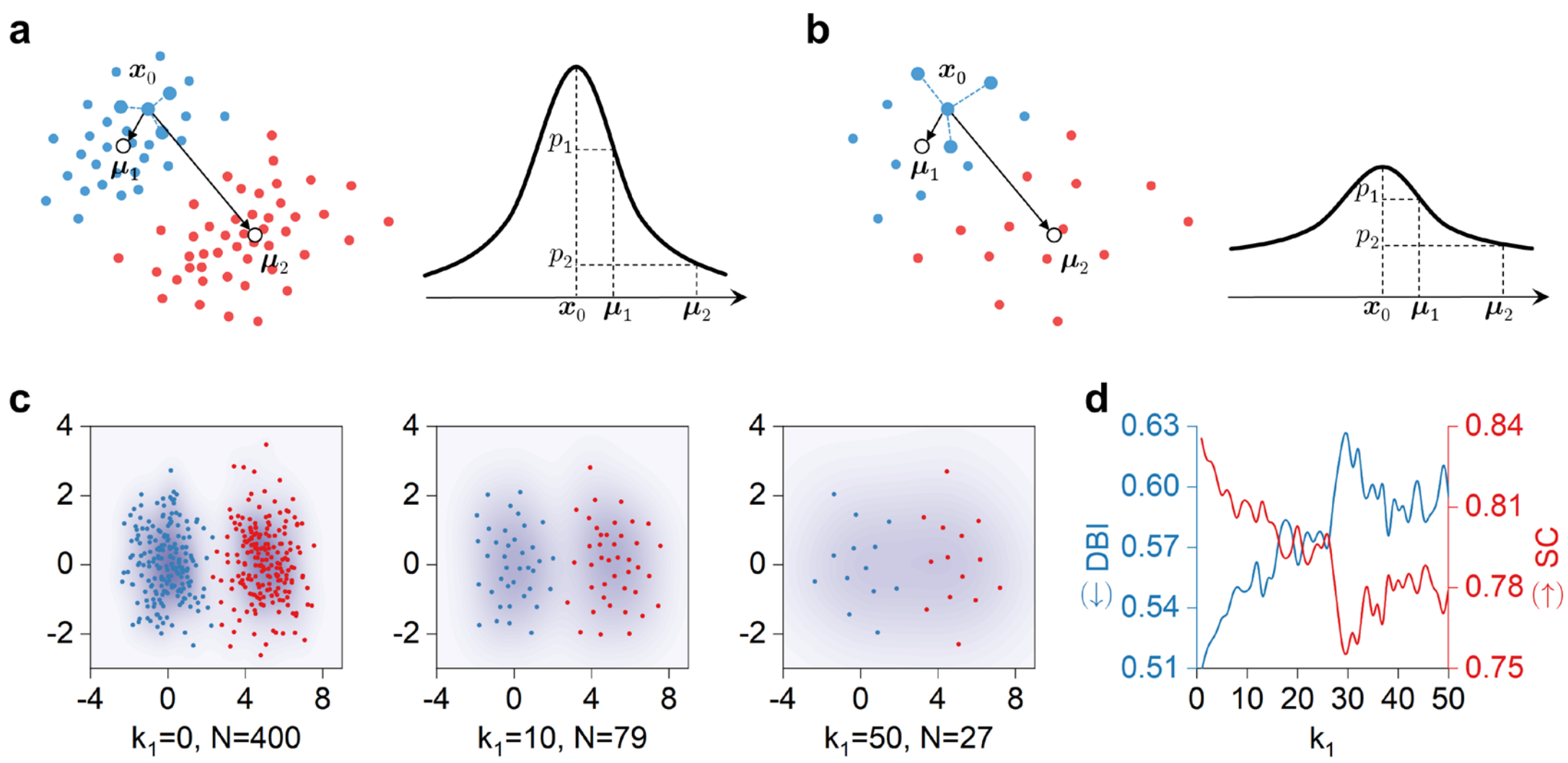

**Supplementary Fig. 9. The impact of undersampling.** (**a**), (**b**), An example on a toy dataset illustrates that undersampling leads to the convergence of intra-cluster similarity and inter-cluster similarity. (**c**), Density distribution of landmarks sampled from a synthetic dataset under different $\boldsymbol{k_1}$. (**d**), Trends of DBI and SC as the sampling rate decreases.

We further validated the undersampling problem on a 2-D synthetic dataset. It consists of two clusters and 400 points. As shown in Supplementary Fig. 9c, the two Gaussian distributions exhibit a distinct separation in the original data. While the between-cluster boundaries become fuzzy when landmarks are sparsely distributed. We used the DBI and Silhouette Coefficient (SC) scores to measure the clustering quality of this dataset under different sampling rates. The DBI overall shows a decreasing trend, while the SC exhibits an increasing trend. It indicates a decline of clustering quality and suggests that the undersampling problem weakens the distinguishability between clusters. This will lead to crowding embeddings in manifold learning.

**Supplementary Note 10: How does early aggregation mitigate the undersampling problem?**

In this paper, we design an early aggregation strategy to maintain the intra-cluster compactness and cope with the undersampling problem. Our basic idea is to enhance the pairwise similarities between the landmarks in the same cluster using their associations in the original data. The original associations are measured by the shared nearest neighbor (SNN) method, which is defined as the sum of RNN values for all common nearest neighbors of any two landmarks (equation (1)). To better understand the effect of early aggregation, we illustrate a toy dataset in Supplementary Fig. 10a, where the landmark positions and Gaussian distribution of two clusters before early aggregation are presented. There is no clear gap between the landmarks of these two clusters. It aligns with the insight discussed in Supplementary Note 9, where undersampling would make the Gaussian distributions of different clusters have larger bandwidths and become smoother, thereby weakening the distinguishability between clusters. By leveraging the associations in the original data, early aggregation can pull landmarks within the same clusters closer and slow down the increase of Gaussian bandwidth as shown in Supplementary Fig. 10b.

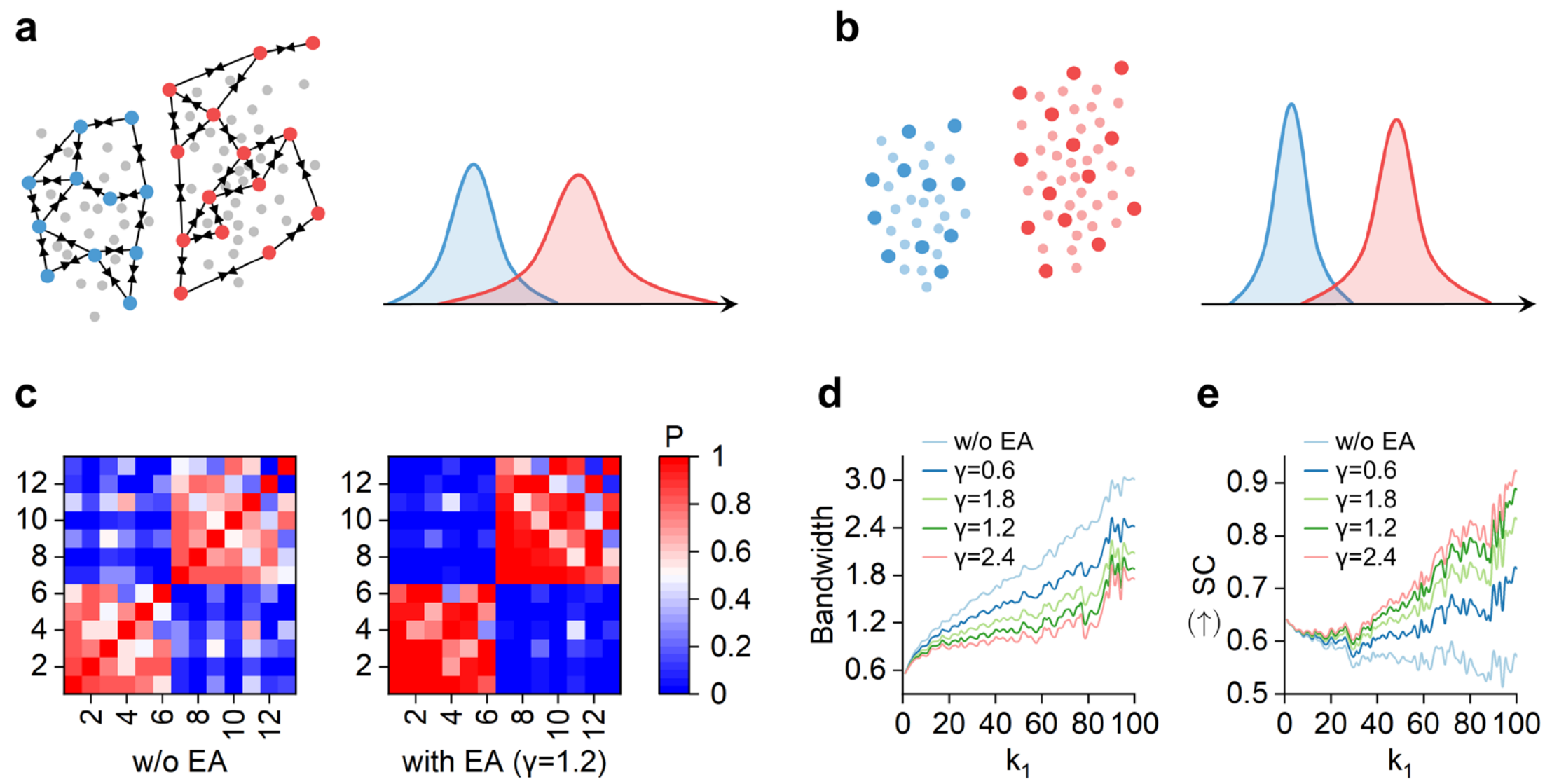

**Supplementary Fig. 10. Validation of the early aggregation method.** An example on a toy dataset that illustrates the landmark positions and Gaussian distributions of two clusters (**a**), before and (**b**), after early aggregation. (**c**), High-dimensional similarity matrices of the landmarks generated by SUDE with and without early aggregation (w/o EA). (**d**), Comparison of the average bandwidth and (**e**), SC score between w/o EA and with EA using different aggregation coefficient $\gamma$.

To validate the effectiveness of early aggregation, we used the synthetic dataset in Supplementary Fig. 9c and compared the high-dimensional similarity matrices of landmarks with ($\gamma = 1.2$) and without early aggregation (w/o EA) at $k_1 = 100$. The result in Supplementary Fig. 10c suggests that early aggregation can make the inter-cluster gap more distinct and preserve the cluster structure. We computed the average bandwidths of the landmarks generated by SUDE with and w/o EA. When EA is not applied ($\gamma = 0$), the average Gaussian kernel bandwidth grows rapidly (Supplementary Fig. 10d), which indicates a sharp decline of inter-cluster distinguishability. In contrast, applying EA can slow down the smoothing rate of the Gaussian distributions, and this suppression effect strengthens as the sampling rate decreases and $\gamma$ increases. A similar pattern is observed for the Silhouette Coefficient (SC) in Supplementary Fig. 10e, where EA can effectively reduce the loss of cluster structure as the sampling rate decreases.

However, increasing the parameter $\gamma$ will also cause a greater loss of the isometric relationship among landmarks. Therefore, in practice, we will not set a very large $\gamma$. As shown in Supplementary Fig. 10d,e, the results of $\gamma = 1.8$ and $\gamma = 2.4$ are already very close, providing less improvement compared to increasing $\gamma$ from 0 to 0.6. There is a trade-off between the preservation of the cluster structure and the isometry. In the implementation of SUDE, we set $\gamma = 1.2$ by default.

**Supplementary Note 11: How does CLLE evade dirty clusters?**

Non-landmark embedding is the final step in SUDE, aiming to utilize the relationships between landmarks and non-landmarks to determine the positions of non-landmarks in the learned low-dimensional space. Since LLE has high time efficiency and is robust to rotations, rescalings, and translations [25], we use LLE to determine the linear relationships between landmarks and non-landmarks and further reconstruct the low-dimensional embeddings of non-landmarks. However, when the clusters are very close to each other, the nearest landmarks of the non-landmarks near cluster boundaries might come from different clusters. As a result, using LLE's linear weights to reconstruct their positions may embed some of them into the between-cluster gaps and form dirty clusters. To address this issue, we develop CLLE by imposing a constraint of the nearest distance on the naïve objective function of LLE.

Suppose that we embed a non-landmark $\boldsymbol{x}$ into 2-D space, CLLE first calculates the corresponding linear weights $w_1$, $w_2$, and $w_3$ of its 3-NN, and then estimates the distance $d_1$ from its nearest landmark $\boldsymbol{y}_1$ in low-dimensional space. Let us denote the low-dimensional position determined by LLE as $\boldsymbol{y}' = \sum_{i=1}^{3} w_i \boldsymbol{y}_i$. There are two cases. When $\|\boldsymbol{y}_1 - \boldsymbol{y}'\| \leq d_1$, that means $\boldsymbol{y}'$ is outside the circle centered at $\boldsymbol{y}_1$ with radius $d_1$. In this case, the low-dimensional coordinates of $\boldsymbol{y}$ can be obtained as presented in Supplementary Note 4

$$\boldsymbol{y} = \boldsymbol{y}_1 - d_1 \frac{\boldsymbol{y}_1 - \boldsymbol{y}'}{\|\boldsymbol{y}_1 - \boldsymbol{y}'\|}$$

Supplementary 11a illustrates that the essence of CLLE aims to move $\boldsymbol{y}'$ along the line determined by $\boldsymbol{y}_1$ and $\boldsymbol{y}'$ onto the circle centered at $\boldsymbol{y}_1$ with radius $d_1$ when $\boldsymbol{y}'$ is outside the circle. It is as tying a rope of length $d_1$ to the $\boldsymbol{y}_1$ end to keep $\boldsymbol{y}$ from moving too far away from $\boldsymbol{y}_1$. This shortens the distance to its nearest landmark and prevents it from being misplaced into between-cluster gaps or incorrect clusters. While, when $\boldsymbol{y}'$ falls inside the circle, we push $\boldsymbol{y}'$ onto this circle to preserve the nearest distance constraint. By integrating local linear representations and distance relationships, errors caused by using only one of them can be reduced. Besides, it is interesting that we found another extrema when $\lambda > 1$ in Supplementary Note 4. This extrema corresponds the other point where the line determined by $\boldsymbol{y}_1$ and $\boldsymbol{y}'$ intersects the circle ($\boldsymbol{y}''$ in Supplementary 11a), which is not the position we want due to its longer distance from $\boldsymbol{y}_1$.

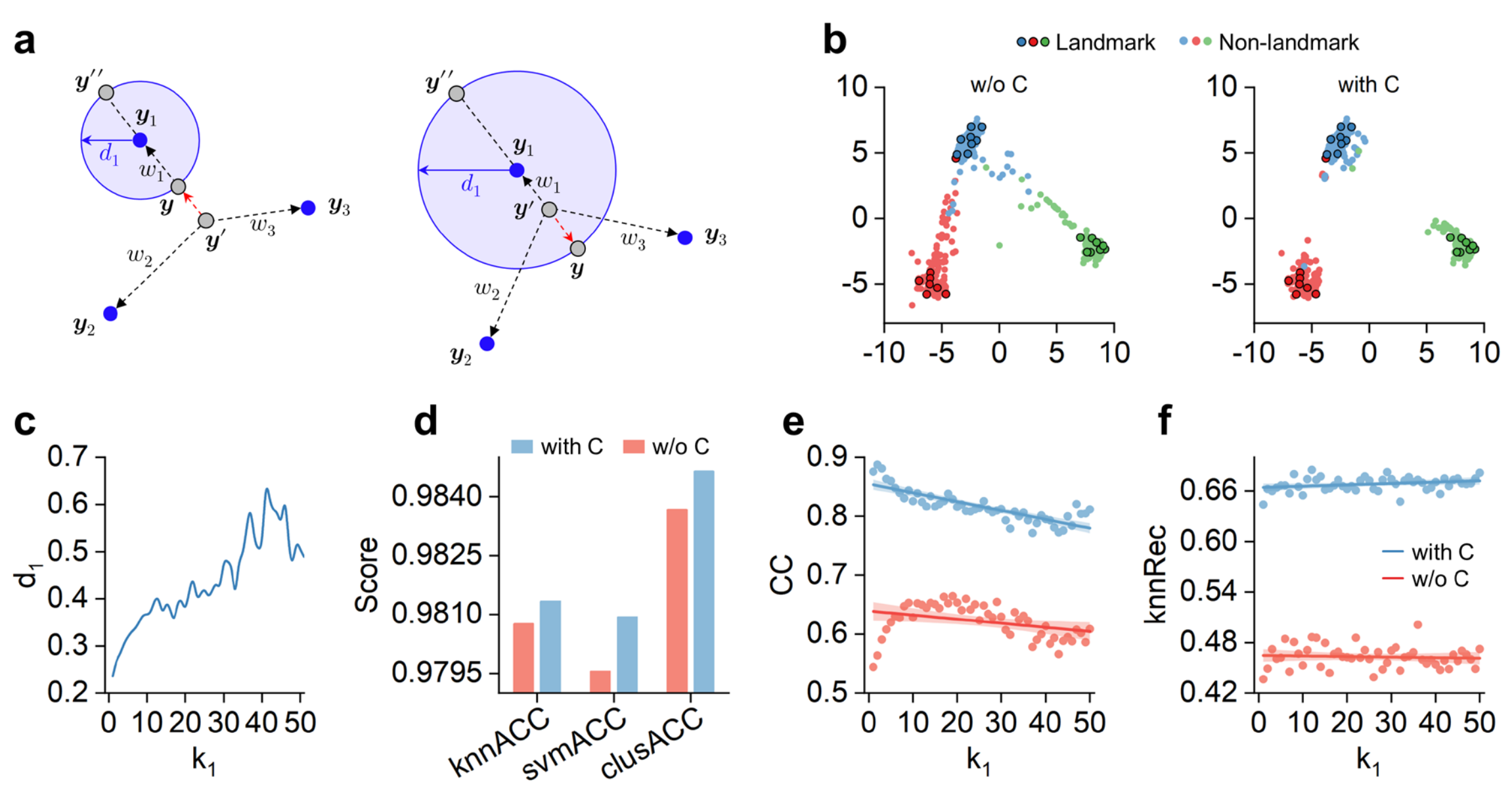

**Supplementary Fig. 11. Validation of the constrained locally linear embedding method.** (**a**), Illustration of how the nearest distance constraint works under two cases. (**b**), Comparison of embeddings generated by SUDE jointly with and without the nearest distance constraint (w/o C) in CLLE when $\boldsymbol{k_1} = \mathbf{50}$. (**c**), Trends of the nearest distance $\boldsymbol{d_1}$ with the increase of $\boldsymbol{k_1}$. (**d**), The average knnACC, svmACC, and clusACC scores by varying $\boldsymbol{k_1}$ from 1 to 50. (**e**), CC, and (**f**), knnRec scores under different $\boldsymbol{k_1}$.

We validated the effectiveness of CLLE on a synthetic 3-D dataset, where 600 points constitute three nearby clusters with clear gaps. Supplementary Fig. 11b presents SUDE embeddings with and without the nearest distance constraint (w/o C) in CLLE when $k_1 = 50$. Under the condition of low sampling rate and sparse landmarks, non-landmarks located near the cluster boundaries are prone to find the nearest landmarks from other clusters, and would be misplaced into the inter-cluster gaps by reconstructing the low-dimensional positions only through linear weights. This causes dirty clusters and outliers. By imposing the nearest distance constraint, non-landmarks can be pulled closer to their nearest landmarks, thereby creating more pronounced separation between clusters. As the sampling rate decreases, $d_1$ exhibits an upward trend in Supplementary Fig. 11c. It indicates that non-landmarks are farther away from their nearest landmarks when landmarks become sparse. The comparison of classification and clustering accuracies in Supplementary Fig. 11d verifies the effectiveness of the distance constraint in preservation of cluster structure. The global structure gradually deteriorates with decreasing sampling rates (Supplementary Fig. 11e), while the distance constraint can effectively enhance its preservation and obtain a higher CC score. This is because it ensures the pairwise distance relationships between non-landmarks and their nearest landmarks across the high-dimensional and embedding spaces. As shown in Supplementary Fig. 11f, CLLE can also enhance the local structure preservation, since this constraint makes the non-landmarks closer to their nearest landmarks when the initial separation distance is sufficiently large.

**Supplementary Note 12: Comparative analysis of UMAP and SUDE embeddings in mouse retina single-cell transcriptomics**

We selected a scRNA-seq dataset from wild-type (WT) mouse retinas and validated that SUDE embedding better aligns with biological significance compared to UMAP. Fig. 5a in the main text presents the visualization of UMAP and SUDE embeddings. It can be observed that UMAP over-clusters cone bipolar cells and amacrine cells into multiple fine-grained subclusters, while SUDE identifies them as two whole clusters identical with the ground-truth cell annotations. Leveraging the CDC clustering algorithm, we further quantified that SUDE can better preserve the cluster integrity as shown in Fig. 5c. In this note, we aim to investigate which method can generate clusters better correspond to biology within these retinal cell populations from three perspectives.

- **The cluster structure in the SUDE embedding is more consistent with the Louvain clusters than UMAP embedding**

To validate the consistency between the embeddings and the original high-dimensional data, we selected the Louvain algorithm to explore the true cluster structure in the high-dimensional data. Louvain clustering is the default clustering algorithm in the Seurat software and is also a benchmark cell annotation method in bioinformatics. We first preprocessed the WT dataset and conducted the Louvain algorithm on 50 PCs by varying its parameter $resolution$ from 0.01 to 1 with an interval of 0.01. The best SC score (0.62) of Louvain achieved when $resolution = 0.02$ as shown in Supplementary 12a, and the corresponding results are illustrated in Supplementary 12d. As the CDC clustering algorithm can handle data with density heterogeneity and weak connectivity, we then used it to identify the cluster structures of the 2-D UMAP and SUDE embeddings by specifying $k = 20$ and varying $ratio$ from 0.80 to 0.99 with an interval of 0.01. Compared to both the true cell types and Louvain results, CDC yielded significantly higher ACC scores on the SUDE embedding than UMAP embedding (Supplementary 12b). The Sankey plot in Supplementary 12c reports that the best CDC results on the SUDE embedding are more consistent with the Louvain clusters compared to UMAP embedding. Louvain identifies 7 clusters from the WT dataset in Supplementary 12d, while CDC detects 9 and 26 clusters from the SUDE and UMAP embeddings Supplementary 12e, respectively. The Louvain results indicate that clusters of cone bipolar cells and amacrine cells have no distinct substructures in the original data. UMAP over-clusters cone bipolar cells and amacrine cells, while SUDE preserves the integrity of these cell clusters.

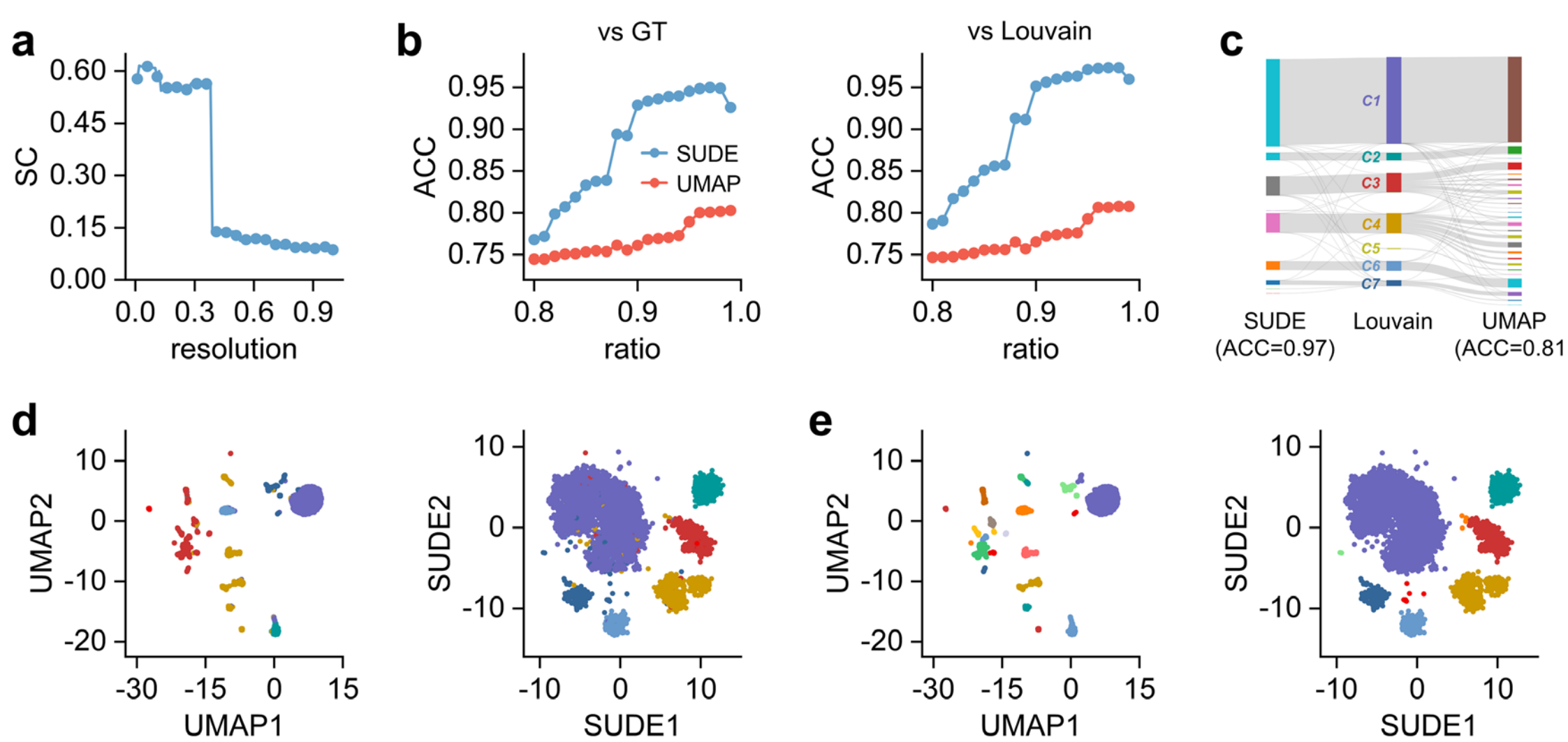

**Supplementary Fig. 12. Evaluation of the cluster structures in UMAP and SUDE embeddings.** (**a**), SC scores of the Louvain clusters under different $resolution$. (**b**), ACC scores obtained by the CDC algorithms with different $ratio$ on the UMAP and SUDE embeddings. (**c**), A Sankey diagram shows the match between the best CDC and Louvain results. (**d**), The best Louvain clustering results ($resolution = 0.02$) on 50 PCs. (**e**), The best CDC clustering results ($ratio = 0.97/0.99$) on the UMAP and SUDE embeddings.

- **Differential gene expression analysis reveals that SUDE preserves the cluster integrity of cone bipolar cells and amacrine cells in WT retinas**

Differentially expressed genes (DEGs) play a critical role in determining and validating the biological relevance of clusters in genomic or single-cell analysis [26]. We employed the *FindMarkers* function in the Seurat package to identify DEGs of the six major cell types (rods, rod bipolar cells, cones, cone bipolar cells, amacrine cells, Muller glia) in WT. The parameters *min.pct* and *logfc.threshold* were set to 0.1 and 0.5, respectively. *min.pct* requires that a gene is expressed in at least a certain percentage of cells in either of the two groups being compared, while *logfc.threshold* refers to the minimum absolute log-scale fold change in average expression between two groups. *FindMarkers* would output five values, $p_val$, $avg_log2FC$, *pct.1*, *pct.2*, and $p_val_adj$ for each gene. $p_val$ and $p_val_adj$ indicates the significance level. *pct.1* and *pct.2* refers to the percentage of cells where the gene is detected in the first and second groups, respectively. Higher $avg_log2FC$ values indicate that the gene is more highly expressed in the first group. To determine the DEGs, we adopted a z-score-based indicator as

$$I = scale(p_val) - scale(avg_log2FC) - scale(pct.1) + scale(pct.2) + scale(p_val_adj)$$

where $scale(\cdot)$ is the function for standardizing data and converting each original value into a z-score. The top 5 genes with the smallest $I$ values for each cell type are listed below:

| **Rods** | *p_val* | *avg_log2FC* | *pct.1* | *pct.2* | *p_val_adj* |
|---|---|---|---|---|---|
| Cnga1 | 0.00E+00 | 4.970419 | 0.647 | 0.098 | 0.00E+00 |
| Guca1b | 0.00E+00 | 4.191949 | 0.587 | 0.148 | 0.00E+00 |
| Nr2e3 | 8.85E-301 | 4.344794 | 0.467 | 0.107 | 2.48E-296 |
| Nrl | 0.00E+00 | 4.192116 | 0.502 | 0.119 | 0.00E+00 |
| Samd11 | 3.62E-213 | 4.647729 | 0.354 | 0.07 | 1.01E-208 |
| **Rod bipolar cells** | *p_val* | *avg_log2FC* | *pct.1* | *pct.2* | *p_val_adj* |
| Sebox | 0.00E+00 | 5.524008 | 0.793 | 0.022 | 0.00E+00 |
| Tpbg | 0.00E+00 | 6.102859 | 0.555 | 0.012 | 0.00E+00 |
| Gm29678 | 3.63E-279 | 7.771408 | 0.198 | 0.002 | 1.02E-274 |
| Car8 | 0.00E+00 | 5.284311 | 0.707 | 0.026 | 0.00E+00 |
| Cacna2d3 | 0.00E+00 | 4.858698 | 0.793 | 0.045 | 0.00E+00 |
| **Cones** | *p_val* | *avg_log2FC* | *pct.1* | *pct.2* | *p_val_adj* |
| Arr3 | 0 | 6.7056 | 0.851 | 0.027 | 0 |
| Pde6h | 0 | 6.399393 | 0.985 | 0.119 | 0 |
| Opn1sw | 0 | 7.777574 | 0.676 | 0.059 | 0 |
| Opn1mw | 0 | 7.248165 | 0.676 | 0.018 | 0 |
| Gnat2 | 0 | 5.976401 | 0.887 | 0.048 | 0 |
| **Cone bipolar cells** | *p_val* | *avg_log2FC* | *pct.1* | *pct.2* | *p_val_adj* |
| Scgn | 0.00E+00 | 5.171301 | 0.547 | 0.016 | 0.00E+00 |
| Gsg1 | 0.00E+00 | 5.78366 | 0.34 | 0.011 | 0.00E+00 |
| Epha3 | 8.71E-157 | 6.679931 | 0.122 | 0.002 | 2.44E-152 |
| A730046J19Rik | 0.00E+00 | 5.564237 | 0.282 | 0.007 | 0.00E+00 |
| BC030499 | 0.00E+00 | 3.867156 | 0.726 | 0.087 | 0.00E+00 |
| **Amacrine cells** | *p_val* | *avg_log2FC* | *pct.1* | *pct.2* | *p_val_adj* |
| Slc32a1 | 0 | 5.970451 | 0.543 | 0.008 | 0 |
| Pax6 | 0 | 4.076734 | 0.863 | 0.047 | 0 |
| Snhg11 | 0 | 4.315827 | 0.893 | 0.110 | 0 |
| Elavl3 | 0 | 5.000974 | 0.493 | 0.012 | 0 |
| Synpr | 0 | 3.9333 | 0.737 | 0.050 | 0 |
| **Muller glia** | *p_val* | *avg_log2FC* | *pct.1* | *pct.2* | *p_val_adj* |
| Spc25 | 0 | 8.00216 | 0.818 | 0.009 | 0 |
| Glul | 0 | 6.567159 | 0.96 | 0.091 | 0 |
| Rlbp1 | 0 | 7.380033 | 0.727 | 0.008 | 0 |
| Aqp4 | 0 | 9.055512 | 0.511 | 0.001 | 0 |
| Pdpn | 0 | 7.865628 | 0.614 | 0.004 | 0 |

Supplementary Fig. 13a displays an expression heatmap of the top 5 DEGs for each cell type. The selected DEGs effectively distinguish between different cell types. Furthermore, the violin plots in Supplementary Fig. 13b demonstrate that the top-ranked DEGs (Cnga1, Sebox, Arr3, Scgn, Slc32a1, and Spc25) of the six major cell types is highly expressed within its corresponding cell population but exhibits minimal expression in other cell types, confirming the accuracy of our DEG selection. Meanwhile, the significant differential expression across these six cell types in the selected DEGs demonstrates the presence of six distinct cluster structures in the high-dimensional space, which aligns consistently with the results obtained from Louvain clustering in Supplementary Fig. 12d.

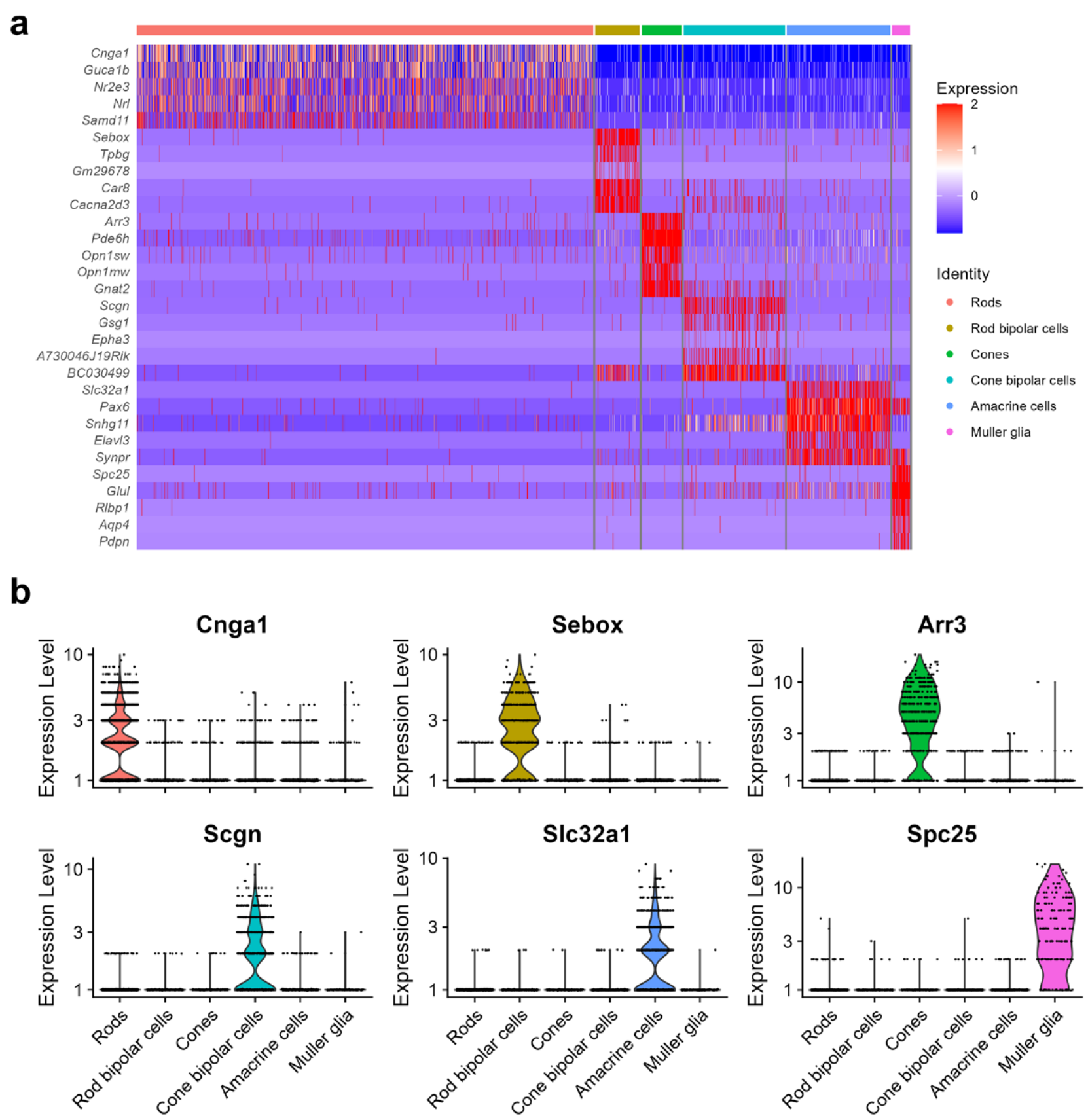

**Supplementary Fig. 13. DEGs of the six major cell types in WT retinas.** (**a**), An expression heatmap for the top 5 markers of each cell type. (**b**), Violin plots show the expression probability distributions of across different cell types.

We illustrated the expression patterns of the above six DEGs in Supplementary Fig. 14. The cone bipolar cells and amacrine cells with high expression levels of Scgn and Slc32a1 on the SUDE plot exhibit a more concentrated distribution than the UMAP plot. This indicates that SUDE can better preserve the cellular similarity reflected by DEGs in low-dimensional embedding.

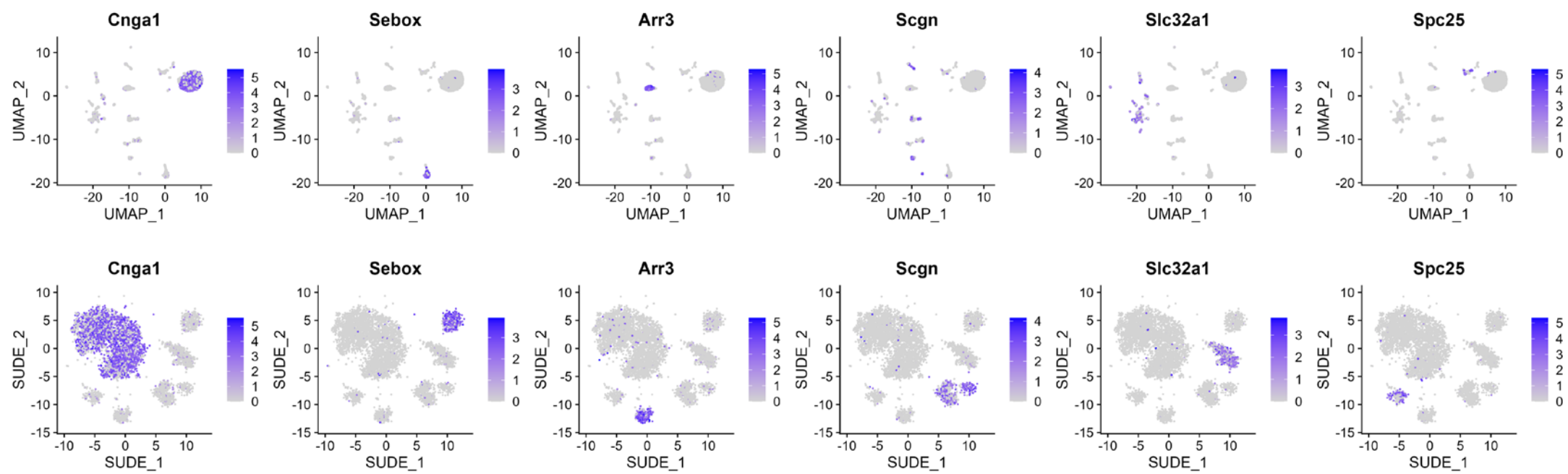

**Supplementary Fig. 14. Expression patterns of the six DEGs on the UMAP and SUDE plots.**

We further extracted cone bipolar cells and amacrine cells from the original dataset and detected their significant DEGs. We first applied the Louvain algorithm to these two cell types separately by varying $resolution$ from 0 to 1 with an interval of 0.01. Then, we selected the best clustering results using SC score and identified the significant DEGs for each cluster. To find out all potential significant DEGs, we defined the following criterion: *pct.1 > 0.5*, *pct.2* < 0.1, and *avg_log2FC* > 0.5 [27]. This criterion ensures that the gene is expressed in at least 50% of cells in the first group, while being expressed in less than 10% of cells in the second group. Meanwhile, the log fold-change in average gene expression between the two groups must be greater than 0.5. This criterion is not overly stringent and is designed to identify as many DEGs as possible. We regard the clusters containing more than 20 significant DEGs as potential cell subtypes. This threshold is empirically setting, which is lenient. For non-rare cell types, the number of significant DEGs usually reaches dozens or even hundreds [28]. It is important to note that the biological significance of a cell cluster cannot be solely determined by the quantity of DEGs. Functional enrichment analysis can provide a complementary approach to validate and interpret the biological relevance of these DEGs.

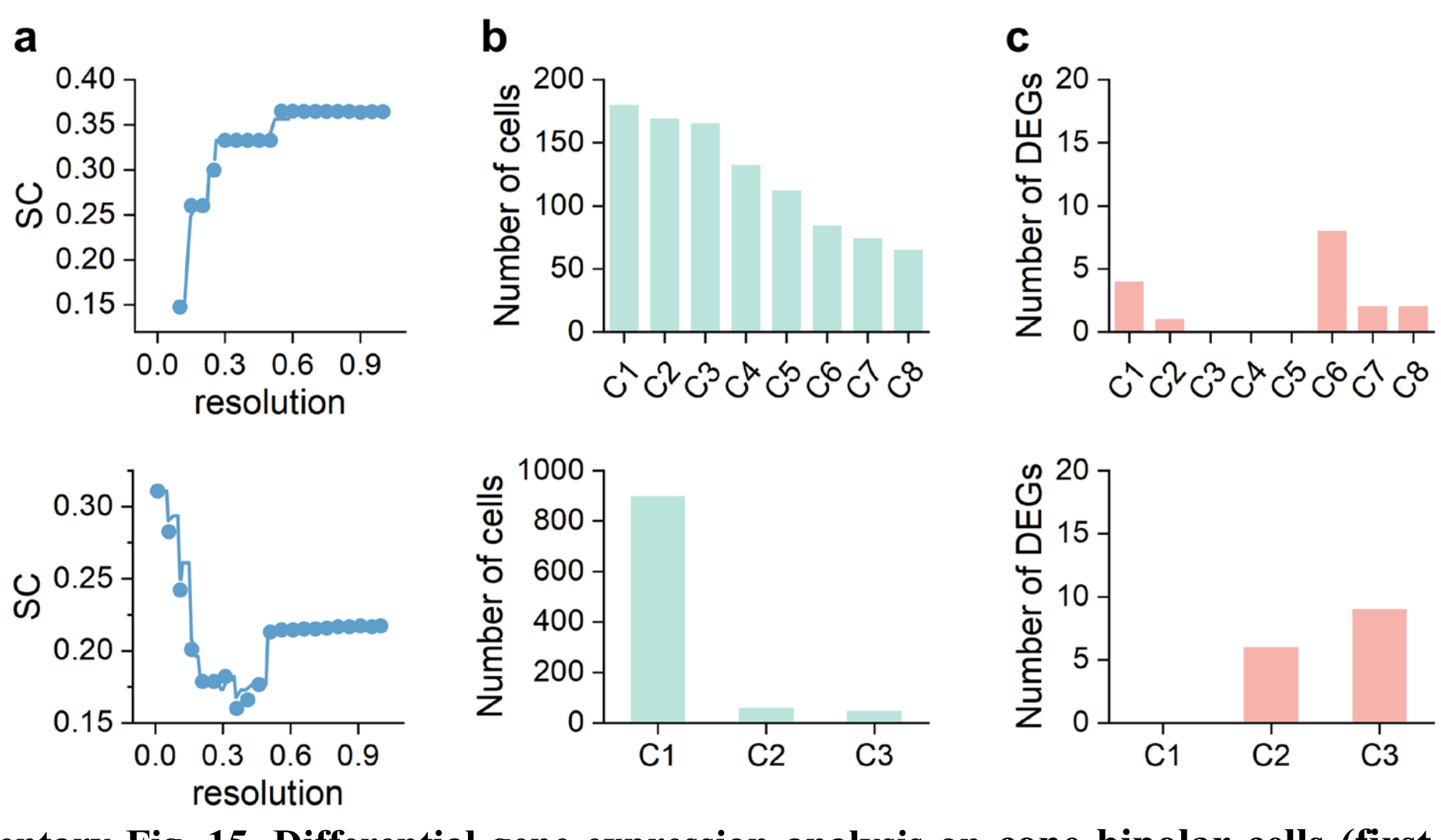

**Supplementary Fig. 15. Differential gene expression analysis on cone bipolar cells (first row) and amacrine cells (second row).** (**a**), SC scores of the Louvain results under different $resolution$. The numbers of (**b**), cells and (**c**), significant DEGs of each cluster in the best Louvain results.

As shown in Supplementary Fig. 15, the SC scores of Louvain on both cone bipolar cells and amacrine cells are much lower than the SC scores (0.62) from the entire dataset (Supplementary Fig. 12a). It indicates that these cells are densely distributed in the high-dimensional original data without distinct gaps. Notably, the Louvain algorithm identifies cone bipolar cells as a single cluster at $resolution < 0.1$, thus yielding no SC score. We selected the best clustering results for significant DEG detection, which are achieved when $resolution = 0.6$ in cone bipolar cells and $resolution = 0.01$ in amacrine cells. Louvain identifies eight clusters in cone bipolar cells and three clusters in amacrine cells, and all of them contain less than 10 significant DEGs. Although the cluster C6 of cone bipolar cells has eight significant DEGs and C3 in amacrine cells has nine DEGs, they contain a small number of cells. Even when considering them as independent subclusters, the number of clusters for each cell type is significantly smaller than the number of subclusters in the UMAP embedding (9 and 14 subclusters of cone bipolar cells and amacrine cells are identified by CDC from the UMAP embedding). This suggests no significant cell subtypes exist for these two cell types in the WT dataset.

- **UMAP embedding over-clustered cone bipolar cells and amacrine cells**

We adopted a model-based hypothesis testing approach, single-cell Significance of Hierarchical Clustering (sc-SHC) [29], to evaluate the statistical significance of the subclusters in UMAP embedding. In this approach, it is not recommended to merge the clusters if p-value is greater than 0.05. The best CDC clustering results of cone bipolar cells and amacrine cells in Supplementary Fig. 12e were used, where these two cell types contain 9 and 14 clusters on the UMAP embedding. The test of sc-SHC suggests merging these subclusters as a whole one by reporting the p-values of cone bipolar cells and amacrine cells as 1.0 and 0.8, respectively. Therefore, it is inappropriate to partition these two cell types into many subclusters; while identifying them as a single cluster aligns better with the high-dimensional structure of cells in the WT dataset.

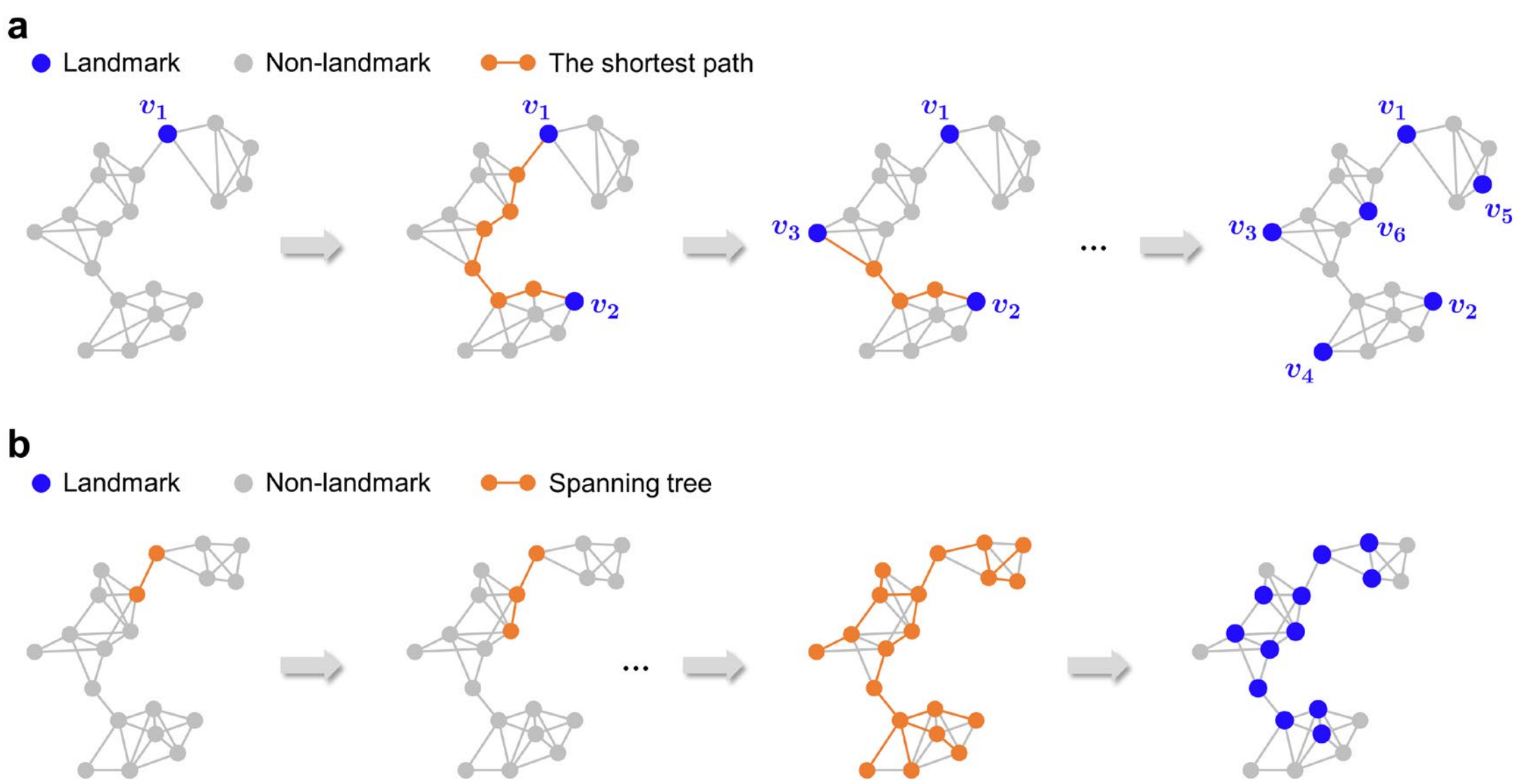

**Supplementary Fig. 16. Workflows of two landmark sampling algorithms.** (**a**), The pruned spanning tree (PST) algorithm aims to construct a random spanning tree in the ε-neighborhood graph. It first randomly selects an edge as the initial tree, and then recursively adds edges between tree nodes and unassigned nodes and makes sure that no cycles form in the tree. This procedure ends until all nodes are incorporated into the tree. Finally, PST selects the points with high vertex degree as the ultimate landmarks. (**b**), The maximum-minimum sampling (MMS) algorithm first constructs the KNN graph, and randomly selects a point as the initial landmark set. In each loop, the point having the maximum distance to the current landmark set would be chose as a landmark, that is, maximizing the minimum distance to any of the points in the current set. The shortest path distance is used to measure the point-pair distance, and the procedure ends until the required number of landmarks is reached.

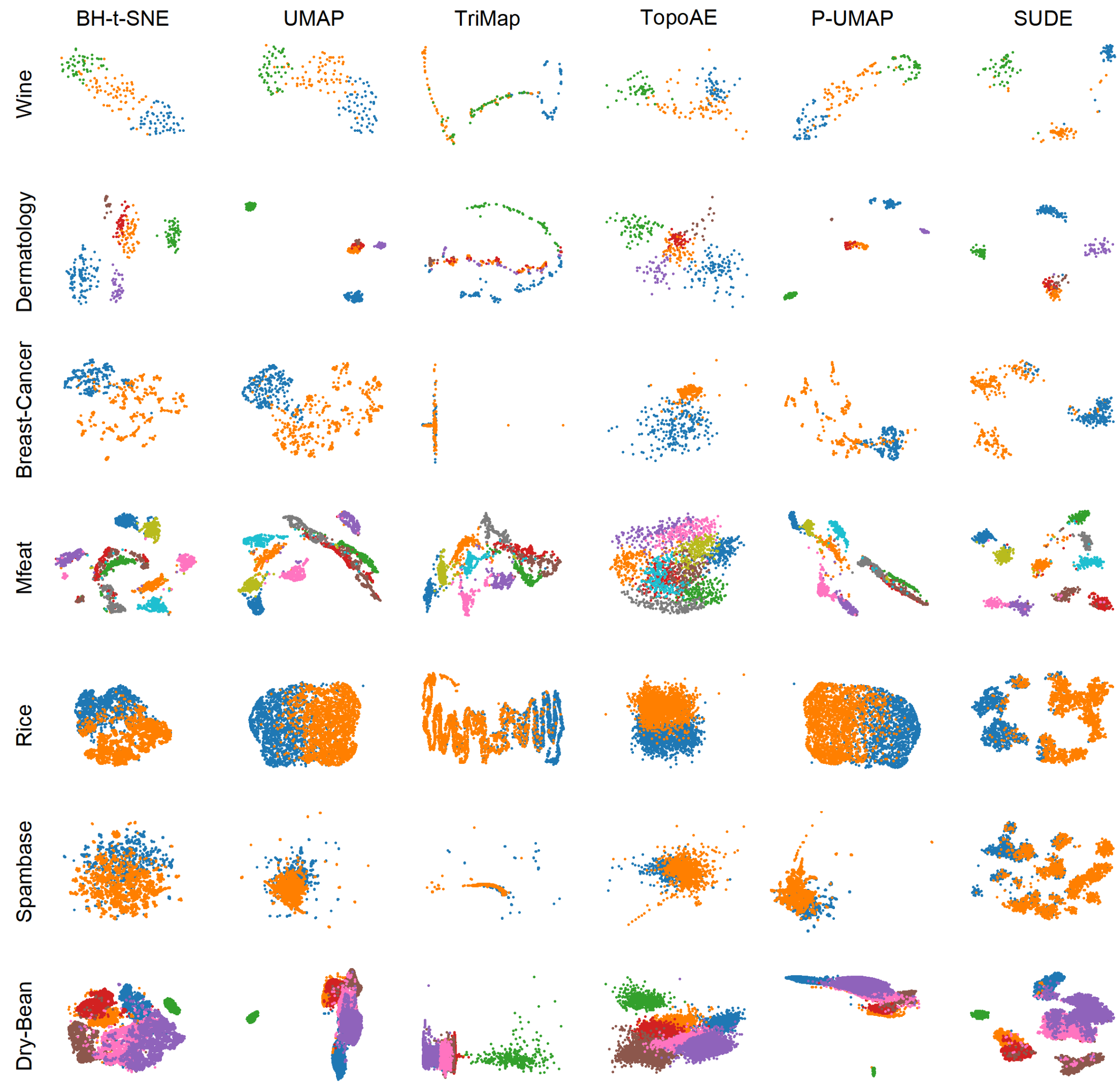

**Supplementary Fig. 17. Comparison of embeddings from seven real-world datasets using BH-t-SNE, UMAP, TriMap, TopoAE, P-UMAP, and SUDE.**

* BH-t-SNE, UMAP, and P-UMAP obtain similar embeddings, where the clusters can be separated clearly. When the clusters exhibit clear gaps in the original high-dimensional data, like Dermatology and Dry-Bean, UMAP and P-UMAP can achieve more distinct cluster separation compared to BH-t-SNE. TriMap generates curved embeddings from Wine and Dermatology, and cannot separate the clusters of Breast-Cancer and Spambase, leading to much lower knnACC, svmACC, and clusACC scores (Fig. 4a). TopoAE preserves a more accurate layout of clusters and achieves the highest CC score on most datasets. However, there is a crowding problem with no distinct gaps between clusters in the TopoAE embeddings on Wine, Dermatology, and Mfeat, which results in a poor performance with respect to the classification and clustering accuracy across these datasets. SUDE can generate more distinct inter-cluster discrimination and separate weakly connected clusters using a heavy-tailed logarithmic similarity kernel and an early aggregation strategy. Taking Dry-Bean as an example, the first five baselines merge the six clusters as a whole one, while SUDE better separates them and identifies them as four subclusters. It obtains the highest knnACC, svmACC, and clusACC scores on Wine, Breast-Cancer, Mfeat, and Dry-Bean.

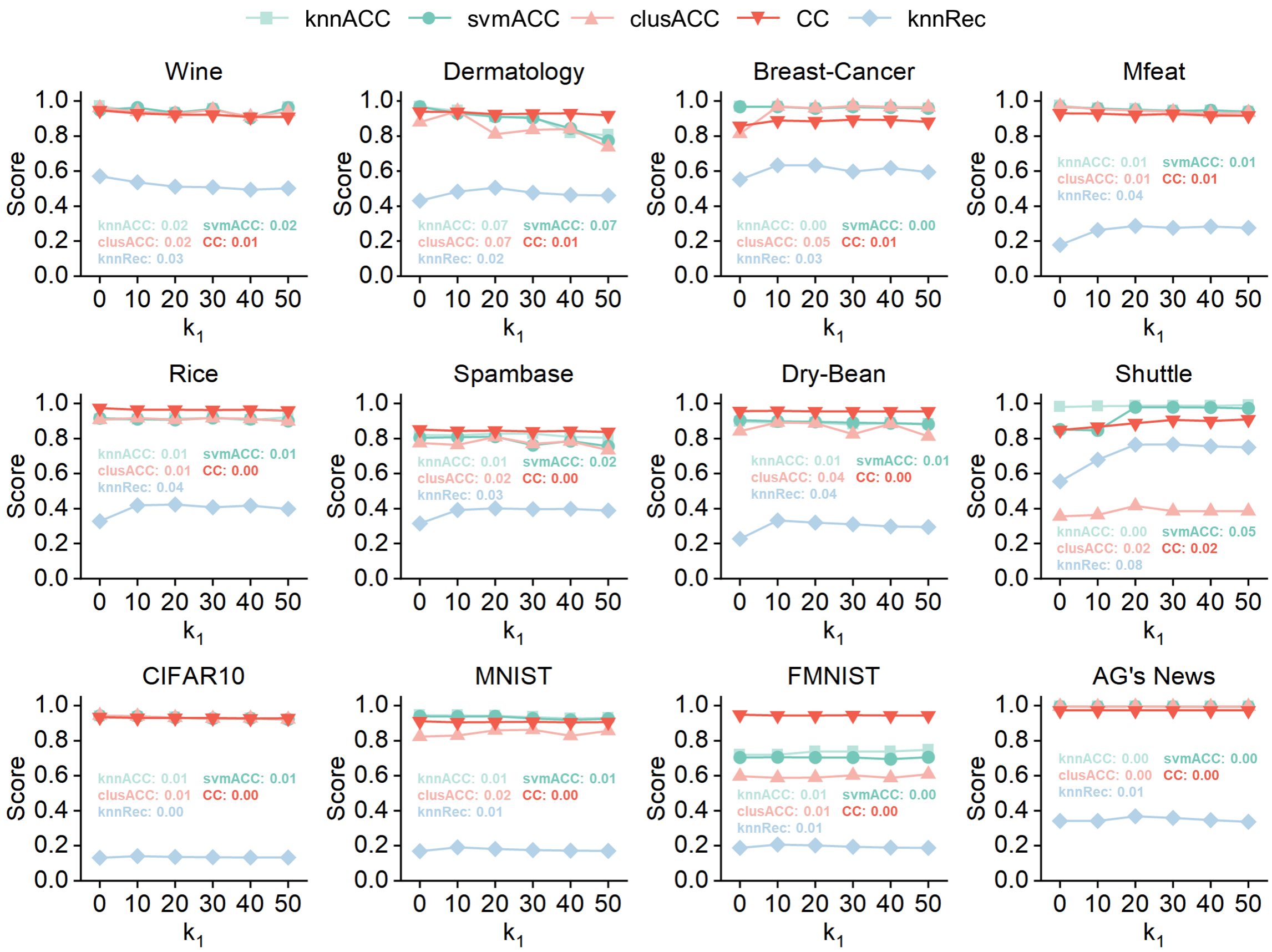

**Supplementary Fig. 18. Five evaluation metrics of SUDE under different $k_1$ on 12 real-world datasets.**

* We used knnACC, svmACC, clusACC, CC, and knnRec scores to evaluate the quality of SUDE embedding with $k_1$ varying from 0 to 50. SUDE exhibits notable sampling robustness on most datasets, with all standard deviations no greater than 0.07. The preservation of global structure reported by CC score is quite stable as the sampling rate decreases, since the PPS algorithm can generate uniform landmarks and better preserve the structure of the entire data (Details can be found in Supplementary Notes 6 and 7). In addition, we observe that knnRec score under $k_1 = 10$ outperforms $k_1 = 0$ on most datasets. This is because SUDE activates CLLE after landmark sampling, and CLLE leverages the nearest distance constraint to better preserve the local structure of data. However, when landmarks become extremely sparse, CLLE is insufficient to fix the local structure, which leads to a decline in knnRec score.

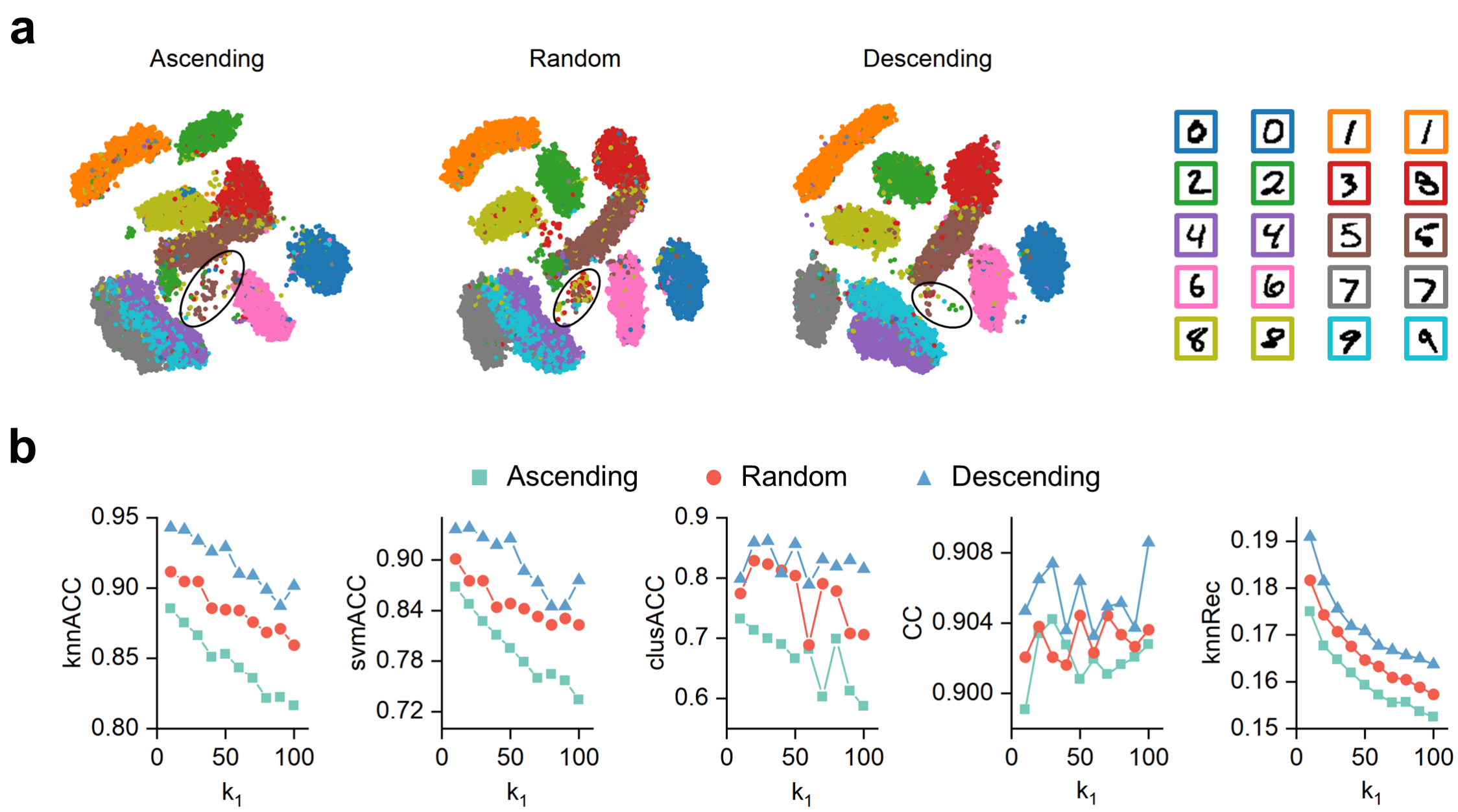

**Supplementary Fig. 19. Performance comparison among SUDE using three RNN sequences in the PPS algorithm on MNIST.** (**a**), Embeddings ($\boldsymbol{k_1} = \mathbf{50}$), and (**b**), five evaluation metrics of PPS using ascending, random and descending RNN sequences under different $\boldsymbol{k_1}$.

* Ascending and random ordering cannot separate digits "7" from the clump of digits "4" and "9" and make partial digits "2" connect with digits "5". Since ascending and random ordering tend to select more boundary points or outliers as landmarks, they embed more non-landmark points in the between-cluster gaps, resulting in more outliers in the embeddings (black circles). In terms of the evaluation metrics, descending ordering demonstrates advantages over ascending and random ordering with respect to classification/clustering accuracy and global/local structure preservation.

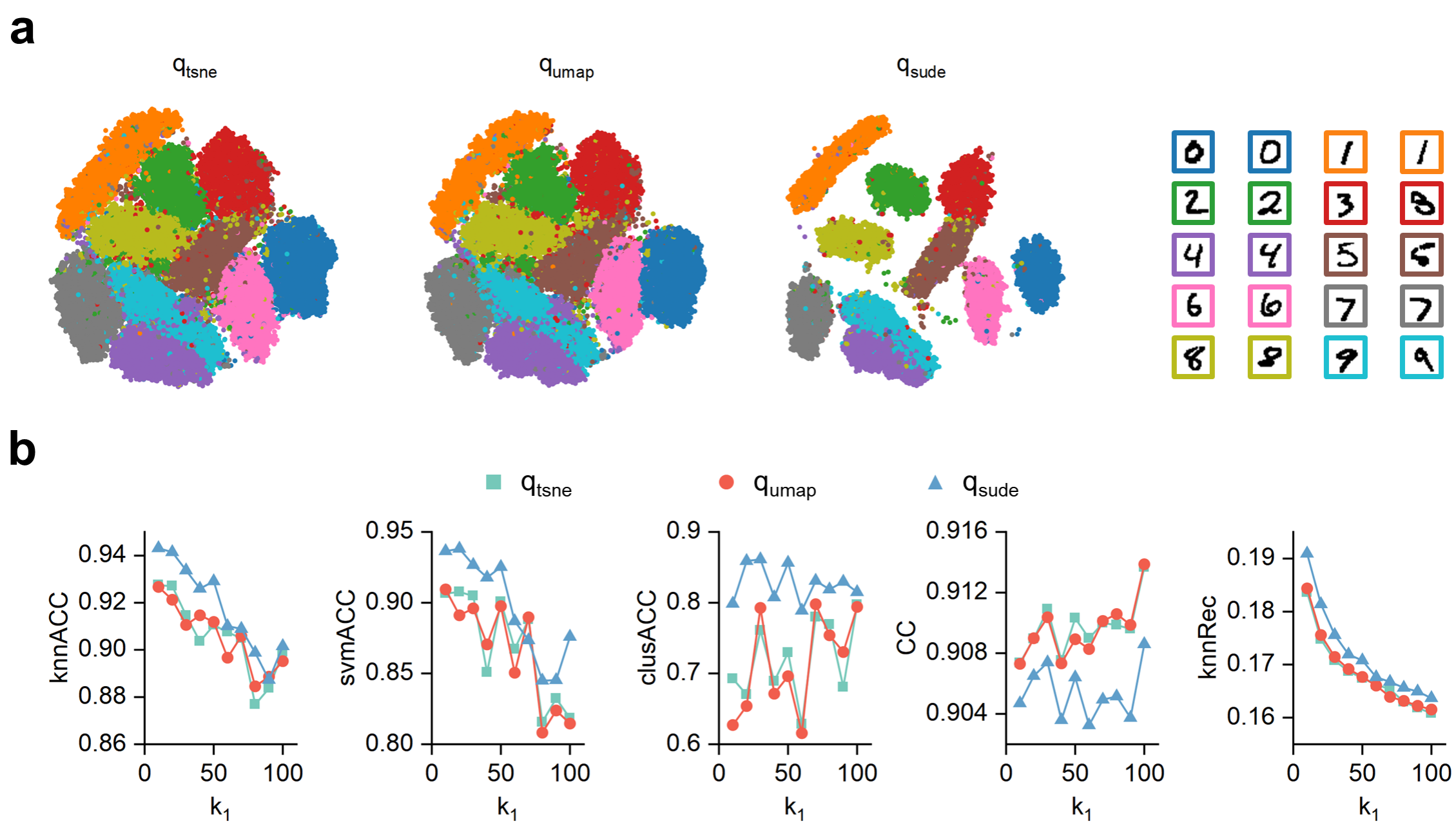

**Supplementary Fig. 20. Performance comparison among SUDE jointly with three low-dimensional probability functions on MNIST.** (**a**), Embeddings ($k_1 = 50$), and (**b**), five evaluation metrics of SUDE using $q_{tsne}=(1+\|\boldsymbol{y}_i-\boldsymbol{y}_j\|^2)^{-1}$, $q_{umap}=(1+a\|\boldsymbol{y}_i-\boldsymbol{y}_j\|^{2b})^{-1}$, and $q_{sude}=\left(1+\log(1+\|\boldsymbol{y}_i-\boldsymbol{y}_j\|^2)\right)^{-1}$ under different $k_1$. We set $a = 1.577$ and $b = 0.895$ for UMAP by default and ensure all other configurations to be the same.

* $q_{tsne}$ and $q_{umap}$ cannot generate distinct inter-cluster separation within 50 epochs (by default in SUDE implementation) and yield lower classification/clustering accuracy and worse local structure compared to $q_{sude}$. Nonetheless, $q_{tsne}$ and $q_{umap}$ can better preserve the global structure with higher CC scores. This reveals an inherent trade-off between cluster separation and global structure preservation in dimension reduction.

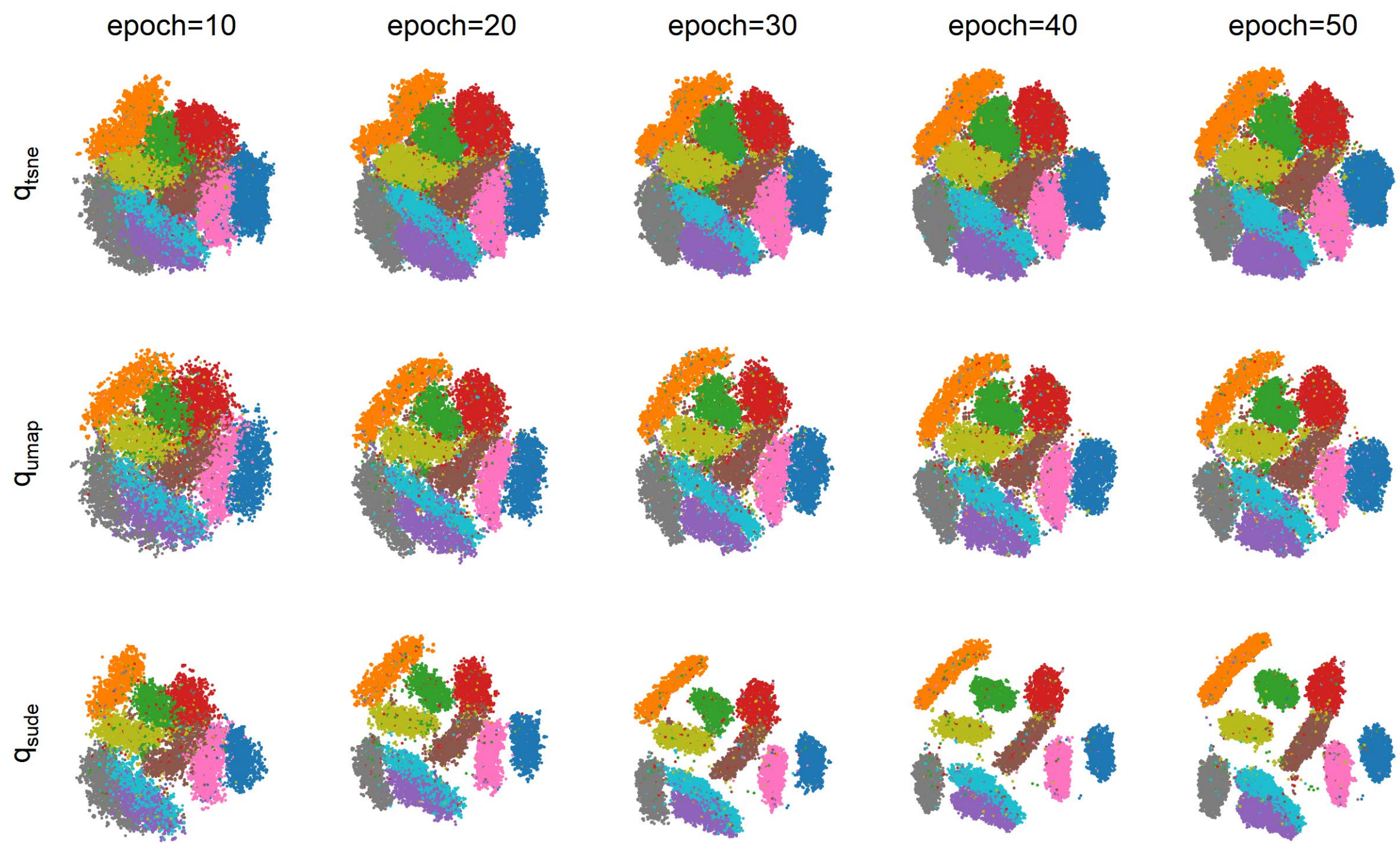

**Supplementary Fig. 21. Embeddings of SUDE using three low-dimensional probability functions under different epochs on MNIST.**

* The results show the effectiveness of the logarithmic probability function. As a heavy-tailed kernel function, $q_{sude}$ is more effective than $q_{tsne}$ and $q_{umap}$ to compress clusters. It can accelerate the convergence speed of embedding learning and yield a promising layout within 30 epochs. (Details can be seen in Supplementary Note 8).

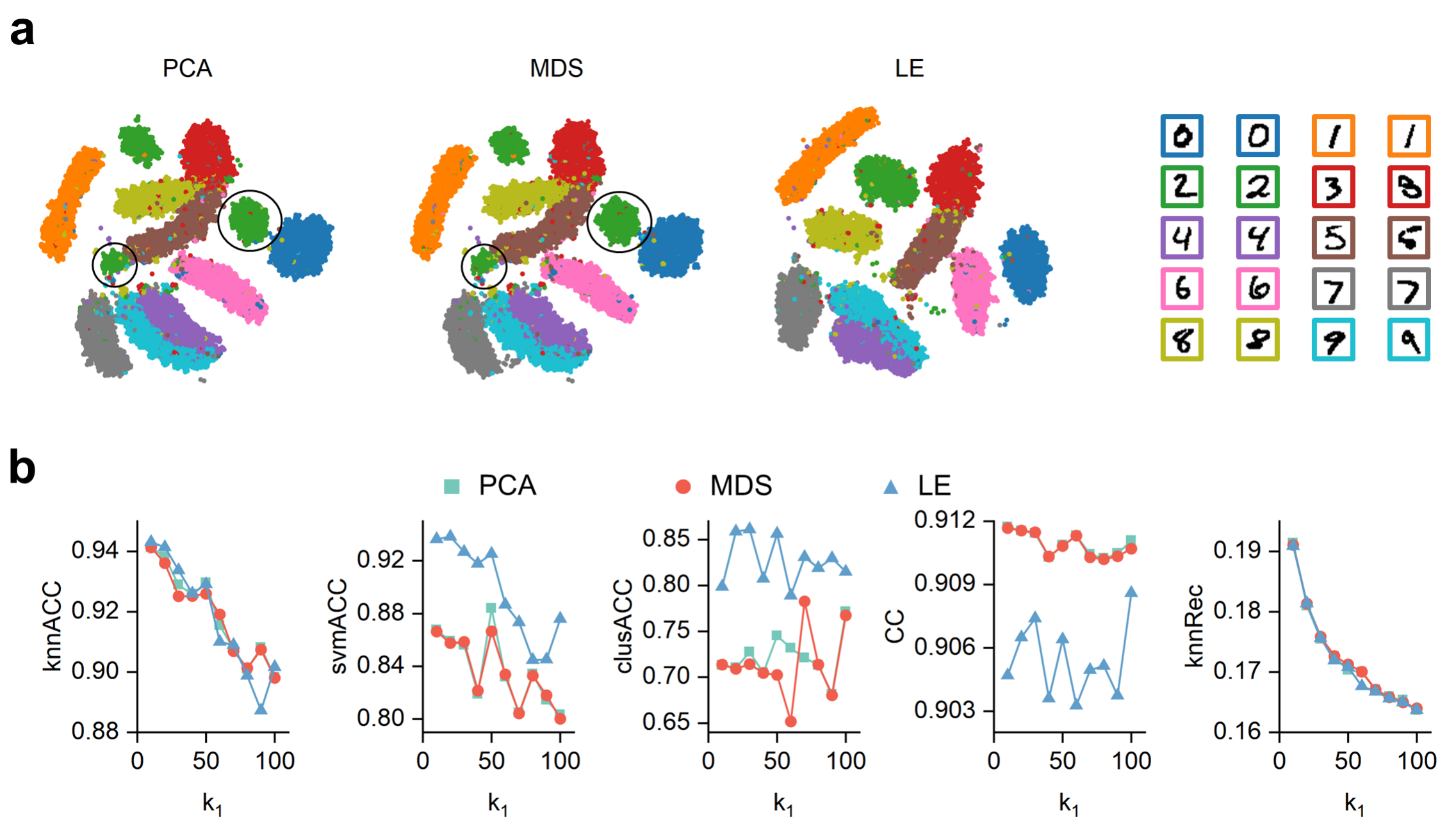

**Supplementary Fig. 22. Performance comparison among SUDE using three initialization methods on MNIST.** (**a**), Embeddings ($\boldsymbol{k_1} = \mathbf{50}$), and (**b**), five evaluation metrics of SUDE using PCA, MDS, and LE to initialize the embedding under different $\boldsymbol{k_1}$.

* In general, the three initialization methods can generate similar local structure reported by knnACC and knnRec scores. PCA and MDS have distinct advantages in preserving isometric relationships and global structure as shown in CC scores. While, they split digits "2" into three subclusters (black circles) and cannot separate digits "8" (olive) from the clump of digits "3" (red) and "5" (brown). In this case, LE outperforms them in svmACC and clusACC scores by a large margin, which indicates that LE can better preserve the cluster integrity and discrimination than MDS and PCA. Our implementation provides these three initialization methods for users to customize their option according to specific application requirements.

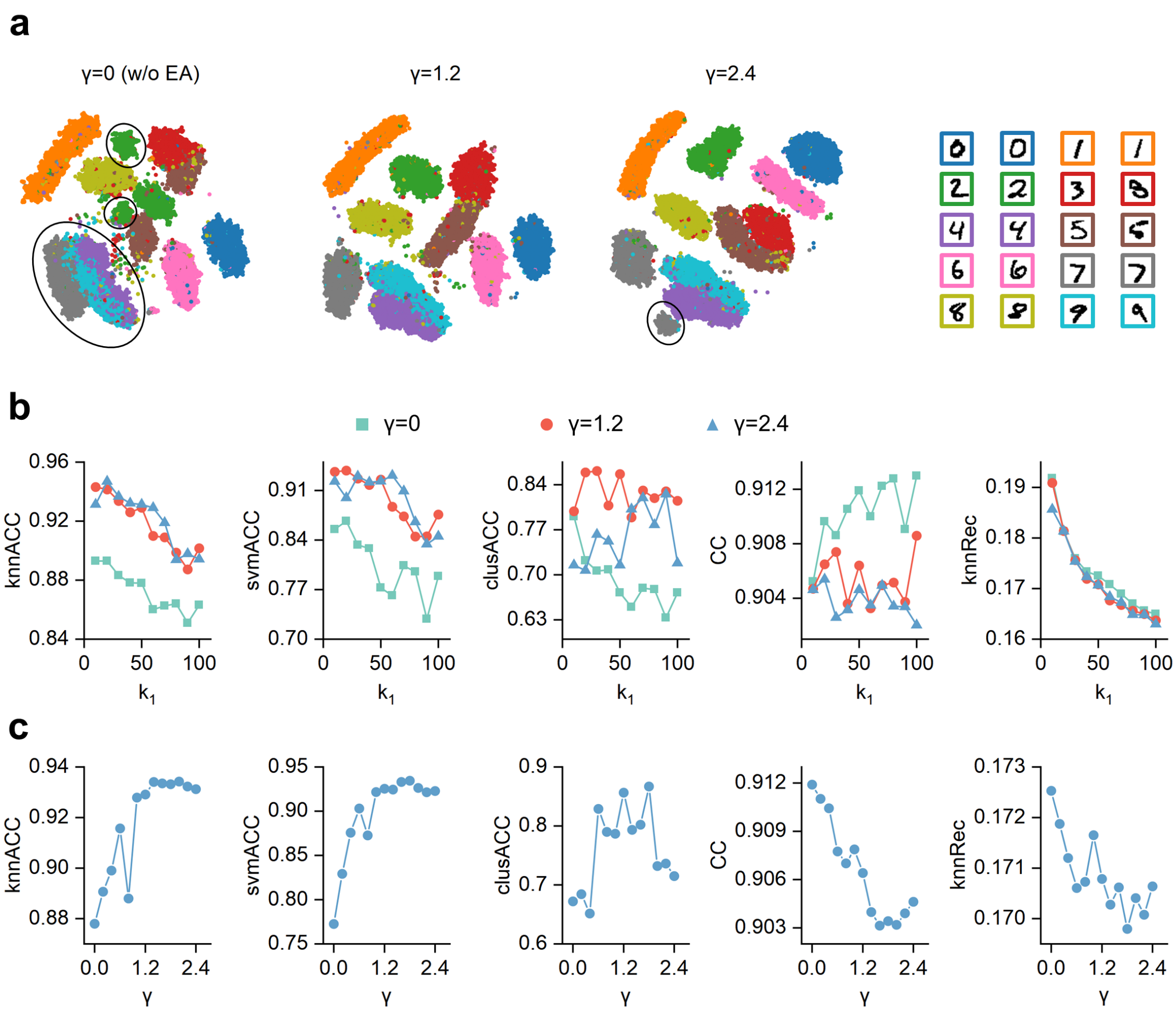

**Supplementary Fig. 23. Performance comparison between SUDE jointly with and without early aggregation (w/o EA) on MNIST.** (**a**), Embeddings ($\boldsymbol{k_1} = \mathbf{50}$), and (**b**), five evaluation metrics of SUDE by setting $\gamma$ to 0, 1.2, and 2.4 under different $\boldsymbol{k_1}$. (**c**), Five evaluation metrics of SUDE under different $\gamma$.

* The results demonstrate that w/o EA can better preserve isometric relationships (evidenced by higher CC scores), but the undersampling problem significantly degrades inter-cluster separation (Supplementary Note 9). In the embedding of w/o EA, digits "7" (gray), "4" (purple), and "9" (cyan) are merged as one cluster, and many sparse outliers are in the inter-cluster gaps (black circles), which leads to lower knnACC, svmACC, and clusACC scores. EA is able to enhance intra-cluster compactness and enlarge inter-cluster gaps. However, an overlarge aggregation coefficient may also destroy cluster integrity due to heterogeneous densities within clusters, as evidenced by the lower clusACC scores at $\gamma = 2.4$ compared to $\gamma = 1.2$. As the aggregation coefficient $\gamma$ increases, classification accuracy grows significantly until reaching a plateau at $\gamma = 1.2$, while clustering accuracy attains its peak within the range of 0.6 to 1.8. The preservation of global structure declines with higher $\gamma$ values. Meanwhile, the knnRec score keeps relative stable with fluctuations below 0.004, despite exhibiting an overall downward trend. Therefore, we set $\gamma = 1.2$ by default by considering the trade-off between the preservation of the cluster structure and the isometry as discussed in Supplementary Note 10.

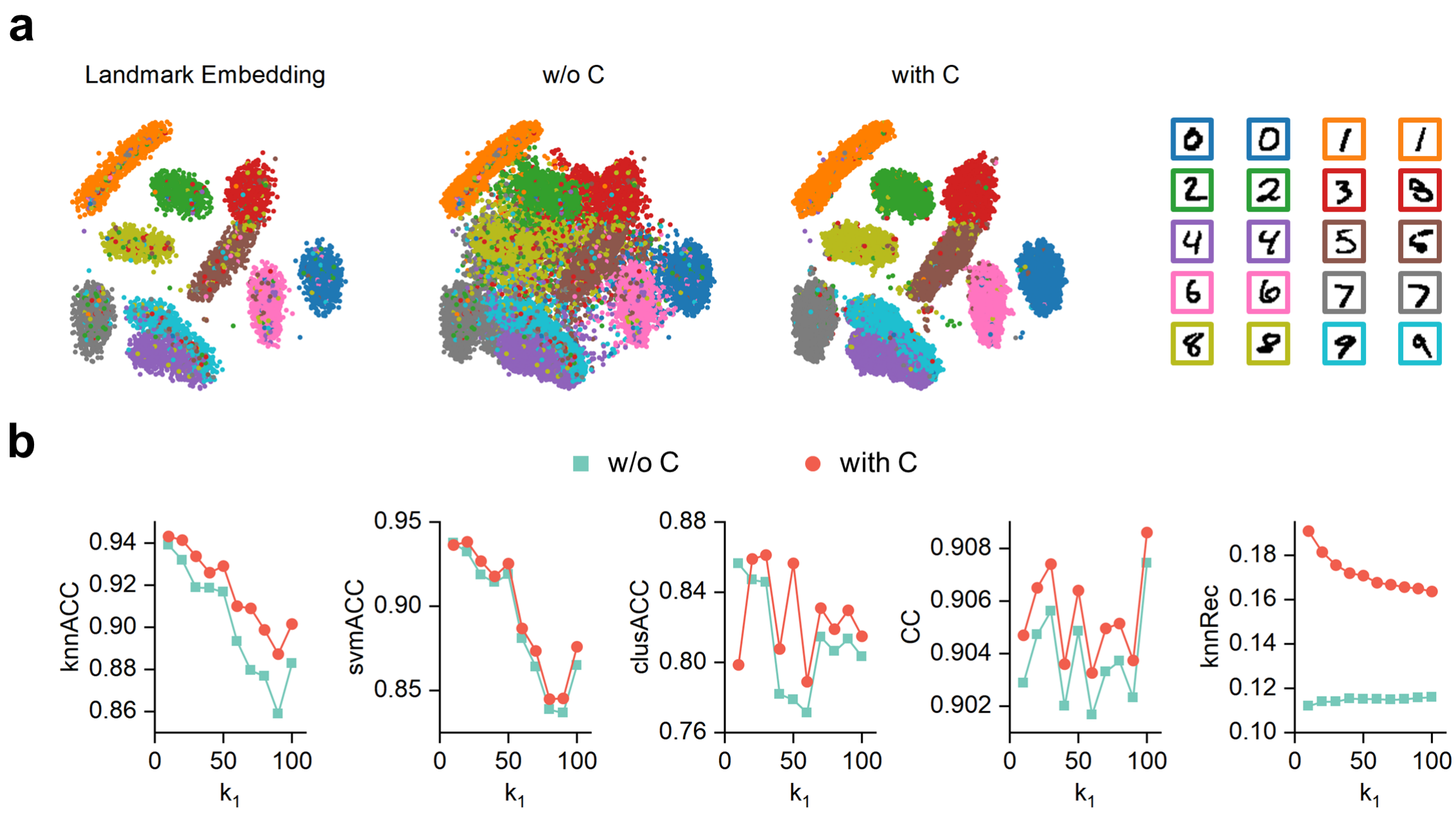

**Supplementary Fig. 24. Performance comparison between SUDE jointly with and without a constraint (w/o C) in LLE on MNIST.** (**a**), The landmark embedding and overall embeddings of SUDE ($k_1 = 50$) using LLE and CLLE for non-landmark. (**b**), Five evaluation metrics under different $k_1$.

* As we analyzed in Supplementary Note 11, w/o C would embed non-landmarks into the inter-cluster gaps and form dirty clusters when the sampling rate is low. The nearest distance constraint improves the performance of SUDE in classification/clustering accuracy and global structure preservation comprehensively. This constraint significantly enhances local structure preservation by enforcing closer proximity between each non-landmark and its nearest landmark.

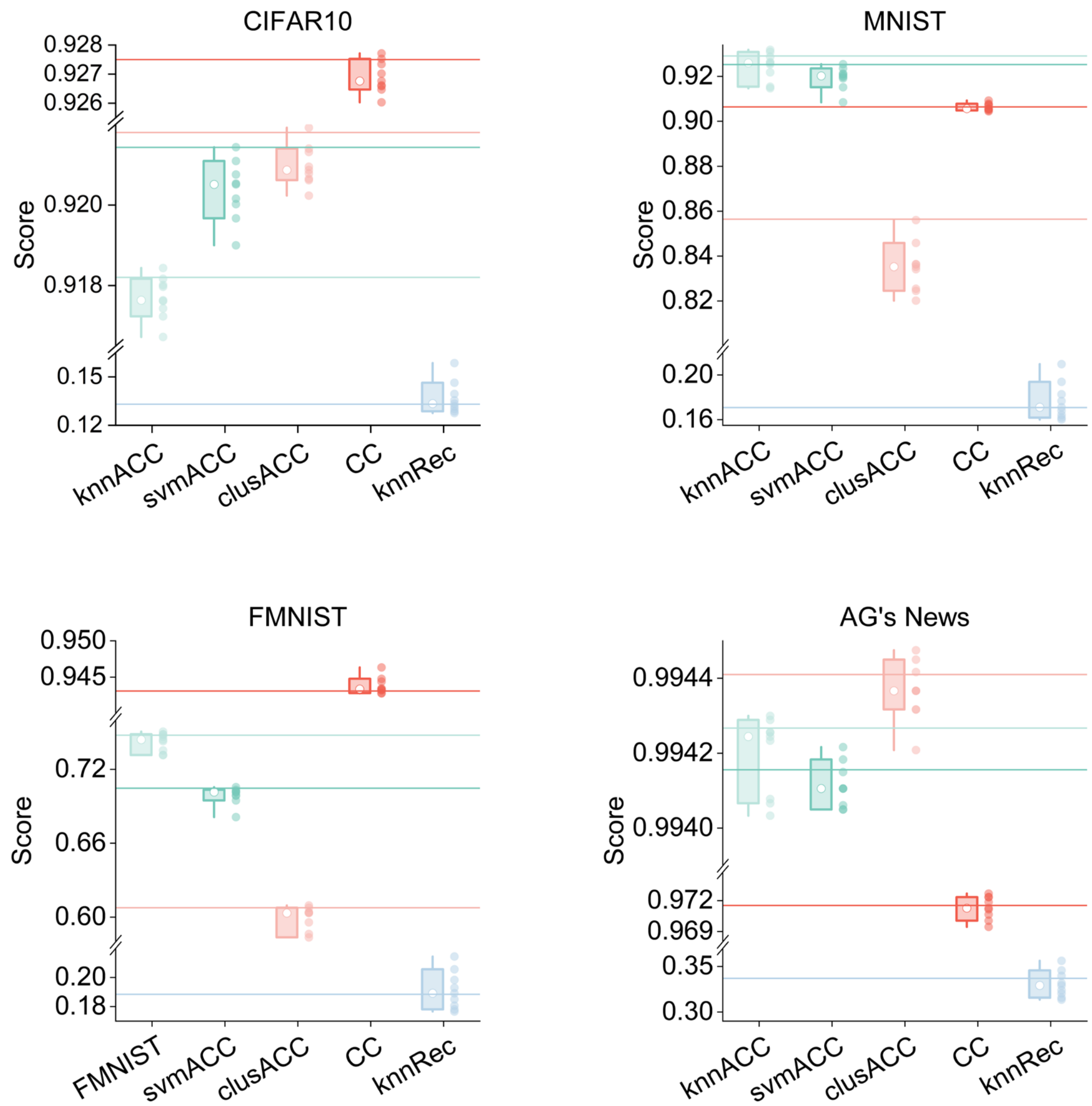

**Supplementary Fig. 25. Five evaluation metrics of SUDE using the adaptive $k_2$ on four real-world datasets.** The boxes and dots denote the knnACC, svmACC, clusACC, CC, and knnRec scores obtained by varied $k_2$ from 10 to 50 with an interval of 5, where the white circles refer to the median values. The lines represent the scores achieved by the adaptive $k_2$ using equation (18) in Methods.

* Most evaluation metrics obtained by the adaptive strategy surpass the median values in the specified parameter space. It demonstrates the effectiveness of the parameter adaption method.

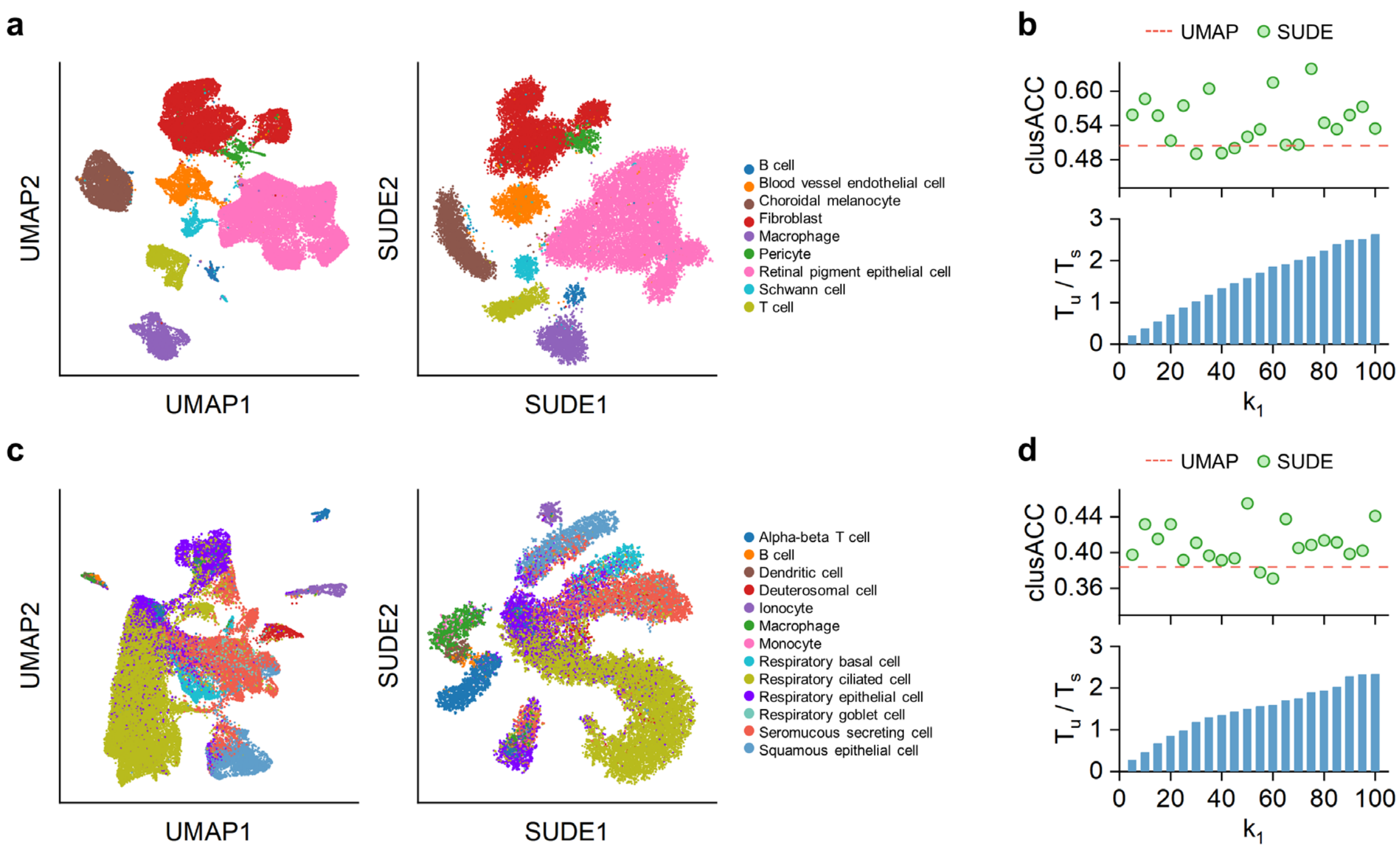

**Supplementary Fig. 26. Performance comparison on two scRNA-seq datasets.** 2-D embeddings of UMAP and SUDE ($\boldsymbol{k_1 = 50}$) on the (**a**), HRPEAC (human retinal pigment epithelium and choroid) dataset, and (**c**), NE (nasopharyngeal epithelium) dataset. The clusACC and $\boldsymbol{T_u/T_s}$ scores of UMAP and SUDE under different $\boldsymbol{k_1}$ on (**b**), HRPEAC and (**d**), NE, where $\boldsymbol{T_u/T_s}$ indicates the ratio of UMAP's runtime to SUDE's runtime.

* Both UMAP and SUDE can clearly distinguish the nine cell types on the HRPEAC dataset. However, SUDE separates blood vessel endothelial cells and Schwann cells from retinal pigment epithelial cells. These two methods can identify three substructures within fibroblasts, but SUDE better preserves the integrity of fibroblasts rather than separate them. For the NE dataset, UMAP can discover more substructures within seromucous secreting cell, and separates deuterosomal cells clearly. However, it mixes B cells, dendritic cells, and macrophages together. SUDE can distinguish these three types of cells. SUDE obtains higher clusACC scores than UMAP under most sampling rates (Supplementary Fig. 26a), since it can not only distinguish cell types, but also preserves cluster integrity. In terms of scalability, SUDE runs faster than UMAP when $k_1 > 30$ ($T_u/T_s > 1$, and the sampling rate is approximately 20%), and is almost 3-fold more efficient than UMAP when $k_1 = 100$.

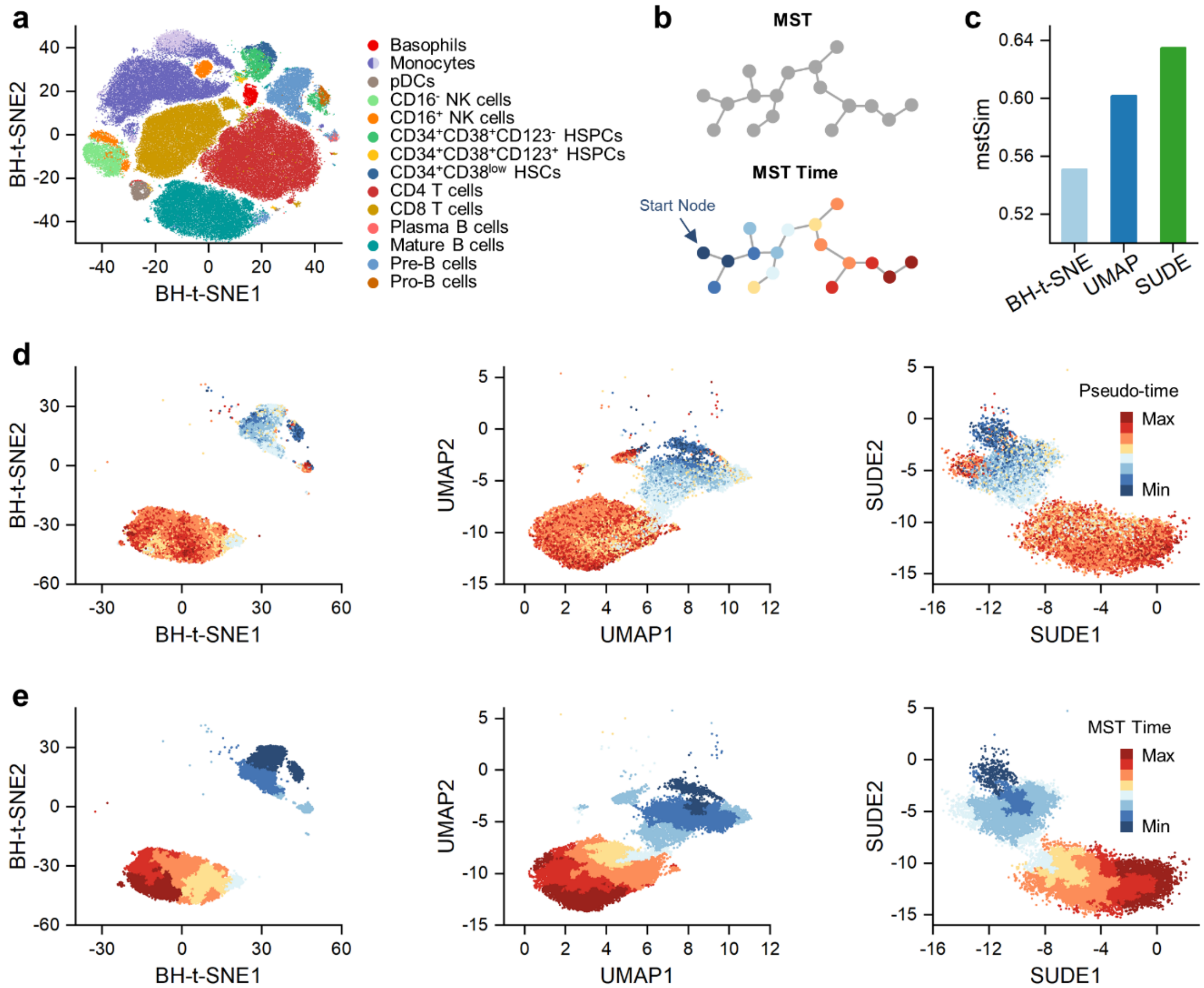

**Supplementary Fig. 27. Quantitative comparison of the three embeddings with respect to pseudo-time analysis on Levine.** (**a**), 2-D embedding of BH-t-SNE. (**b**), Illustration of the mstSim metric that measures the similarity between the inferred pseudo-time and minimum spanning tree (MST) time. We first construct the MST based on the 2-D embedding. Then, we select the cell whose pseudo-time is equal to 0 as the start node, and calculate the shortest path distances as MST time. mstSim is defined as the Spearman's rank correlation coefficient between pseudo-time and MST time. (**c**), mstSim scores obtained from the embeddings of BH-t-SNE, UMAP, and SUDE. (**d**), The development landscape of B cells on BH-t-SNE, UMAP, and SUDE plots, which are colored by the pseudo-time generated after selecting an early Pro-B cell for initiation. (**e**), BH-t-SNE, UMAP, and SUDE plots colored by the MST time.

* The results of mstSim indicate that the cell distribution in the SUDE plot is more in line with the development process of B cells. BH-t-SNE embedding pushes the mature B cells away from other B cells, which destroys the continuous transition states. Compared to UMAP, SUDE makes the distribution of different cells more compact and can also allow plasma B cells to obtain a later MST time, which is more consistent with the real differentiation process.

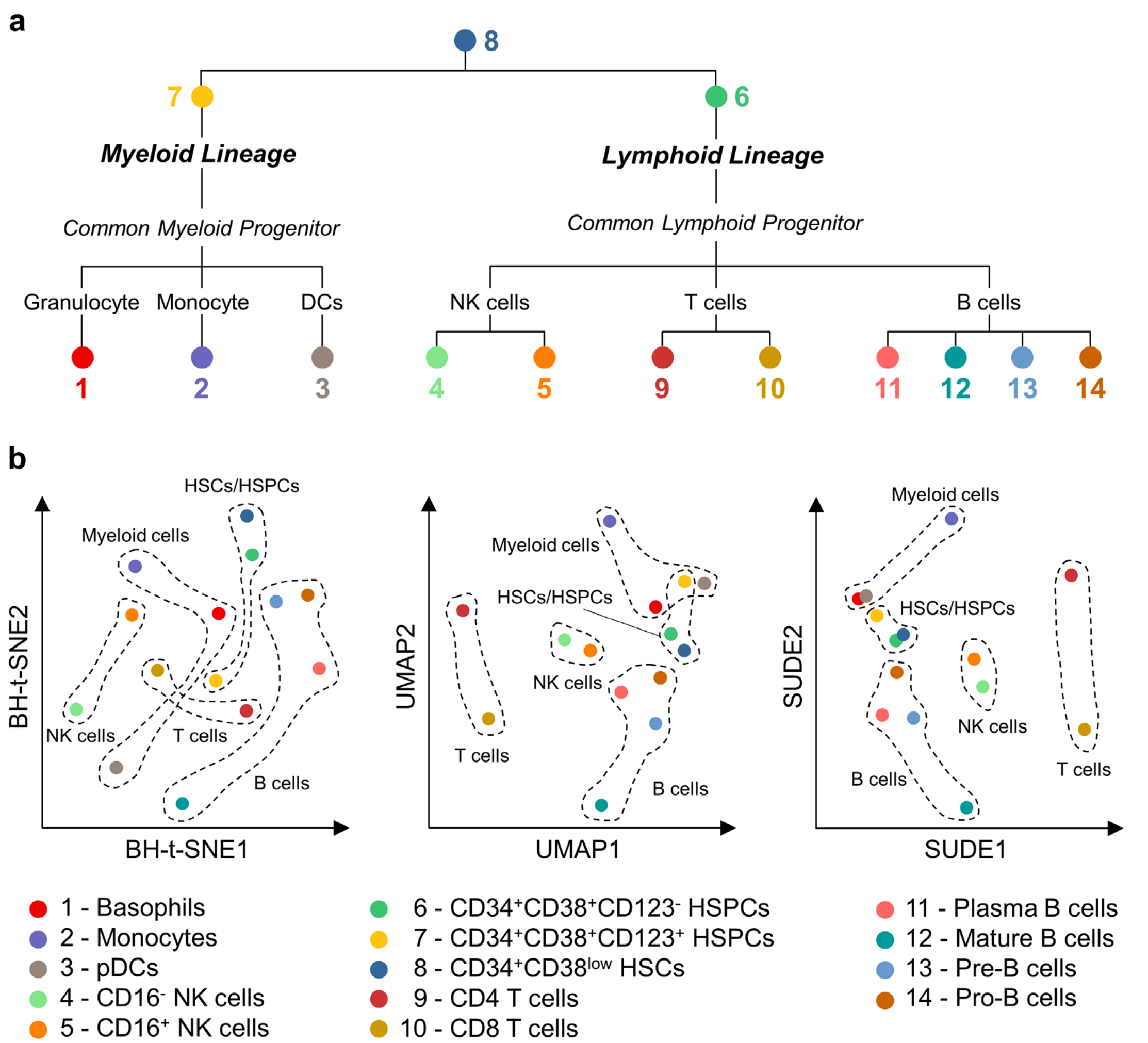

**Supplementary Fig. 28. Comparison of the three embeddings in preserving cellular differentiation hierarchy on Levine.** (**a**), A lineage tree shows the differentiation hierarchy of 14 healthy human bone marrow cell types. (**b**), The scatter plots show the distributions of cluster centroids in the BH-t-SNE, UMAP, and SUDE embeddings.

* The hematopoietic hierarchy originates from $CD34^{+}CD38^{low}$ HSCs, which differentiate into $CD34^{+}CD38^{+}$ HSPCs. Within HSPCs, CD123 expression delineates lineage bias: $CD123^{-}$ progenitors exhibit lymphoid priming, while $CD123^{+}$ progenitors drive myeloid specification. Common myeloid progenitors further diverge into granulocyte-monocyte precursors giving rise to basophils and monocytes, alongside dendritic cell precursors generating pDCs. Simultaneously, common lymphoid progenitors lead to NK cells, T cells, and B cells. t-SNE fails to preserve cellular hierarchy, with cells from the same differentiation level becoming dispersed in the embedding space. Both UMAP and SUDE demonstrate capacity to distinguish myeloid and lymphoid lineages, achieving clear separation among NK cells, T cells, and B cell subsets. However, UMAP mixes up the three myeloid cell populations and $CD34^{+}CD38^{+}CD123^{+}$ HSPCs. SUDE effectively preserves the distinction between myeloid cells and HSCs/HSPCs.

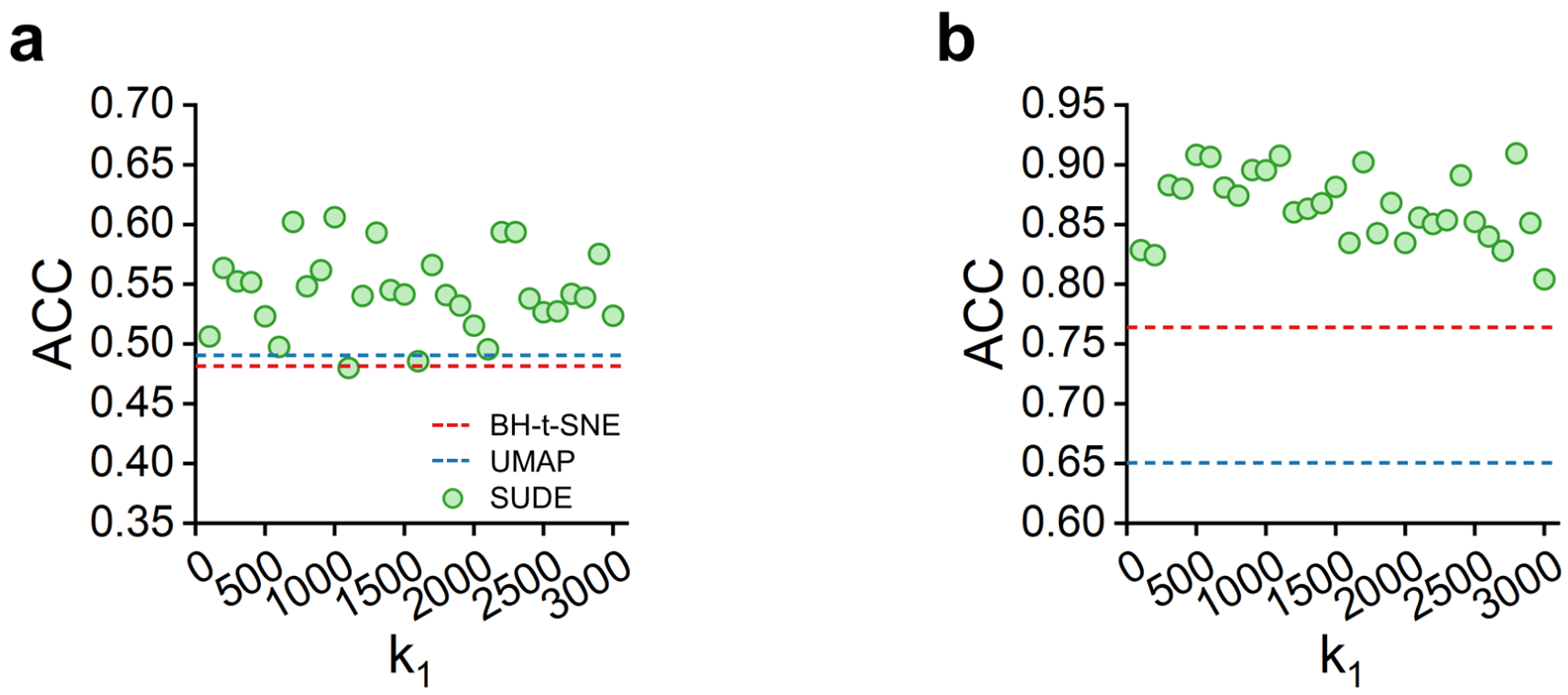

**Supplementary Fig. 29. The clustering performance on the BH-t-SNE, UMAP, and SUDE embeddings for the Levine dataset.** ACC scores of the (**a**), K-means and (**b**), CDC ($\boldsymbol{k = 20, ratio = 0.95}$) algorithms on the embeddings by varying $\boldsymbol{k_1}$ from 100 to 3,000 with an interval of 100, respectively.

* Most ACC scores upon the SUDE embedding are higher. Especially in the CDC results, SUDE leads BH-t-SNE and UMAP by a large margin.

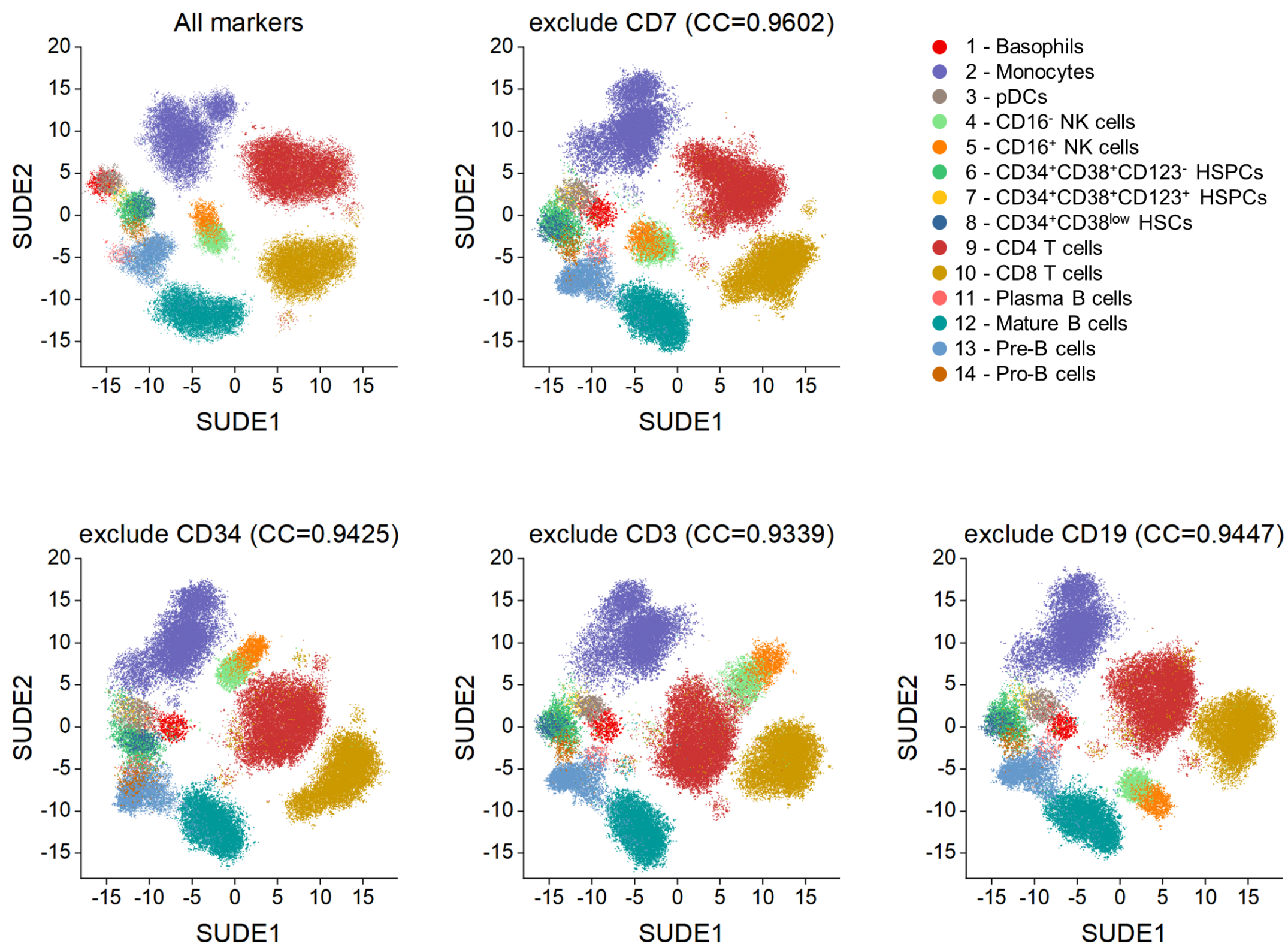

**Supplementary Fig. 30. 2-D embeddings of SUDE on Levine by removing different canonical markers.** The SUDE maps in the last four panels are generated by 31 surface markers, where the marker listed above the panel was excluded in that run.

* SUDE is robust to the marker selection. Despite removing the canonical markers of NK cells, HSPCs, T cells, and B cells (CD7, CD34, CD3, and CD19, respectively), cell types can be identified and separated well in each panel. The high CC scores show that the leave-one-out SUDE embeddings are almost identical to the embedding with all surface markers.

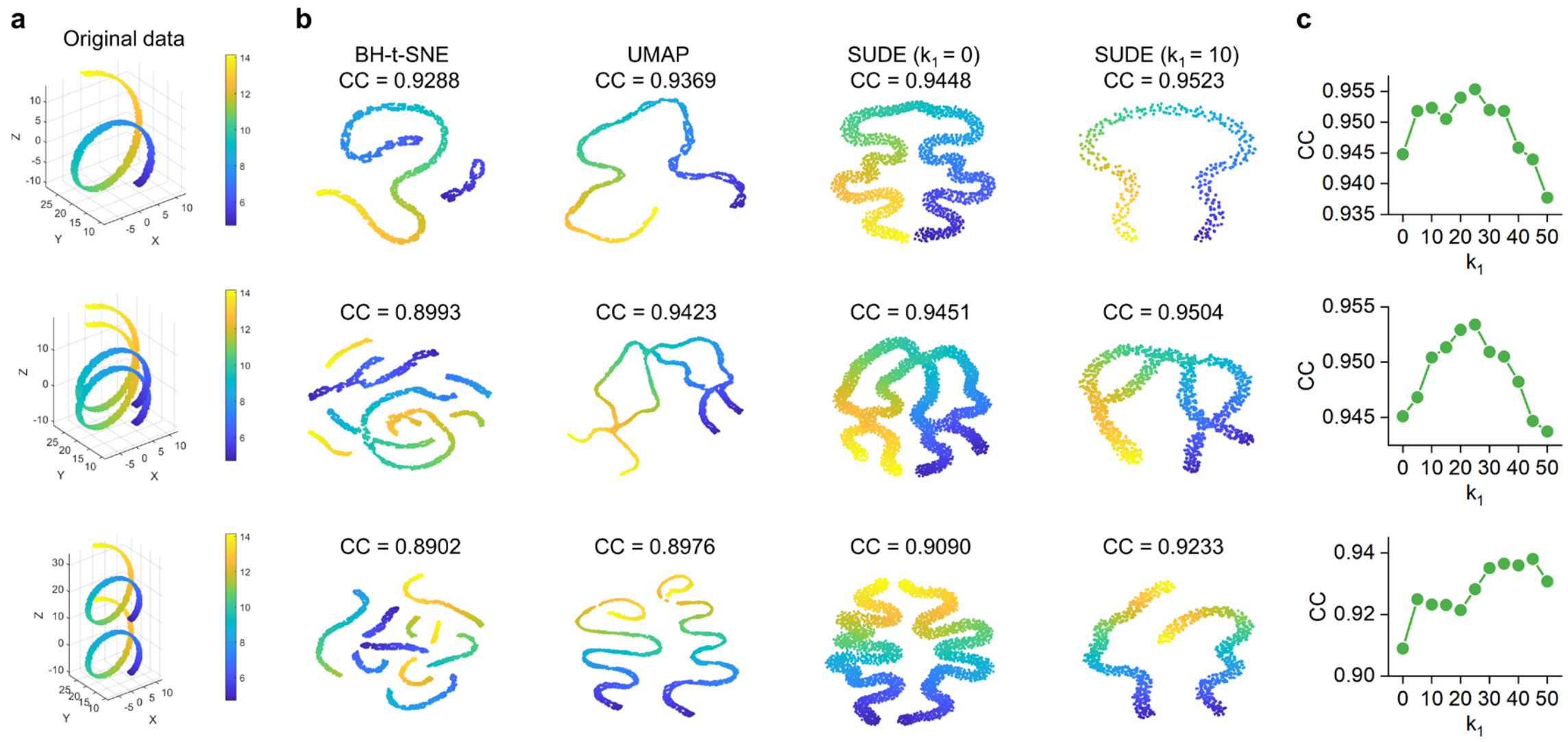

**Supplementary Fig. 31. Performances of BH-t-SNE, UMAP, and SUDE on three toy datasets. (a)**, The original data in 3-D space. **(b)**, 2-D embeddings of BH-t-SNE, UMAP, and SUDE. (**c**), CC scores of SUDE under different $\boldsymbol{k_1}$.

* We compared the performances of BH-t-SNE, UMAP, and SUDE on three toy datasets to further demonstrate SUDE's ability to preserve continuous variations. As presented in Supplementary Fig. 31a, the first dataset contains a smooth continuous ribbon comprising 2,000 points, which is essentially a Swiss Roll stretched spirally along the Y axis. The second dataset contains two same ribbons with three intersection areas between them, while these two ribbons are separate without any overlap in the third dataset. It can be observed that BH-t-SNE tends to break up the continuous structures in Supplementary Fig. 31b. Both UMAP and SUDE is able to preserve data coherence, and SUDE exhibits slightly higher CC scores than UMAP with $k_1 = 0$ (without applying sampling procedure). SUDE's CC score presents an upward-then-downward trend as $k_1$ increases (sampling rate decreases) in Supplementary Fig. 31c. This arises because landmark sampling expands the neighborhood range for embedding learning and enables to capture long distance relationships. However, landmark sampling also causes loss of structural details simultaneously. For instance, SUDE merges the yellow endpoints of the two ribbons in the second dataset under $k_1 = 10$ (sampling rate is 18%). In the third dataset, UMAP splits the yellow endpoint regions into disconnected parts, whereas SUDE maintains both the separation and continuous of the two ribbons. In summary, SUDE can handle continuously varying data structures. Moderate landmark sampling enhances SUDE's global structure preservation, but excessive sampling still destroys the structural coherence.

**Supplementary Table 1. Differences between t-SNE, UMAP and SUDE in the embedding learning process.**

| | t-SNE | UMAP | SUDE |
|---|---|---|---|
| Dissimilarity (take Euclidean distance for example) | $d_{ij}=\lVert\boldsymbol{x}_i-\boldsymbol{x}_j\rVert$ | $d_{ij}=\lVert\boldsymbol{x}_i-\boldsymbol{x}_j\rVert$ | $d_{j\mid i}=\left(1-\frac{SNN_{ij}}{\max_i SNN_{ij}}\right)^{\gamma}\lVert\boldsymbol{x}_i-\boldsymbol{x}_j\rVert$ |
| High-dimensional conditional probabilities | $p_{j\mid i}=\exp\left(-\frac{{d_{ij}}^2}{2{\sigma_i}^2}\right)$ | $p_{j\mid i}=\exp\left(-\frac{d_{ij}-\rho_i}{\sigma_i}\right)$ | $p_{j\mid i}=\exp\left(-\frac{{d_{j\mid i}}^2}{2{\sigma_i}^2}\right)$ |
| Gaussian bandwidth | $Perp=2^{-\sum_j p_{j\mid i}\log p_{j\mid i}}$ | $\sum_{\boldsymbol{x}_j\in KNN(\boldsymbol{x}_i)}\exp\left(-\frac{d_{ij}-\rho_i}{\sigma_i}\right)=\log_2 k$ | $\sigma_i=\frac{1}{k}\sum_{\boldsymbol{x}_j\in KNN(\boldsymbol{x}_i)}d_{j\mid i}$ |
| Normalization and symmetrization | $p_{ij}=\frac{p_{i\mid j}+p_{j\mid i}}{2n\sum_{k\neq l}p_{k\mid l}}$ | $p_{ij}=p_{j\mid i}+p_{i\mid j}-p_{j\mid i}p_{i\mid j}$ | $p_{ij}=\frac{p_{i\mid j}+p_{j\mid i}}{2\sum_{k\neq l}p_{k\mid l}}$ |
| Low-dimensional conditional probabilities | $q_{ij}=\frac{(1+\lVert\boldsymbol{y}_i-\boldsymbol{y}_j\rVert^2)^{-1}}{\sum_{k\neq l}(1+\lVert\boldsymbol{y}_k-\boldsymbol{y}_l\rVert^2)^{-1}}$ | $q_{ij}=(1+a\lVert\boldsymbol{y}_i-\boldsymbol{y}_j\rVert^{2b})^{-1}$ | $q_{ij}=\frac{\left(1+\log(1+\lVert\boldsymbol{y}_i-\boldsymbol{y}_j\rVert^2)\right)^{-1}}{\sum_{k\neq l}\left(1+\log(1+\lVert\boldsymbol{y}_k-\boldsymbol{y}_l\rVert^2)\right)^{-1}}$ |
| Cost function | $\mathcal{L}=\sum_{i\neq j}p_{ij}\log\frac{p_{ij}}{q_{ij}}$ | $\mathcal{L}=\sum_{i\neq j}p_{ij}\log\frac{p_{ij}}{q_{ij}}+(1-p_{ij})\log(\frac{1-p_{ij}}{1-q_{ij}})$ | $\mathcal{L}=\sum_{i\neq j}p_{ij}\log\frac{p_{ij}}{q_{ij}}$ |
| Initialization | *Random initialization* | *Laplacian eigenmaps* | *Laplacian eigenmaps* |
| Optimizer | $\boldsymbol{y}_i^{(t)}=\boldsymbol{y}_i^{(t-1)}-\eta\frac{\partial\mathcal{L}}{\partial\boldsymbol{y}_i^{(t)}}+\alpha(t)(\boldsymbol{y}_i^{(t-1)}-\boldsymbol{y}_i^{(t-2)})$ | *Stochastic gradient descent and negative sampling* | $\boldsymbol{y}_i^{(t)}=\boldsymbol{y}_i^{(t-1)}-\eta(t)\left(\frac{\partial\mathcal{L}}{\partial\boldsymbol{y}_i^{(t)}}+\alpha(t)\frac{\partial\mathcal{L}}{\partial\boldsymbol{y}_i^{(t-1)}}\right)$ |

**Supplementary Table 2. Information of the 13 real-world datasets.**

| Dataset | Description | # of Samples | Dimensions | Classes |
|---|---|---|---|---|
| Wine | Chemical Constituents of Wines | 178 | 13 | 3 |
| Dermatology | Eryhemato-Squamous Disease | 358 | 34 | 6 |
| Breast-Cancer | Wisconsin Breast Cancer | 683 | 10 | 2 |
| Mfeat | Handwritten Digits | 2,000 | 649 | 10 |
| Rice | Grain Images of Rice | 3,810 | 7 | 2 |
| Spambase | Email Content and Headers | 4,601 | 57 | 2 |
| Dry-Bean | Grain Images of Dry Beans | 13,611 | 16 | 7 |
| Shuttle | Spacecraft Sensors | 14,500 | 9 | 7 |
| CIFAR10 | Natural Images | 60,000 | 1,024 | 10 |
| MNIST | Handwritten Digits | 70,000 | 784 | 10 |
| FMNIST | Fashion Products | 70,000 | 784 | 10 |
| AG's News | Text Documents | 120,000 | 100 | 4 |
| Yahoo | Text Documents | 1,400,000 | 100 | 10 |

**Wine**, **Dermatology**, **Breast-Cancer**, **Mfeat**, **Rice**, **Spambase**, **Dry-Bean**, and **Shuttle** are eight UC Irvine datasets, in which features denote some attributes of real-world instances.

**CIFAR10** is a dataset including 60,000 32×32 color images in 10 classes, with 6,000 images per class. Each image has been applied to a convolutional neural network to generate 1024-D features.

**MNIST** contains 70,000 grayscale images of handwritten digits from 0 to 9 with the size of 28×28. We convert the 2-D images to 1-D vectors and input into the process of dimension reduction.

**FMNIST** consists of 70,000 28×28 pixel grayscale images and includes ten categories of fashion products (T-shirt, trouser, pullover, dress, coat, sandal, shirt, sneaker, bag, and ankle boot). Like with MNIST, FMNIST is also treated as 70,000 different 784-D vectors.

**AG's News** collects news articles in four categories (world, sports, business, and science technique). Each article is represented by a 100-D vector generated by FastText [30].

**Yahoo** has 1,400,000 news articles, which have been applied to FastText to extract 100-D features.

**Supplementary Table 3. Hyperparameter settings of the baselines.**

| Parameter | t-SNE / BH-t-SNE | UMAP | TriMap |
|---|---|---|---|
| *Distance Metric* | Euclidean | Euclidean | Euclidean |
| *Number of Dimensions* | 2 | 2 | 2 |
| *Number of Neighbors* | 3**Perplexity* | 15 | 10 |
| *Number of Iterations* | 1000 | 200 | 400 |
| *Learning Rate* | 500 | 1 | 1000 |
| *KNN Method* | KD-Tree/Exhaustive | NN-Descent | HNSW |
| *Initialization Method* | Random | Laplacian Eigenmaps | PCA |
| *Optimization Method* | Momentum | SGD | Delta-Bar-Delta |
| *Loss Function* | KLD | CE | Triplet loss |
| *Others* | *Perplexity*: 30<br>*Exaggeration*: 4<br>*Theta*: 0.5 | *Min-dist*: 0.3<br>*Negative Sampling Rate*: 5<br>*a*: 1.577<br>*b*: 0.895 | *Number of Outliers*: 5<br>*Number of Random Triplets per Point*: 5<br>*Weight Adjustment*: 500 |

| Parameter | TopoAE | P-UMAP |
|---|---|---|
| *Distance Metric* | Euclidean | Euclidean |
| *Number of Dimensions* | 2 | 2 |
| *Number of Neighbors* | / | 15 |
| *Number of Iterations* | 10 | 10 |
| *Batch Size* | 64 | 1000 |
| *Learning Rate* | 1e-3 | 1e-3 |
| *KNN Method* | / | NN-Dscent |
| *Initialization Method* | Random | Random & Laplacian Eigenmaps |
| *Optimization Method* | Adam | Adam |
| *Activation Function* | ReLU | ReLU |
| *Loss Function* | MSE + topological loss | MSE + UMAP loss |
| *Neural Network Architecture* | Encoder: 3-layer hidden neurons (1000-500-250)<br>Bottleneck: 2-neuron hidden layer<br>Decoder: 3-layer hidden neurons (250-500-1000) | Encoder: 3-layer hidden neurons (100-100-100)<br>Bottleneck: 2-neuron hidden layer<br>Decoder: 3-layer hidden neurons (100-100-100) |

***a*, *b*** - The parameters of the low-dimensional probability $q_{ij}=(1+a\|\boldsymbol{y}_i-\boldsymbol{y}_j\|^{2b})^{-1}$ in UMAP.

***Activation Function*** - The function applied to the output of a neuron for introducing non-linearity.

***Batch Size*** - The number of samples used in each of these smaller batches during training.

***Distance Metric*** - The metric of distance when calculating distance between observations.

***Exaggeration*** - The tightness of the natural clusters in high-dimensional data at the start of the optimization.

***Initialization Method*** - The method for generating the initial low-dimensional embedding in t-SNE, UMAP, TriMap, and P-UMAP, and for specifying the initial neural network weights in TopoAE and P-UMAP.

***KNN Method*** - The method for the search of nearest neighbors. t-SNE finds the neighbors with KD-Tree when the data dimension is less than or equal to 10. Otherwise, the exhaustive algorithm is used. UMAP and P-UMAP adopt NN-Descent for the approximate KNN search with 1% accuracy loss. TriMap utilizes a graph-based approximate KNN method HNSW.

***Learning Rate*** - The learning rate of the optimization process.

***Loss Function*** - The objective of the manifold learning method.

***Min-dist*** - The cutoff distance that controls how tightly UMAP packs points together.

***Neural Network Architecture*** - The components of the autoencoder used in TopoAE and P-UMAP.

***Negative Sampling rate*** - The number of negative samples per positive sample in the optimization process.

***Number of Dimensions*** - The number of dimensions of the low-dimensional embedding.

***Number of Iterations*** - The maximum number of iterations for optimization process.

***Number of Neighbors*** - The number of nearest neighbors.

***Number of Outliers*** - The number of outliers for forming the nearest neighbor triplets.

***Number of Random Triplets per Point*** - The number of random triplets added for each point.

***Optimization Method*** - The method for optimizing the low-dimensional embedding. t-SNE uses the gradient descent with momentum. UMAP leverages the stochastic gradient descent (SGD) method. TriMap employs Delta-Bar-Delta method [31]. TopoAE and P-UMAP utilize Adam optimizer [32].

***Perplexity*** - The effective number of local neighbors of each point for determining the Gaussian bandwidth.

***Theta*** - The argument to trade-off of speed and accuracy for Barnes-Hut algorithm.

***Weight Adjustment*** - The value of gamma for the log-transformation

**Supplementary Table 4. Assessment of BH-t-SNE, UMAP, TriMap, TopoAE, P-UMAP, and SUDE on 12 real-world datasets.**

| | | knnACC | svmACC | clusACC | CC | knnRec | Runtime (s) |
|---|---|---|---|---|---|---|---|
| Wine | BH-t-SNE | 0.887 | 0.880 | 0.899 | 0.934 | **0.626** | 0.44 |
| | UMAP | 0.902 | 0.880 | 0.899 | 0.933 | 0.597 | 0.99 |
| | TriMap | 0.692 | 0.714 | 0.725 | 0.846 | 0.605 | 0.68 |
| | TopoAE | 0.842 | 0.842 | 0.832 | **0.938** | 0.374 | 5.98 |
| | P-UMAP | 0.880 | 0.895 | 0.843 | 0.926 | 0.582 | 1.02 |
| | SUDE | **0.932** | **0.932** | **0.927** | 0.921 | 0.501 | **0.02** ($k_1$=20) |
| Dermatology | BH-t-SNE | 0.899 | 0.929 | 0.631 | 0.928 | **0.593** | 0.66 |
| | UMAP | 0.914 | 0.914 | 0.821 | 0.895 | 0.527 | 1.14 |
| | TriMap | 0.687 | 0.661 | 0.327 | 0.849 | 0.583 | 0.80 |
| | TopoAE | 0.888 | 0.847 | 0.779 | **0.950** | 0.362 | 8.07 |
| | P-UMAP | **0.966** | **0.978** | 0.777 | 0.911 | 0.552 | 3.12 |
| | SUDE | 0.925 | 0.933 | **0.844** | 0.925 | 0.461 | **0.03** ($k_1$=20) |
| Breast-Cancer | BH-t-SNE | 0.943 | 0.940 | 0.816 | 0.845 | **0.669** | 1.98 |
| | UMAP | 0.941 | 0.941 | 0.710 | 0.864 | 0.582 | 1.47 |
| | TriMap | 0.641 | 0.600 | 0.647 | 0.357 | 0.623 | 1.10 |
| | TopoAE | 0.957 | 0.956 | 0.953 | **0.979** | 0.506 | 10.87 |
| | P-UMAP | 0.941 | 0.936 | 0.741 | 0.844 | 0.615 | 2.98 |
| | SUDE | **0.957** | **0.957** | **0.959** | 0.884 | 0.594 | **0.07** ($k_1$=20) |
| Mfeat | BH-t-SNE | 0.913 | 0.802 | 0.752 | 0.932 | **0.646** | 2.87 |
| | UMAP | 0.904 | 0.893 | 0.801 | 0.916 | 0.547 | 1.81 |
| | TriMap | 0.900 | 0.897 | 0.709 | 0.836 | 0.505 | 2.20 |
| | TopoAE | 0.679 | 0.699 | 0.609 | 0.932 | 0.246 | 23.16 |
| | P-UMAP | 0.881 | 0.890 | 0.800 | **0.973** | 0.247 | 25.30 |
| | SUDE | **0.953** | **0.949** | **0.945** | 0.920 | 0.270 | **0.62** ($k_1$=20) |
| Rice | BH-t-SNE | 0.893 | 0.899 | 0.901 | 0.965 | 0.625 | 5.79 |
| | UMAP | 0.896 | 0.906 | 0.903 | 0.979 | 0.461 | 2.47 |
| | TriMap | 0.853 | 0.878 | 0.863 | 0.921 | **0.657** | 3.44 |
| | TopoAE | 0.911 | **0.919** | **0.915** | **0.988** | 0.412 | 39.59 |
| | P-UMAP | 0.883 | 0.892 | 0.890 | 0.973 | 0.388 | 38.21 |
| | SUDE | **0.914** | 0.906 | 0.907 | 0.964 | 0.398 | **0.26** ($k_1$=20) |
| Spambase | BH-t-SNE | 0.824 | 0.777 | 0.771 | 0.868 | **0.585** | 11.53 |
| | UMAP | 0.822 | 0.788 | 0.778 | 0.800 | 0.452 | 2.98 |
| | TriMap | 0.732 | 0.684 | 0.692 | 0.709 | 0.519 | 3.72 |
| | TopoAE | 0.809 | **0.813** | 0.541 | **0.934** | 0.261 | 47.20 |
| | P-UMAP | 0.823 | 0.807 | 0.776 | 0.817 | 0.446 | 55.33 |
| | SUDE | **0.827** | 0.810 | **0.809** | 0.845 | 0.388 | **1.35** ($k_1$=20) |
| Dry-Bean | BH-t-SNE | 0.887 | 0.877 | 0.611 | 0.918 | **0.534** | 23.86 |
| | UMAP | 0.871 | 0.876 | 0.678 | 0.962 | 0.321 | 15.80 |
| | TriMap | 0.634 | 0.608 | 0.527 | 0.844 | 0.514 | 10.11 |
| | TopoAE | 0.867 | 0.872 | 0.639 | **0.989** | 0.231 | 129.96 |
| | P-UMAP | 0.878 | 0.873 | 0.695 | 0.949 | 0.243 | 101.78 |
| | SUDE | **0.890** | **0.894** | **0.736** | 0.954 | 0.294 | **5.87** ($k_1$=20) |
| Shuttle | BH-t-SNE | **0.993** | 0.832 | 0.256 | 0.815 | 0.713 | 29.13 |
| | UMAP | 0.992 | 0.830 | 0.316 | 0.863 | 0.550 | 4.80 |
| | TriMap | 0.981 | 0.922 | 0.343 | 0.889 | 0.475 | 12.00 |
| | TopoAE | 0.991 | 0.922 | 0.482 | 0.987 | 0.190 | 139.22 |
| | P-UMAP | 0.967 | 0.792 | **0.788** | **0.999** | 0.446 | 111.51 |
| | SUDE | 0.986 | **0.978** | 0.346 | 0.887 | **0.750** | **2.67** ($k_1$=20) |
| CIFAR10 | BH-t-SNE | **0.963** | **0.960** | **0.955** | 0.934 | **0.292** | 494.60 |
| | UMAP | 0.959 | 0.958 | 0.938 | 0.930 | 0.132 | 107.00 |
| | TriMap | 0.959 | 0.959 | 0.801 | 0.894 | 0.105 | **95.00** |
| | TopoAE | 0.657 | 0.694 | 0.557 | **0.944** | 0.108 | 592.60 |
| | P-UMAP | 0.926 | 0.905 | 0.689 | 0.942 | 0.115 | 704.24 |
| | SUDE | 0.919 | 0.921 | 0.922 | 0.928 | 0.133 | 109.45 ($k_1$=50)/69.05 ($k_1$=100) |
| MNIST | BH-t-SNE | 0.970 | 0.829 | 0.725 | 0.902 | **0.396** | 524.00 |
| | UMAP | 0.968 | 0.962 | 0.815 | 0.893 | 0.169 | 105.00 |
| | TriMap | **0.971** | 0.952 | 0.510 | 0.673 | 0.167 | **93.00** |
| | TopoAE | 0.442 | 0.474 | 0.370 | **0.917** | 0.110 | 671.41 |
| | P-UMAP | 0.969 | **0.968** | 0.818 | 0.892 | 0.137 | 761.53 |
| | SUDE | 0.929 | 0.925 | **0.856** | 0.906 | 0.171 | 89.20 ($k_1$=50)/58.53 ($k_1$=100) |
| FMNIST | BH-t-SNE | **0.823** | 0.620 | 0.511 | 0.906 | **0.369** | 528.00 |
| | UMAP | 0.742 | 0.707 | 0.579 | 0.928 | 0.175 | 126.00 |
| | TriMap | 0.738 | **0.707** | 0.521 | 0.892 | 0.166 | **93.00** |
| | TopoAE | 0.610 | 0.623 | 0.487 | 0.862 | 0.119 | 682.53 |
| | P-UMAP | 0.651 | 0.482 | 0.503 | **0.962** | 0.122 | 730.19 |
| | SUDE | 0.748 | 0.705 | **0.608** | 0.943 | 0.188 | 96.63 ($k_1$=50)/61.19 ($k_1$=100) |
| AG's News | BH-t-SNE | 0.965 | 0.743 | 0.594 | 0.869 | **0.536** | 506.00 |
| | UMAP | 0.985 | 0.985 | 0.990 | 0.973 | 0.232 | 223.00 |
| | TriMap | 0.978 | 0.963 | 0.976 | 0.975 | 0.241 | 128.00 |
| | TopoAE | 0.991 | 0.992 | 0.988 | 0.983 | 0.186 | 1147.38 |
| | P-UMAP | 0.987 | 0.782 | 0.982 | **0.989** | 0.186 | 1087.75 |
| | SUDE | **0.994** | **0.994** | **0.994** | 0.972 | 0.337 | **62.13** ($k_1$=50)/31.64 ($k_1$=100) |

## References

[1] Li, K. & Malik, J. Fast k-nearest neighbour search via dynamic continuous indexing. In *Proceedings of the 33rd International Conference on Machine Learning*, pp. 671-679, (2016).

[2] Beyer, K., Goldstein, J., Ramakrishnan, R. & Shaft, U. When is "nearest neighbor" meaningful?. In *Proceedings of the 7th International Conference on Database Theory*, pp. 217-235, (1999).

[3] Aggarwal, C. C., Hinneburg, A. & Keim, D. A. On the surprising behavior of distance metrics in high dimensional space. In *Proceedings of the 8th International Conference on Database Theory*, pp. 420-434, (2001).

[4] Mani, P. et al. The hubness phenomenon in high-dimensional spaces. In *Research in Data Science* (eds Gasparovic, E. & Domeniconi, C.) vol 17, pp. 15-45 (2019).

[5] Mani, P. & Domeniconi, C. Hub-based subspace clustering. *Neurocomputing* **413**, 193-209 (2020).

[6] Verleysen, M. & François, D. The curse of dimensionality in data mining and time series prediction. In *International Work-Conference on Artificial Neural Networks*, pp. 758-770, (2005).

[7] Bach, F. Breaking the curse of dimensionality with convex neural networks. *J. Mach. Learn. Res.* **18**, 1-53 (2017).

[8] Peng, D., Gui, Z. & Wu, H. Interpreting the curse of dimensionality from distance concentration and manifold effect. Preprint at https://arxiv.org/abs/2401.00422 (2023).

[9] Altman, N. & Krzywinski, M. The curse(s) of dimensionality. *Nat. Methods* **15**, 399-400 (2018).

[10] Debie, E. & Shafi, K. Implications of the curse of dimensionality for supervised learning classifier systems: theoretical and empirical analyses. *Pattern Anal. Appl.* **22**, 519-536 (2019).

[11] Meilă, M., & Zhang, H. Manifold learning: What, how, and why. *Annu. Rev. Stat. Appl.* **11**, 393-417 (2024).

[12] Hall, P., Marron, J. S. & Neeman, A. Geometric representation of high dimension, low sample size data. *J. R. Stat. Soc. Ser. B Stat. Methodol.* **67**, 427-444 (2005).

[13] Ahn, J., Marron, J. S., Muller, K. M. & Chi, Y. Y. The high-dimension, low-sample-size geometric representation holds under mild conditions. *Biometrika* **94**, 760-766 (2007).

[14] Jung, S. & Marron, J. S. PCA consistency in high dimension, low sample size context. *Ann. Statist.* **37**, 4104-4130 (2009).

[15] Shen, D., Shen, H. & Marron, J. S. A General Framework for Consistency of Principal Component Analysis. *J. Mach. Learn. Res.* **17**, 1-34 (2016).

[16] Tomasev, N., Radovanovic, M., Mladenic, D. & Ivanovic, M. The role of hubness in clustering high-dimensional data. *IEEE Trans. Knowl. Data Eng.* **26**, 739-751 (2013).

[17] Radovanovic, M., Nanopoulos, A. & Ivanovic, M. Hubs in space: Popular nearest neighbors in high-dimensional data. *J. Mach. Learn. Res.* **11**, 2487-2531 (2010).

[18] Brock, A., Donahue, J. & Simonyan, K. Large scale GAN training for high fidelity natural image synthesis. Preprint at https://doi.org/10.48550/arXiv.1809.11096 (2018).

[19] Liang, Y., Wu, J., Lai, Y. K. & Qin, Y. Exploring and exploiting hubness priors for high-quality GAN latent sampling. In *Proceedings of the 39th International Conference on Machine Learning*, pp. 13271-13284 (2022).

[20] Ning, K., Ng, H. K., Srihari, S., Leong, H. W. & Nesvizhskii, A. I. Examination of the relationship between essential genes in PPI network and hub proteins in reverse nearest neighbor topology. *BMC Bioinf.* **11**, 1-14 (2010).

[21] Radovanović, M., Nanopoulos, A. & Ivanović, M. Nearest neighbors in high-dimensional data: The emergence and influence of hubs. *In Proceedings of the 26th International Conference on Machine Learning*, pp. 865-872 (2009).

[22] Radovanović, M., Nanopoulos, A. & Ivanović, M. Reverse nearest neighbors in unsupervised distance-based outlier detection. *IEEE Trans. Knowl. Data Eng.* **27**, 1369-1382 (2014).

[23] Yang, Z., King, I., Xu, Z. & Oja, E. Heavy-tailed symmetric stochastic neighbor embedding. In *Advances in Neural Information Processing Systems*, pp. 2169-2177 (2009).

[24] Kobak, D., Linderman, G., Steinerberger, S., Kluger, Y. & Berens, P. Heavy-tailed kernels reveal a finer cluster structure in t-SNE visualisations. In *Joint European Conference on Machine Learning and Knowledge Discovery in Databases* (eds Brefeld, U., Fromont, E., Hotho, A., Knobbe, A., Maathuis, M., Robardet, C.) pp. 124-139 (2019).

[25] Saul, L. K. & Roweis, S. T. An introduction to locally linear embedding. (2000). Available at: https://www.stat.cmu.edu/~cshalizi/350/lectures/14/lecture-14.pdf.

[26] Yin, H. et al. Unlocking biological insights from differentially expressed Genes: Concepts, methods, and future perspectives. *J. Adv. Res.*, (2024). https://doi.org/10.1016/j.jare.2024.12.004.

[27] Osabe, T., Shimizu, K. & Kadota, K. Differential expression analysis using a model-based gene clustering algorithm for RNA-seq data. *BMC Bioinf.* **22**, 1-20 (2021).

[28] Zhao, B., Erwin, A. & Xue, B. How many differentially expressed genes: a perspective from the comparison of genotypic and phenotypic distances. *Genomics* **110**, 67-73 (2018).

[29] Grabski, I. N., Street, K. & Irizarry, R. A. Significance analysis for clustering with single-cell RNA-sequencing data. *Nat. Methods* **20**, 1196-1202 (2023).

[30] Joulin, A., Grave, E., Bojanowski, P. & Mikolov, T. Bag of tricks for efficient text classification. Preprint at https://arxiv.org/abs/1607.01759 (2016).

[31] Jacobs, R. A. Increased rates of convergence through learning rate adaptation. *Neural Netw.* **1**, 295-307 (1988).

[32] Kingma, D. P. & Ba, J. Adam: A method for stochastic optimization. Preprint at https://doi.org/10.48550/arXiv.1412.6980 (2014).